\RequirePackage{fix-cm}
\documentclass[twocolumn]{svjour3}          % twocolumn
\smartqed  % flush right qed marks, e.g. at end of proof

\usepackage{graphicx}

\usepackage{url}
\usepackage{xcolor}
\usepackage{hyperref}
 \hypersetup{
     colorlinks=true,
     linkcolor=orange,
     filecolor=orange,
     citecolor=orange,      
     urlcolor=orange,
     }
\usepackage{color, colortbl}
\usepackage{amsfonts}
\usepackage{amsmath}
\usepackage{algorithm, setspace}
\usepackage{algpseudocode}
\usepackage{csquotes}
\usepackage{amsfonts}
\usepackage{booktabs}
\usepackage{siunitx}

\newcommand{\p}[1]{\medskip \noindent \textbf{{#1}.}}

\journalname{Autonomous Robots}

\begin{document}

\title{Learning Latent Actions to Control Assistive Robots%\thanks{Grants or other notes
}
% \subtitle{Do you have a subtitle?\\ If so, write it here}


\author{Dylan P. Losey \and
        Hong Jun Jeon \and Mengxi Li \and Krishnan Srinivasan \and Ajay Mandlekar \and Animesh Garg \and Jeannette Bohg \and Dorsa Sadigh
}


\institute{D. Losey \at
              Mechanical Engineering Department, Virginia Tech \\
              \email{losey@vt.edu}           %  \\
           \and
           H. Jeon, M. Li, K. Srinivasan, A. Mandlekar, J. Bohg, and D. Sadigh \at
              Computer Science Department, Stanford University
            \and 
            A. Garg \at
            Computer Science Department, University of Toronto
}


\maketitle

\begin{abstract}

Assistive robot arms enable people with disabilities to conduct everyday tasks on their own. These arms are dexterous and \textit{high-dimensional}; however, the interfaces people must use to control their robots are \textit{low-dimensional}. Consider teleoperating a $7$-DoF robot arm with a $2$-DoF joystick. The robot is helping you eat dinner, and currently you want to cut a piece of tofu. Today's robots assume a pre-defined mapping between joystick inputs and robot actions: in one mode the joystick controls the robot's motion in the $x$-$y$ plane, in another mode the joystick controls the robot's $z$-$yaw$ motion, and so on. But this mapping misses out on the task you are trying to perform! Ideally, one joystick axis should control how the robot stabs the tofu, and the other axis should control different cutting motions. Our insight is that we can achieve intuitive, user-friendly control of assistive robots by \textit{embedding} the robot's high-dimensional actions into low-dimensional and human-controllable \textit{latent actions}. We divide this process into three parts. First, we explore models for learning latent actions from offline task demonstrations, and formalize the properties that latent actions should satisfy. Next, we combine learned latent actions with autonomous robot assistance to help the user reach and maintain their high-level goals. Finally, we learn a personalized alignment model between joystick inputs and latent actions. We evaluate our resulting approach in four user studies where non-disabled participants reach marshmallows, cook apple pie, cut tofu, and assemble dessert. We then test our approach with two disabled adults who leverage assistive devices on a daily basis.

\end{abstract}

\keywords{Assistive Robotics \and Teleoperation \and Shared Autonomy \and Latent Representations}

\section{Introduction}

\begin{figure*}[t]
	\begin{center}
		\includegraphics[width=2.0\columnwidth]{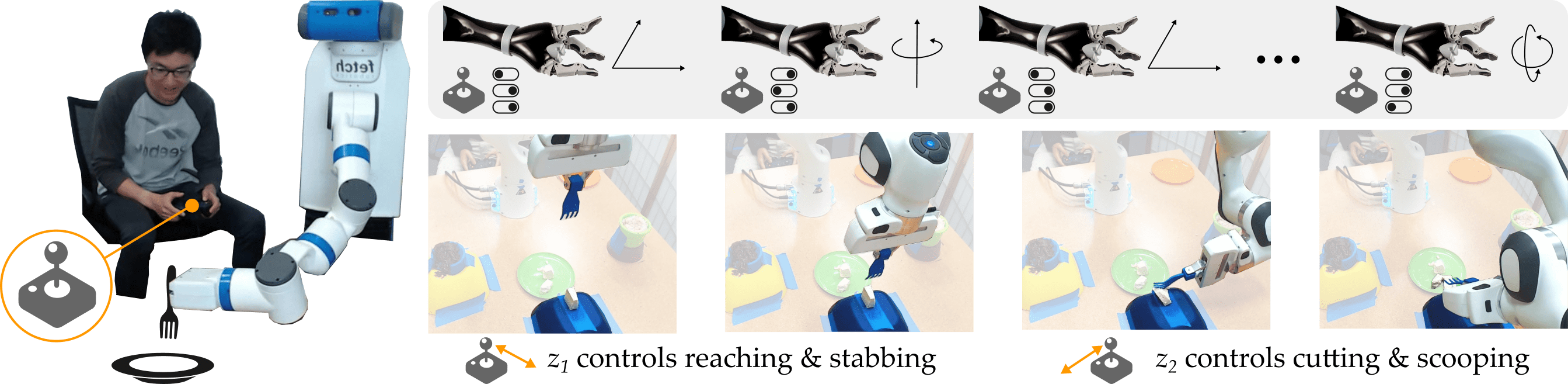}
		
		\caption{Our approach makes it easier for users to control assistive robots. (Left) assistive robot arms are dexterous and high-dimensional, but humans must teleoperate these robots with low-dimensional interfaces, such as $2$-DoF joysticks. (Right) we focus on assistive eating tasks; for example, trying to get a piece of tofu. (Top) existing work maps joystick inputs to end-effector motion. Here the user must toggle back and forth between multiple modes to control their desired end-effector motion. (Bottom) we learn task specific mapping that embeds the robot's high-dimensional actions into low-dimensional \textit{latent actions} $z$. Now pressing up and down controls the robot along a reaching and stabbing motion, while pressing right and left moves the robot arm through precise cutting and scooping motions. The user no longer needs to change modes.}
		\label{fig:front}
	\end{center}

\end{figure*}

For over one million American adults living with physical disabilities, performing everyday tasks like grabbing a bite of food or pouring a glass of water presents a significant challenge \cite{taylor2018americans}. Assistive devices --- such as wheelchair-mounted robot arms --- have the potential to improve these people's independence and quality of life \cite{argall2018autonomy,jacobsson2000people,mitzner2018closing,carlson2013brain}. A key advantage of these robots is their \textit{dexterity}: assistive arms move along multiple degrees-of-freedom (DoFs), orchestrating complex motions like stabbing a piece of tofu or pouring a glass of water. Unfortunately, this very dexterity makes assistive arms hard to control.

Imagine that you are leveraging an assistive robot arm to eat dinner (see Figure~\ref{fig:front}). You want the robot to reach for some tofu on the table in front of you, cut off a piece, and then pick it up with its fork. Non-disabled persons can use their own body to show the robot how to perform this task: for instance, the human grabs the tofu with their own arm, and the robot mimics the human's motion \cite{rakita2017motion,rakita2019shared}. But mimicking is not feasible for people living with physical disabilities --- instead, these users are limited to \textit{low-dimensional} controllers. Today's assistive robot arms leverage joysticks \cite{herlant2016assistive}, sip-and-puff devices \cite{argall2018autonomy}, or brain-computer interfaces \cite{muelling2017autonomy}. So to get a bite of tofu, you must carefully coordinate the dexterous robot arm while only pressing up-down-left-right on a joystick. Put another way, users are challenged by an inherent mismatch between low-dimensional interfaces and high-dimensional robots.

Existing work on assistive robots tackles this problem with \textit{pre-defined} mappings between user inputs and robot actions. These mappings incorporate modes, and the user switches between modes to control different robot DoFs \cite{herlant2016assistive,aronson2018eye,newman2018harmonic}. For instance, in one mode the user's $2$-DoF joystick controls the $x$-$y$ position of the end-effector, in a second mode the joystick controls the $z$-$yaw$ position of the end-effector, and so on. Importantly, these pre-defined mappings miss out on the human's underlying task. Consider teleoperating the robot to cut off a piece of tofu and then stab it with its fork. First you must use the $x$-$y$ mode to align the fork above the tofu, then $roll$-$pitch$ to orient the fork for cutting, then $z$-$yaw$ to move the fork down into the tofu, and then back to $roll$-$pitch$ to return the fork upright, and finally $z$-$yaw$ to stab the tofu --- and this is assuming you never undo a motion or make a correction! 

Controlling assistive robots becomes easier when the joystick inputs map directly to task-related motions. Within our example, one joystick DoF could produce a spectrum of stabbing motions, while the other DoF teleoperates the robot through different cutting motions. To address the fundamental mismatch between high-DoF robot arms and low-DoF control interfaces, we learn a mapping between these spaces:
\begin{center} \vspace{-0.0em}
    \emph{We make it easier to control high-dimensional robots by} embedding \emph{the robot's actions into low- dimensional and human-controllable} latent actions.
\end{center} \vspace{-0.0em}
Latent actions here refer to a low-DoF representation that captures the most salient aspects of the robot's motions. Intuitively, we can think of these latent actions as similar to the eigenvectors of a matrix composed of high-dimensional robot motions. Returning to our motivation, imagine that you have eaten dinner with the assistive robot many times. Across all of these meals there are some common motions: reaching for food items, cutting, pouring, scooping, etc. At the heart of our approach we \textit{learn} an embedding that captures these underlying motion patterns, and enables the human to \textit{control} via these learned embeddings, which we refer to as latent actions.

Overall, we make the following contributions\footnote{Parts of this work have been published at the International Conference on Robotics and Automation \cite{losey2020controlling}, Robotics: Science and Systems \cite{jeon2020shared}, and the International Conference on Intelligent Robots and Systems \cite{li2020learning}.}:

\p{Learning Latent Actions} Given a dataset of task-related robot motions, we develop a framework for learning to map the user's low-dimensional inputs to high-dimensional robot actions. For instance, imagine using a $2$-DoF joystick to teleoperate a $7$-DoF assistive robot arm. To reach for a piece of tofu, you need a mapping function --- something that interprets your joystick inputs into robot actions. Of course, not just any mapping will do; you need something that is intuitive and meaningful, so that you can easily coordinate all the robot's joints to move towards your tofu. In \textbf{Section~\ref{sec:latent-action}} we introduce a set of properties that user-friendly latent actions must satisfy, and formulate learning models that capture these properties.

\p{Integrating Shared Autonomy} But what happens once you've guided the robot to reach the tofu (i.e., your high-level goal)? Next, you need to precisely manipulate the robot arm in order to cut off a piece and pick it up with your fork. Here relying on latent actions alone is challenging, since small changes in your joystick input may accidentally move the robot away from your goal. To address this problem, in \textbf{Section~\ref{sec:shared-autonomy}} we incorporate shared autonomy, where both the human and robot arbitrate control over the robot's motion. Here the robot autonomously helps the user reach and maintain their desired high-level goals, while the user leverages latent actions to perform precise manipulation tasks (e.g., cutting, stabbing, and scooping). We show convergence bounds on the robot’s distance to the most likely goal, and develop a training procedure to ensure the human can still guide the robot to different goals if they change their mind.

\p{Personalizing Alignment} Throughout the assistive eating task your joystick inputs have produced different robot motions. To guide the robot towards the tofu, you pressed the joystick up; to orient the fork for cutting, you pressed the joystick right; and to stab your piece of tofu, you pressed the joystick down. This alignment between joystick inputs and robot outputs may make sense to you --- but different users will inevitably have different preferences! Accordingly, in \textbf{Section~\ref{sec:alignment}} we leverage user expectations to personalize the alignment between joystick directions and latent actions. Part of this alignment process involves asking the human what they prefer (e.g., what joystick direction should correspond to scooping?). We minimize the number of queries by formalizing and leveraging the priors that humans expect when controlling robotic systems.

\p{Conducting User Studies} In order to compare our approach to the state-of-the-art, we performed four user studies inspired by assistive eating tasks. Non-disabled participants teleoperated a $7$-DoF robot arm using a $2$-DoF joystick to reach marshmallows, make a simplified apple pie, cut tofu, and assemble dessert. We compared our latent action approach to both pre-defined mappings and shared autonomy baselines, including the HARMONIC dataset \cite{newman2018harmonic}. We found that latent actions help users complete high-level reaching and precise manipulations with their preferred alignment, resulting in improved objective and subjective performance.

\p{Evaluating with Disabled Users} We applied our proposed approach with two disabled adults who leverage assistive devices when eating on a daily basis. These adults have a combined five years of experience with assistive robot arms, and typically control their arms with pre-defined mappings. In our case study both participants cut tofu and assembled a marshmallow dessert using either learned latent actions or a pre-defined mapping. Similar to our results with non-disabled users, here latent actions helped these disabled participants more quickly and accurately perform eating tasks.

\p{Unifying Previous Research} This paper combines our earlier work from \cite{losey2020controlling,jeon2020shared,li2020learning}. We build on these preliminary results by integrating each part into an overarching formalism (\textbf{Section~\ref{sec:alg}}), demonstrating how each component relates to the overall approach, and evaluating the resulting approach with disabled members of our target population (\textbf{Section~\ref{disabled}}).
\section{Related Work}

Our approach learns latent representations of dexterous robot motions, and then combines those representations with shared autonomy to facilitate both coarse reaching and precise manipulation tasks. We apply this approach to assistive robot arms --- specifically for assistive eating --- so that users intuitively teleoperate their robot through a spectrum of eating-related tasks.

\p{Assistive Eating} Making and eating dinner without the help of a caretaker is particularly important to people living with physical disabilities \cite{mitzner2018closing,jacobsson2000people}. As a result, a variety of robotic devices and algorithms have been developed for assistive eating \cite{brose2010role,naotunna2015meal}. We emphasize that these devices are \textit{high-dimensional} in order to reach and manipulate food items in 3D space \cite{argall2018autonomy}. When considering how to control these devices, prior works break the assistive eating task into three parts: i) \textit{reaching} for the human's desired food item, ii) \textit{manipulating} the food item to get a bite, and then iii) \textit{returning} that bite back to the human's mouth. Recent research on assistive eating has explored \textit{automating} this process: here the human indicates what type of food they would like using a visual or audio interface, and then the robot autonomously reaches, manipulates, and returns a bite of the desired food to the user \cite{feng2019robot,park2019toward,gallenberger2019transfer,gordon2019adaptive,canal2016personalization}. However, designing a fully autonomous system to handle a task as variable and personalized as eating is exceedingly challenging: consider aspects like bite size or motion timing. Indeed --- when surveyed in \cite{bhattacharjee2020more} --- users with physical disabilities indicated that they preferred \textit{partially autonomy} during eating tasks, since this better enables the user to convey their own preferences. In line with these findings, we develop a partially autonomous algorithm that assists the human while letting them maintain control over the robot's motion.

\p{Latent Representations} Carefully orchestrating complex movements of high-dimensional robots is difficult for humans, especially when users are limited to a low-dimensional control interface \cite{bajcsy2018learning}. Prior work has tried to prune away unnecessary control axes in a data-driven fashion through Principal Component Analysis \cite{ciocarlie2009hand,artemiadis2010emg,matrone2012real}. Here the robot records demonstrated motions, identifies the first few eigenvectors, and leverages these eigenvectors to map the human's inputs to high-dimensional motions. But PCA produces a \textit{linear} embedding --- and this embedding remains constant, regardless of where the robot is or what the human is trying to accomplish. To capture intricate \textit{non-linear} embeddings, we turn to recent works that learn latent representations from data \cite{jonschkowski2014state}. Robots can learn low-dimensional models of states \cite{pacelli2020learning}, dynamics \cite{watter2015embed,xie2020learning}, movement primitives \cite{noseworthy2020task}, trajectories \cite{reyes2018self}, plans \cite{lynch2019learning}, policies \cite{edwards2019imitating}, skills \cite{pertsch2020accelerating}, and action representations for reinforcement learning \cite{chandak2019learning}. One common theme across all of these works is that there are underlying patterns in high-DoF data, and the robot can succinctly capture these patterns with a low-DoF latent space. A second connection is that these works typically leverage \textit{autoencoders} \cite{kingma2013auto,doersch2016tutorial} to learn the latent space. Inspired by these latent representation methods, we similarly adapt an autoencoder model to extract the underlying pattern in high-dimensional robot motions. But unlike prior methods, we give the human control over this embedding --- putting a human-in-the-loop for assistive teleoperation.

\p{Shared Autonomy} Learning latent representations provides a mapping from low-dimensional inputs to high-dimensional actions. But how do we \textit{combine} this learned mapping with control theory to ensure that the human can accurately complete their desired task? Prior work on assistive arms leverages shared autonomy, where the robot's action is a combination of the human's input and autonomous assistance \cite{dragan2013policy,javdani2018shared,jain2019probabilistic,broad2020data}. Here the human controls the robot with a low-DoF interface (typically a joystick), and the robot leverages a \textit{pre-defined} mapping with toggled modes to convert the human's inputs into end-effector motion \cite{aronson2018eye,herlant2016assistive,newman2018harmonic}. To assist the human, the robot maintains a belief over a discrete set of possible goal objects in the environment: the robot continually updates this belief by leveraging the human's joystick inputs as evidence in a Bayesian framework \cite{dragan2013policy,javdani2018shared,jain2019probabilistic,gopinath2016human,nikolaidis2017human}. As the robot becomes increasingly confident in the human's goal, it provides assistance to autonomously guide the end-effector towards that target. We emphasize that so far the robot has employed a pre-defined input mapping --- but more related to our approach are \cite{reddy2018shared,broad2020data,reddy2018you}, where the robot proposes or learns suitable dynamics models to translate user inputs to robot actions. For instance, in \cite{reddy2018shared,broad2020data} the robot leverages a reinforcement learning framework to identify how to interpret and assist human inputs. Importantly, here the input space has the same number of dimensions as the action space, and so no embedding is required. We build upon this previous research in shared autonomy by helping the user reach and maintain their high-level goals, but we do so by leveraging latent representations to learn a mapping from \textit{low}-DoF human inputs to \textit{high}-DoF robot outputs.
\section{Problem Setting}
\label{sec:problem}

We consider settings where a human user is teleoperating an assistive robot arm. The human interacts with the robot using a low-dimensional interface: this could be a joystick, sip-and-puff device, or brain-computer interface. We specifically focus on interfaces with a continuous control input (or an input that could be treated as continuous). For clarity, we will assume the teleoperation interface is a \textit{joystick} throughout the rest of the paper, and we will use a joystick input in all our experiments. The assistive robot's first objective is to \textit{map} these joystick inputs to meaningful high-dimensional motions. But assistive robots can do more than just interpret the human's inputs --- they can also act autonomously to help the user reach and maintain their goals. Hence, the robot's second objective is to \textit{integrate} the learned mapping with shared autonomy. In practice, the mappings that the robot learns for one user may be counter-intuitive for another. Our final objective is to \textit{align} the human's joystick inputs with the latent actions, so that users can intuitively convey their desired motions through the control interface. 

In this section we formalize our problem setting, and outline our proposed solutions to each objective. We emphasize the main variables in Table~\ref{table:def}.

\begin{table}[t]
\label{table:def}
\caption{Key Variables and their Definition}
\begin{center}
\begin{tabular}{ l l }
    % \caption{Key Variables and their Definition}
    % \label{table:def}
    \toprule
    $s$ & robot's state (or the world's state) \\
    $b$ & robot's belief over high-level goals $g \in \mathcal{G}$ \\
    $c$ & robot's context: we consider $c=s, ~c=(s, b)$ \\ 
    $u$ & human's joystick input \\
    $z$ & latent action commanded by the human \\
    $f$ & alignment model $z = f(u, c)$ \\
    $a_h$ & human's commanded high-DoF robot action \\
    $\phi$ & learned decoder $a_h = \phi(z, c)$ \\
    $a_r$ & autonomous assistive action \\
    $a$ & robot's action, where $a = (1-\alpha)\cdot a_h + \alpha \cdot a_r$ \\
    \bottomrule
\end{tabular}
\end{center}
\end{table}

\p{Task} The human operator has a task in mind that they want the robot to accomplish. We formulate this task as a Markov decision process: $\mathcal{M} = (\mathcal{S}, \mathcal{A}, \mathcal{T}, R, \gamma, \rho_0)$. Here $s \in \mathcal{S} \subseteq \mathbb{R}^n$ is the state and $a \in \mathcal{A} \subseteq \mathbb{R}^m$ is the robot's high-DoF action. Because we are focusing on the high-dimensional robot arm, we refer to $s$ as the robot's state, but in practice the state $s$ may contain both the robot's arm position and the location of other objects in the environment (e.g., the position of the tofu).

The robot transitions between states according to $\mathcal{T}(s, a)$, and receives reward $R(s)$ at each timestep. We let $\gamma \in [0, 1)$ denote the discount factor, and $\rho_0$ captures the initial state distribution. During each interaction the robot is not sure what the human wants to accomplish (i.e., the robot does not know $R$). Returning to our running example, the robot does not know whether the human wants a bite of tofu, a drink of water, or something else entirely. The human communicates their desired task through joystick inputs $u \in \mathbb{R}^d$. Because the human's input is of lower dimension than the robot's action, we know that $d < m$. 

\p{Dataset} Importantly, this is not the first time the user has guided their robot through the process of eating dinner. We assume access to a dataset of task demonstrations: these demonstrations can be kinesthetically provided by a caregiver or collected beforehand by the disabled user with a baseline teleoperation scheme. For example, the disabled user leverages their standard, pre-defined teleoperation mapping to guide the robot through the process of reaching for objects on the table (e.g., the plate, a glass of water) and manipulating these objects (e.g., scooping rice, picking up the glass). We collect these demonstrations and employ them to train our latent action approach. Formally, we have a dataset $\mathcal{D} = \{(c_0, a_0), (c_1, a_1), \ldots \}$ of context-action pairs that demonstrate high-dimensional robot actions. Notice that here we introduce the \textit{context} $c \in \mathcal{C}$: this context captures the information available to the robot. For now we can think of the context $c$ as the same as the robot's state (i.e., $c = s$), but later we will explore how the robot can also incorporate its understanding of the human's goal into this context.

\begin{figure*}[t]
	\begin{center}
		\includegraphics[width=2.0\columnwidth]{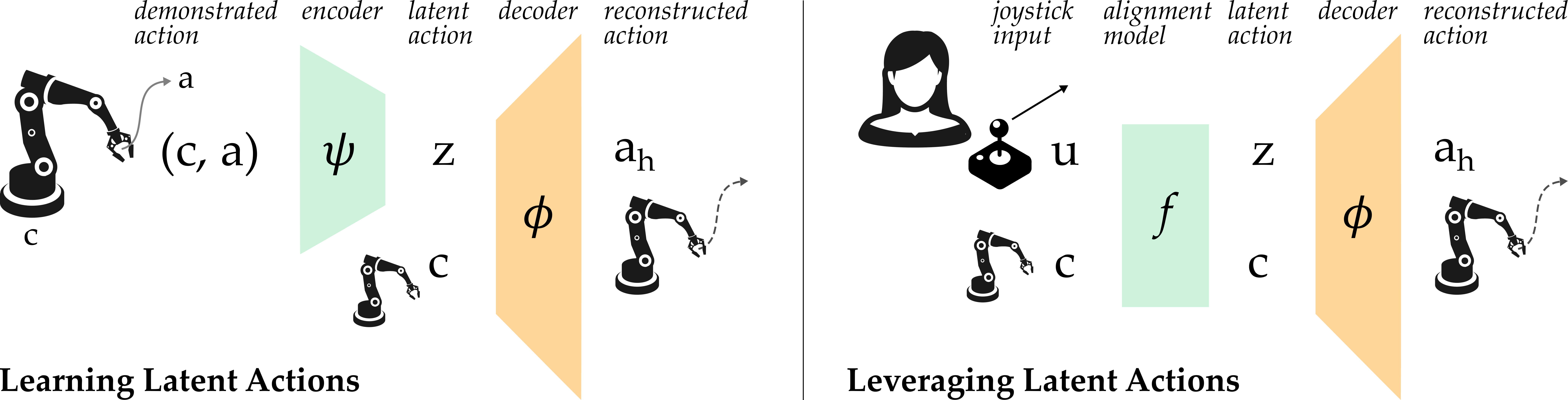}
		
		\caption{Model for learning and leveraging latent actions. (Left) given a dataset of context-action pairs, we embed the robot's high-dimensional behavior $(c,a)$ into a low-dimensional latent space $z \in \mathcal{Z}$. The encoder and decoder are trained to ensure user-friendly properties while minimizing the error between the commanded human action $a_h$ and the demonstrated action $a$. As a result, the decoder $\phi(z, c)$ provides an intuitive mapping from low-dimensional latent actions to high-dimensional robot actions. (Right) at run time the human controls the robot via these low-dimensional latent actions. For now we simplify the alignment model so that $z = u$, meaning that the human's joystick inputs directly map to latent actions.}
		\label{fig:model}
	\end{center}
\end{figure*}

\p{Latent Actions} Given this dataset, we first learn a latent action space $\mathcal{Z} \subset \mathbb{R}^d$, as well as a \textit{decoder} function $\phi : \mathcal{Z} \times \mathcal{C} \rightarrow \mathcal{A}$. Here $\mathcal{Z}$ is a low-dimensional embedding of $\mathcal{D}$ --- we specify the dimensionality of $\mathcal{Z}$ to match the number of degrees-of-freedom of the joystick, so that the user can directly input latent actions $z \in \mathcal{Z}$. Based on the human's latent action $z$ as well as the current \textit{context} $c \in \mathcal{C}$, the robot leverages the decoder $\phi$ to reconstruct a high-dimensional action (see Figure~\ref{fig:model}): 
\begin{equation} \label{eq:p1}
a_h = \phi(z, c)    
\end{equation}
Notice that we use $a_h$ here: this is because this robot action is commanded by the human's input. Consider pressing the joystick right to cause the robot arm to cut some tofu. The joystick input is a low-DoF input $u$, we map this input to a latent action $z \in \mathcal{Z}$, and then leverage $\phi$ to decode $z$ into a high-DoF commanded action $a_h$ that cuts the tofu. We formalize properties of $\mathcal{Z}$ and models for learning $\phi$ in \textbf{Section~\ref{sec:latent-action}}.

\p{Shared Autonomy} The human provides joystick inputs $u$ --- which we treat as latent actions $z$ --- and these latent actions map to high-dimensional robot actions $a_h$. But how can the robot assist the human through its own autonomous behavior? More formally, how should the robot choose autonomous actions $a_r$ that help guide the user? Similar to recent work on shared autonomy \cite{jain2019probabilistic,dragan2013policy,newman2018harmonic}, we define the robot's overall action as the linear combination of $a_h$ (the human's commanded action) and $a_r$ (the robot's autonomous guidance):
\begin{equation} \label{eq:p2}
    a = (1 - \alpha) \cdot a_h + \alpha \cdot a_r
\end{equation}
In the above, $\alpha \in [0, 1]$ parameterizes the trade-off between direct human teleoperation ($\alpha = 0$) and complete robot autonomy ($\alpha = 1$). We specifically focus on autonomous actions that help the user reach and maintain their high-level goals. Let $\mathcal{G}$ be a discrete set of goal positions the human might want to reach (e.g., their tofu, the rice, or a glass of water), and let $g^* \in \mathcal{G}$ be the human's true goal (e.g., the tofu). The robot assists the user towards goals it thinks are likely:
\begin{equation} \label{eq:p3}
    a_r = \sum_{g \in \mathcal{G}} b(g) \cdot (g - s)
\end{equation}
Here $a_r$ is a change of state (i.e., a joint velocity) that moves from $s$ towards the mode of the inferred goal position, and $b$ denotes the robot's \textit{belief}. This belief is a probability distribution over the candidate goals, where $b(g) = 1$ indicates that the robot is completely convinced that $g$ is what the human wants. We analyze dynamics of combining Equations~(\ref{eq:p1}-\ref{eq:p3}) in \textbf{Section~\ref{sec:shared-autonomy}}. 

\p{Alignment} Recall that the human's joystick input is $u$, and that our approach treats this joystick input as a latent actions $z$. A naive robot will simply set $z = u$. But this misses out on how different users \textit{expect} the robot to interpret their commands. For example, let us say the assistive robot is directly above some tofu. One user might expect pressing right to cause a \textit{stabbing} motion, while a second user expects pressing right to \textit{cut} the tofu. To personalize our approach to match individual user expectations, we learn an alignment function:
\begin{equation} \label{eq:P4}
    z = f(u, c)
\end{equation}
Unlike Equation (\ref{eq:p1}), this is \textit{not} an embedding, since both $u$ and $z$ have $d$ dimensions. But like Equation (\ref{eq:p1}), the alignment model \textit{does} depend on the robot's current context. A user might expect pressing right to stab the tofu when they are directly above it --- but when they are interacting with a glass of water, that same user expects pressing right to tilt and pour the glass. We learn $f$ across different contexts in \textbf{Section~\ref{sec:alignment}}.
\section{Learning Latent Actions}
\label{sec:latent-action}

\begin{figure*}[t]
	\begin{center}
		\includegraphics[width=2.0\columnwidth]{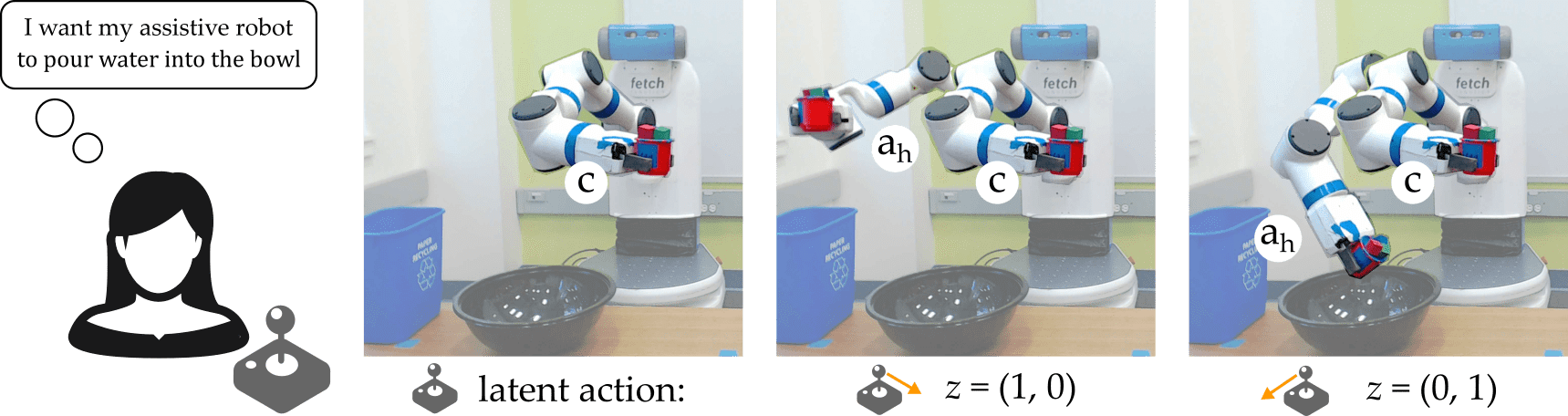}
		
		\caption{Controlling an assistive robot with learned latent actions. The robot has been trained on demonstrations of pouring tasks, and learns a $2$-DoF latent space. One axis of the latent space moves the cup across the table, and the other latent dimension pours the cup. This latent space satisfies our \textit{conditioning} property because the decoded action $a_h$ depends on the current context $c$. This latent space also satisfies \textit{controllability} because the human can leverage the two learned latent dimensions to complete the demonstrated pouring tasks (e.g., pour water into the bowl).}
		\label{fig:latent_action_task}
	\end{center}

\end{figure*}

In this section we focus on learning latent actions. We define latent actions as a low-dimensional embedding of high-dimensional robot actions, and we learn this embedding from the dataset $\mathcal{D}$ of offline demonstrations. Overall, we will search for two things: i) a latent action space $\mathcal{Z} \subset \mathbb{R}^d$ that is of lower dimension than the robot's action space $\mathcal{A} \subseteq \mathbb{R}^m$, and ii) a decoder function $\phi$ that maps from latent actions to robot actions. In practice, latent actions provide users a non-linear mapping for robot control: e.g., pressing right on the joystick causes the robot to perform a stabbing motion. We emphasize that these latent actions do not always have semantic meanings --- they are not always ``stabbing'' or ``cutting'' --- but generally embed the robot's high-dimensional motion into a low-dimensional space.

Recall our motivating example, where the human is trying to teleoperate their assistive robot using a joystick to reach and manipulate food items. When controlling the robot, there are several \textit{properties} that the human expects: e.g., smooth changes in the joystick input should not cause abrupt changes in robot motion, and when the human holds the joystick in a constant direction, the robot's motion should not suddenly switch direction. In what follows, we first \textit{formalize} the properties necessary for latent actions to be intuitive. These properties will guide our approach, and provide a principled way of assessing the usefulness of latent actions with humans-in-the-loop. Next, we will explore different \textit{models} for learning latent actions that capture our intuitive properties.

\subsection{Latent Action Properties}

We identified four properties that user-friendly latent actions should have: conditioning, controllability, consistency, and scalability.

\p{Conditioning} Because $\phi$ maps from latent actions to robot actions, at first glance it may seem intuitive for $\phi$ to \textit{only} depend on $z$, i.e., $a_h = \phi(z)$. But this quickly breaks down in practice. Imagine that you are controlling the robot arm to get some tofu: at the start of the task, you press right and left on the joystick to move the robot towards your target. But as the robot approaches the tofu, you no longer need to keep moving towards a goal; instead, you need to use those same joystick inputs to carefully align the orientation of the fork, so that you can cut off a piece. Hence, latent actions must convey different meanings in different contexts. This is especially true because the latent action space is smaller than the robot action space, and so there are more actions to convey than we can capture with $z$ alone. We therefore introduce $c \in \mathcal{C}$, the robot's current context, and \textit{condition} the decoder on $c$, so that $a_h = \phi(z, c)$. In the rest of this section we will treat the robot's state $s$ as its context, so that $c = s$.

\p{Controllability} For latent actions to be useful, human operators must be able to leverage these actions to control the robot through their desired task. Recall that the dataset $\mathcal{D}$ includes relevant task demonstrations, such as picking up a glass, reaching for the kitchen shelf, and scooping rice. We want the user to be able to control the robot through these same tasks when leveraging latent actions. Let $s_i, s_j \in \mathcal{D}$ be two states from the dataset of demonstrations, and let $s_{1}, s_{2}, ..., s_{K}$ be the sequence of states that the robot visits when starting in state $s_0 = s_i$ and taking latent actions $z_1, ..., z_K$. The robot transitions between the visited states using its transition function $\mathcal{T}$ and the learned decoder $\phi$: $s_{k} = \mathcal{T}(s_{k-1}, \phi(z_{k-1}, s_{k-1}))$. Formally, we say that a latent action space $\mathcal{Z}$ is \textit{controllable} if for every pair of states $(s_i, s_j)$ there exists a sequence of latent actions $\{z_k\}_{k=1}^K, z_k \in \mathcal{Z}$ such that $s_j = s_K$. In other words, a latent action space is controllable if it can move the robot between pairs of start and goal states from the demonstrated tasks.

\p{Consistency} Let us say you are using a one-DoF joystick to guide the robot arm along a line. When you hold the joystick to the right, you expect the robot to immediately move right --- but more than that, you expect the robot to move right at every point along the line! For example, the robot should not move right for a while, then suddenly go left, and switch back to going right again. To capture this, we define a latent action space $\mathcal{Z}$ as \textit{consistent} if the same latent action $z \in \mathcal{Z}$ has a similar effect on how the robot behaves in nearby states. We formulate this similarity via a task-dependent metric $d_M$: e.g., in pouring tasks $d_M$ could measure the orientation of the robot's end-effector, and in reaching tasks $d_M$ could measure the position of the end-effector. Applying this metric, consistent latent actions should satisfy: 
\begin{equation}
    d_M(\mathcal{T}(s_1,\phi(z, s_1)),\mathcal{T}(s_2,\phi(z, s_2))) < \epsilon    
\end{equation}
when $\|s_1 - s_2\| < \delta$ for some $\epsilon, \delta>0$. We emphasize that we do not need to know $d_M$ for our approach; we only introduce this metric as a way of quantifying similarity.

\p{Scalability} Our last property is complementary to consistency. Thinking again about the example of teleoperating a robot along a line, when you press the joystick slightly to the right, you expect the robot to move slowly; and when you hold the joystick all the way to the right, you anticipate that the robot will move quickly. Smaller inputs should cause smaller motions, and larger inputs should cause larger motions. Formally, we say that a latent action space $\mathcal{Z}$ is \textit{scalable} if $\| s-s'\| \to \infty$ as $\| z \| \to \infty$, where $s'=\mathcal{T}(s, \phi(z, s))$. When put together, our consistency and scalability properties ensure that the decoder function $\phi$ is Lipschitz continuous.

\subsection{Models for Learning Latent Actions}

Now that we have formally introduced the properties that a user-friendly latent space should satisfy, we will explore low-DoF embeddings that capture these properties; specifically, models which learn $\phi :\mathcal{Z} \times \mathcal{C} \rightarrow \mathcal{A}$ from offline demonstrations $\mathcal{D}$. We are interested in models that balance \textit{expressiveness} with \textit{intuition}: the embedding must reconstruct high-DoF actions while remaining controllable, consistent, and scalable. We assert that only models which reason over the robot's context when decoding the human's inputs can accurately and intuitively interpret the latent action. Our overall model structure is outlined in Figure~\ref{fig:model}.

\p{Reconstructing Actions} Let us return to our assistive eating example: when the person applies a low-dimensional joystick input, the robot completes a high-dimensional action. We use \textit{autoencoders} to move between these low- and high-DoF action spaces. Define $\psi : \mathcal{C} \times \mathcal{A} \rightarrow \mathcal{Z}$ as an \textit{encoder} that embeds the robot's behavior into a latent space, and define $\phi : \mathcal{Z} \rightarrow \mathcal{A}$ as a \textit{decoder} that reconstructs a high-DoF robot action $a_h$ from this latent space. Intuitively, the reconstructed robot action $a_h$ should match the demonstrated action $a$. To encourage models to learn latent actions that accurately reconstruct high-DoF robot behavior, we incorporate the reconstruction error $\|a - a_h\|^2$ into the model's loss function. Let $\mathcal{L}$ denote the \textit{loss function} our model is trying to minimize; when we only focus on reconstructing actions, our loss function is: 
\begin{equation} \label{eq:ae}
    \mathcal{L} = \|a - \phi(\psi(c, a)) \|^2
\end{equation}
Both principal component analysis (\textbf{PCA}) and autoencoder (\textbf{AE}) models minimize this loss function.

\p{Regularizing Latent Actions} When the user slightly tilts the joystick, the robot should not suddenly cut the entire block of tofu. To better ensure this consistency and scalability, we incorporate a \textit{normalization} term into the model's loss function. Let us define $\psi : \mathcal{C} \times \mathcal{A} \rightarrow \mathbb{R}^d \times \mathbb{R}_+^d$ as an encoder that outputs the mean $\mu$ and covariance $\sigma$ over the latent action space. We penalize the divergence between this latent action space and a normal distribution: $KL(\mathcal{N}(\mu, \sigma) ~\|~ \mathcal{N}(0, 1))$. When we incorporate this normalizer, our loss function becomes:
\begin{equation} \label{eq:vae}
    \mathcal{L} = \|a - \phi(z) \|^2 + \lambda \cdot KL\big[\mathcal{N}(\mu, \sigma) ~\|~ \mathcal{N}(0, 1)\big]
\end{equation}
Variational autoencoder (\textbf{VAE}) models \cite{kingma2013auto,doersch2016tutorial} minimize this loss function by trading-off between reconstruction error and normalization.

\p{Conditioning on State} Importantly, we recognize that the \textit{meaning} of the human's joystick input often depends on what part of the task the robot is performing. When the robot is above a block of tofu, pressing down on the joystick indicates that the robot should stab the food; but --- when the robot is far away from the tofu --- it does not make sense for the robot to stab! So that robots can associate meanings with latent actions, we \textit{condition} the interpretation of the latent action on the robot's current context. Define $\phi : \mathcal{Z} \times \mathcal{C} \rightarrow \mathcal{A}$ as a decoder that now makes decisions based on both $z$ and $c$. Leveraging this conditioned decoder, our final loss function is:
\begin{equation} \label{eq:cvae}
    \mathcal{L} = \|a - \phi(z,c)\|^2 + \lambda \cdot KL\big[\mathcal{N}(\mu, \sigma) ~\|~ \mathcal{N}(0, 1)\big]
\end{equation}
We expect that conditional autoencoders (\textbf{cAE}) and conditional variational autoencoders (\textbf{cVAE}) which use $\phi$ will learn more expressive and controllable actions than their non-context conditioned counterparts. Note that cVAEs minimize Equation~(\ref{eq:cvae}), while cAEs do not include the normalization term (i.e., $\lambda = 0$).

\p{Relation to Properties} Models trained to minimize the listed loss functions are encouraged to satisfy our user-friendly properties. For example, in Equation~(\ref{eq:cvae}) the decoder is \textit{conditioned} on the current context, while minimizing the reconstruction loss $\|a - \phi(z,c)\|^2$ ensures that the robot can reproduce the demonstrations, and is therefore \textit{controllable}. Enforcing \textit{consistency} and \textit{scalability} are more challenging --- particularly when we do not know the similarity metric $d_M$ --- but including the normalization term prevents the latent space from assigning arbitrary and irregular values to $z$. To better understand how these models enforce our desired properties, we conduct a set of controlled simulations in \textbf{Section~\ref{sim:icra}}.
\section{Combining Latent Actions with Shared Autonomy}
\label{sec:shared-autonomy}

\begin{figure*}[t]
	\begin{center}
		\includegraphics[width=2.0\columnwidth]{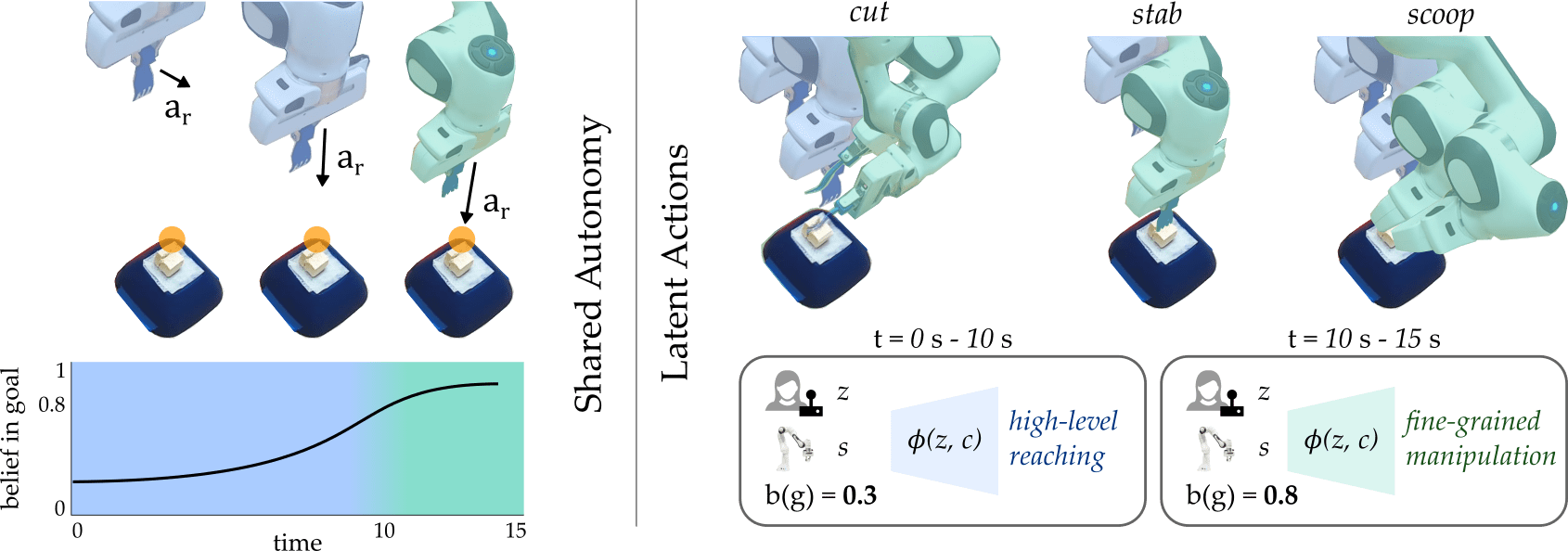}
		
		\caption{Shared autonomy with learned latent actions. (Left) as the human teleoperates the robot towards their desired goal, the robot's belief in that goal increases, and the robot selects assistive actions $a_r$ to help the human autonomously reach and maintain their high-level goal. (Right) the meaning of the latent actions changes as a function of the robot's belief. At the start of the task --- when the robot is not sure about any goal --- the latent actions $z$ produce high-level reaching motions (shown in blue). As the robot becomes confident in the human's goal the meaning of the latent actions becomes more refined, and $z$ increasingly controls fine-grained manipulation (shown in green). }
		\label{fig:shared_autonomy}
	\end{center}

\end{figure*}

The latent actions learned in \textbf{Section~\ref{sec:latent-action}} provide an expressive mapping between low-dimensional user inputs and high-dimensional robot actions. But controlling an assistive robot with latent actions alone still presents a challenge: any \textit{imprecision} or \textit{noise} in either the user's inputs or latent space is reflected in the decoded actions. Recall our eating example: at the start of the task, the human uses latent actions to guide the robot towards their \textit{high-level goal} (i.e., reaching the tofu). Once the robot is close to the tofu, however, the human no longer needs to control reaching motions --- instead, the human leverages latent actions to \textit{precisely manipulate} the tofu, performing \textit{low-level} cutting and stabbing tasks. Here the human's inputs should not unintentionally cause the robot arm to drift away from the tofu or suddenly jerk into the table. Instead, the robot should \textit{maintain} the human's high-level goal. In this section we incorporate shared autonomy alongside latent actions: this approach assists the human towards their high-level goals, and then maintains these goals as the human focuses on low-level manipulation. We visualize shared autonomy with learned latent actions in Figure~\ref{fig:shared_autonomy}.

\subsection{Latent Actions with Shared Autonomy}

We first explain how to combine latent actions with shared autonomy. Remember that the human's joystick input is $u$ and the latent action is $z$. For now we assume some pre-defined mapping from $u$ to $z$ (i.e., $z=u$) so that the human's joystick inputs are treated as latent actions. In the last section we learned a decoder $a_h = \phi(z,c)$, where the output of this decoder is a high-dimensional robot action \textit{commanded} by the human. Here we combine this commanded action with $a_r$, an autonomous \textit{assistive} action that helps the user reach and maintain their high-level goals.

\p{Belief over Goals} Similar to \cite{dragan2013policy,javdani2018shared,gopinath2016human}, we assume access to a discrete set of high-level goals $\mathcal{G}$ that the human may want to reach. Within our eating scenario these goals are food items (e.g., the tofu, rice, a plate, marshmallows). Although the robot knows which goals are possible, the robot does not know the human's current goal $g^* \in \mathcal{G}$. We let $b=P(g ~|~ s^{0:t}, u^{0:t})$ denote the robot’s belief over this space of candidate goals, where $b(g) = 1$ indicates that the robot is convinced that $g$ is the human’s desired goal. Here $s^{0:t}$ is the history of states and $u^{0:t}$ is the history of human inputs: we use Bayesian inference to update the robot's belief $b$ given the human's past decisions:
\begin{equation} \label{eq:bayes}
    b^{t+1}(g) \propto P(u^t \mid s^t, g) \cdot b^t(g)
\end{equation}
This Bayesian inference approach for updating $b$ is explored by prior work on shared autonomy \cite{jain2019probabilistic}.

Importantly, the \textit{meaning} of the human's joystick inputs changes as a function of the robot's belief. Imagine that you are using a 1-DoF joystick to get the tofu in our eating example. At the start of the task --- when the robot is unsure of your goal --- you press left and right on the joystick to move towards your high-level goal. Once you've reached the tofu --- and the robot is confident in your goal --- you need to use those same joystick inputs to carefully align the orientation of the fork. In order to learn latent action spaces that can continuously alternate along a spectrum of high-level goals and fine-grained preferences, we now condition $\phi$ on the robot's current state \textit{as well as its belief}. Hence, instead of $c=s$, we now have $c = (s, b)$. Conditioning on belief enables the meaning of latent actions to change based on the robot's confidence. As a result of this proposed structure, latent actions \textit{purely indicate the desired goal} when the robot is unsure; and once the robot is confident about the human's goal, latent actions gradually change to \textit{convey the precise manipulation}. We note that $b$ is available when collecting demonstrations $\mathcal{D}$, since the robot can compute its belief using the Bayesian update above based on the demonstrated trajectory. Take a demonstration that moves the robot to the tofu: initially the robot has a uniform belief over goals, but as the demonstration moves towards the tofu, the robot applies Equation~(\ref{eq:bayes}) to increase its belief $b$ over the tofu goal.

\p{Shared Autonomy} Recall that the robot applies assistance via action $a_r$ in Equation~(\ref{eq:p2}). In order to assist the human, the robot needs to understand the human's \textit{intent} --- i.e., which goal they want to reach. The robot's understanding of the human's intended goal is captured by belief $b$, and we leverage this belief to select an assistive action $a_r$. As shown in Equation~(\ref{eq:p3}), the robot selects $a_r$ to guide the robot towards each discrete goal $g \in \mathcal{G}$ in proportion to the robot's confidence in that goal\footnote{Our approach is not tied to this particular instantiation of shared autonomy. Other instances of shared autonomy can similarly be used.}. Combining these equations with our learned latent actions, we find the robot's overall action $a$:
\begin{equation} \label{eq:sala}
    a = (1 - \alpha) \cdot \phi(z,c) + \alpha \cdot \sum_{g \in \mathcal{G}} b(g) \cdot (g - s)
\end{equation}
Recall that $\alpha \in [0, 1]$ arbitrates between human control ($\alpha = 0$) and assistive guidance ($\alpha = 1$). In practice, if the robot has a uniform prior over which morsel of food the human wants to eat, $a_r$ guides the robot to the center of these morsels. And --- when the human indicates a desired morsel --- $a_r$ moves the robot towards that target before maintaining the target position.

\subsection{Reaching and Changing Goals}

In Equation~(\ref{eq:sala}) we incorporated shared autonomy with latent actions to tackle assistive eating tasks that require high-level reaching and precise manipulation. Both latent actions and shared autonomy have an \textit{independent} role within this method: but how can we be sure that the \textit{combination} of these tools will remain effective? Returning to our eating example --- if the human inputs latent actions, will shared autonomy correctly guide the robot to the desired morsel of food? And what if the human has multiple goals in mind (e.g., getting a chip and then dipping it in salsa) --- can the human leverage latent actions to change goals even when shared autonomy is confident in the original goal?

\p{Converging to the Desired Goal} We first explore how our approach ensures that the human reaches their desired goal. Consider the Lyapunov function:
\begin{equation} \label{eq:T1}
    V(t) = \frac{1}{2} \|e(t)\|^2, \quad e(t) = g^* - s(t)
\end{equation}
where $e$ denotes the error between the robot's current state $s$ and the human's goal $g^*$. We want the robot to choose actions that minimize Equation~(\ref{eq:T1}) across a spectrum of user skill levels and teleoperation strategies. Let us focus on the common setting in which $s$ is the robot's joint position and $a$ is the joint velocity, so that $\dot{s}(t) = a(t)$. Taking the derivative of Equation~(\ref{eq:T1}) and substituting in this transition function, we reach\footnote{For notational simplicity we choose $\alpha = 0.5$, so that both human and robot inputs are equally weighted. Our results generalize to other $\alpha$.}:
\begin{equation} \label{eq:T2}
    \dot{V}(t) = -\frac{1}{2}e^\top \Big[\phi(z, c) + \sum_{g \in \mathcal{G}}b(g)\cdot (g - s)\Big]
\end{equation}
We want Equation~(\ref{eq:T2}) to be negative, so that $V$ (and thus the error $e$) decrease over time. A sufficient condition for $\dot{V} < 0$ is:
\begin{equation} \label{eq:T3}
    b(g^*) \cdot \|e\| > \|\phi(z,c)\| + \sum_{g \in \mathcal{G}'} b(g)\cdot \|g - s\|
\end{equation}
where $\mathcal{G}'$ is the set of all goals except $g^*$. As a final step, we bound the magnitude of the decoded action, such that $\| \phi(\cdot)\| < \sigma_h$, and we define $\sigma_r$ as the distance between $s$ and the furthest goal: $\sigma_r = \max_{g \in \mathcal{G}'} \|g - s\|$. Now we have $\dot{V} < 0$ if:
\begin{equation} \label{eq:T4}
    b(g^*)\cdot \|e\| > \sigma_h + \big(1 - b(g^*)\big)\cdot \sigma_r
\end{equation}
We define $\delta := \sigma_h + \big(1 - b(g^*)\big)\cdot \sigma_r$. We therefore conclude that our approach in Equation~(\ref{eq:sala}) yields \textit{uniformly ultimately bounded stability} about the human's goal, where $\delta$ affects the \textit{radius} of this bound \cite{spong2006robot}. As the robot's confidence in $g^*$ increases, $\delta \rightarrow \sigma_h$, and the robot's error $e$ decreases so long as $\|e(t) \| > \sigma_h$. Intuitively, this guarantees that the robot will move to some ball around the human's goal $g^*$ (even if we treat the human input as a disturbance), and the radius of that ball decreases as the robot becomes more confident.

\p{Changing Goals} Our analysis so far suggests that the robot becomes \textit{constrained} to a region about the most likely goal. This works well when the human correctly conveys their intentions to the robot --- but what if the human makes a mistake, or changes their mind? How do we ensure that the robot is not \textit{trapped} at an undesired goal? Re-examining Equation~(\ref{eq:T4}), it is key that --- in every context $c$ --- the human can convey sufficiently large actions $\|\phi(z,c)\|$ towards their preferred goal, ensuring that $\sigma_h$ does not decrease to zero. Put another way, the human must be able to \textit{increase} the radius of the bounding ball, reducing the constraint imposed by shared autonomy.

To encourage the robot to learn latent actions that increase this radius, we introduce an additional term into our model's loss function $\mathcal{L}$. We reward the robot for learning latent actions that have high \textit{entropy} with respect to the goals; i.e., in a given context $c$ there exist latent actions $z$ that cause the robot to move towards \textit{each} of the goals $g \in \mathcal{G}$. Define $p_{c}(g)$ as proportional to the total \emph{score} $\eta$ accumulated for goal $g$:
\begin{equation} \label{eq:T5}
    p_{c}(g) \propto \sum_{z\in \mathcal{Z}} \eta(g, c, z)
\end{equation}
where the score function $\eta$ indicates how well action $z$ taken from context $c$ conveys the intent of moving to goal $g$, and the distribution $p_{c}$ over $\mathcal{G}$ captures the proportion of latent actions $z$ at context $c$ that move the robot toward each goal. Intuitively, $p_{c}$ captures the comparative ease of moving toward each goal: when $p_{c}(g) \rightarrow 1$, the human can easily move towards goal $g$ since \emph{all} latent actions at $c$ induce movement towards goal $g$ and consequently, \emph{no} latent actions guide the robot towards any other goals. We seek to \textit{avoid} learning latent actions where $p_{c}(g) \rightarrow 1$, because in these scenarios the teleoperator \textit{cannot} correct their mistakes or move towards a different goal. Recall from \textbf{Section~\ref{sec:latent-action}} that the model should minimize the reconstruction error while regularizing the latent space. We now argue that the model should additionally maximize the Shannon entropy of $p$, so that the loss function becomes:
\begin{multline} \label{eq:T6}
   \mathcal{L} = \|a - \phi(z,c)\|^2 + \lambda_1 \cdot KL\big[\mathcal{N}(\mu, \sigma) ~\|~ \mathcal{N}(0, 1)\big] \\ + \lambda_2 \cdot \sum_{g \in \mathcal{G}} p(g) \log{p(g)} 
\end{multline}
Here the hyperparameter $\lambda_2 > 0$ determines how much importance is assigned to maximizing the entropy over goals. When combining shared autonomy with latent actions, we employ this loss function to train the decoder $\phi$ from dataset $\mathcal{D}$.

\section{Aligning Latent Actions with User Preferences}
\label{sec:alignment}

\begin{figure}[t]

	\begin{center}
		\includegraphics[width=1.0\columnwidth]{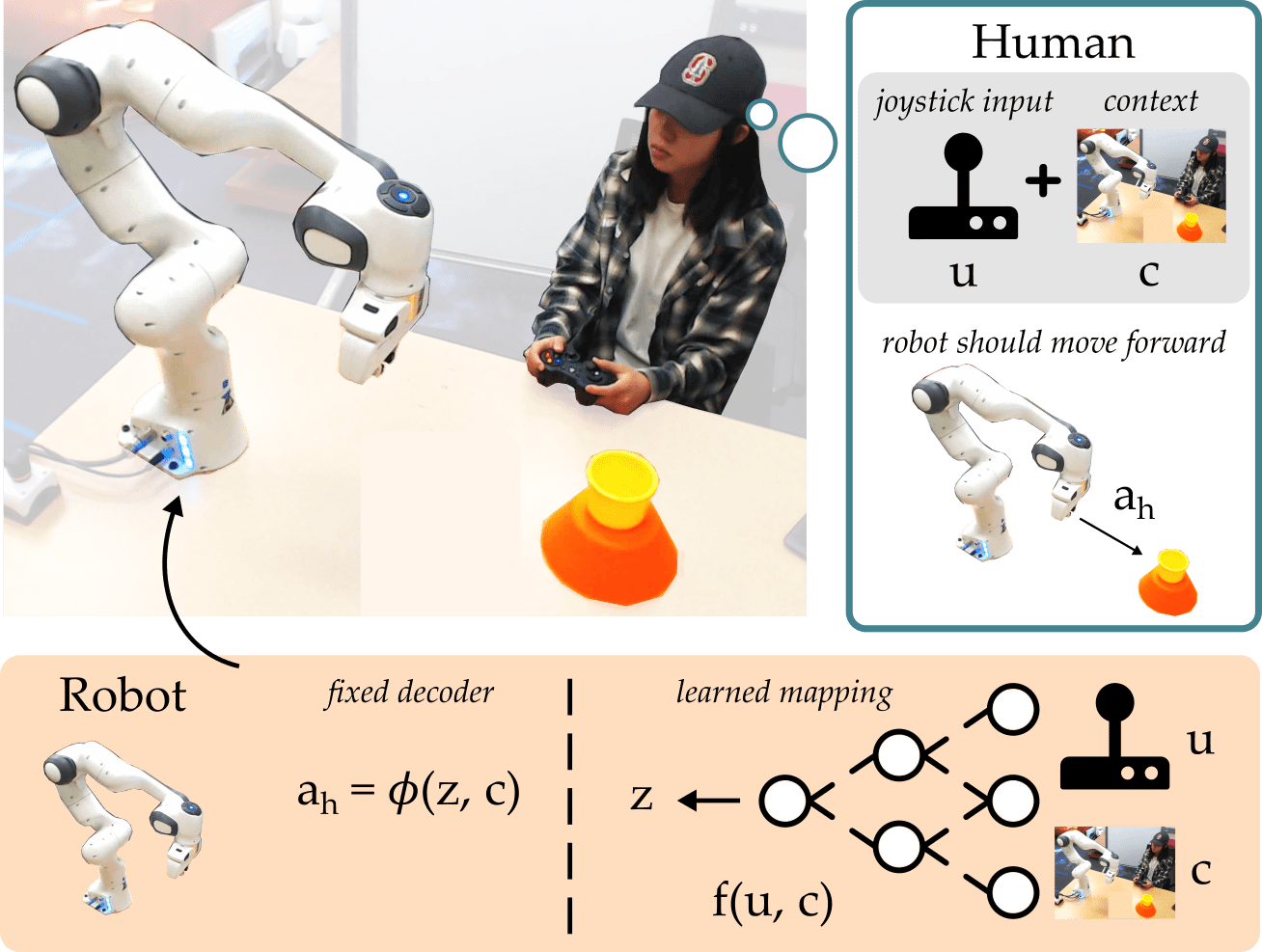}

		\caption{Overview of alignment model. The human has in mind a preferred mapping between their joystick inputs $u$ and their commanded actions $a_h$. We break this into a two step process: \textit{aligning} the joystick inputs with latent actions $z$, and then decoding $z$ into a high-DoF action $a_h$. In previous sections we focused on the decoder $\phi$; now we learn a personalized alignment model $f$. The robot learns $f$ offline in a \textit{semi-supervised} manner by combining labeled queries with intuitive priors that capture the human's underlying expectations of how the control mapping should behave.}

		\label{fig:align_front}
	\end{center}
	
\end{figure}

In \textbf{Section~\ref{sec:latent-action}} we learned latent actions, and in \textbf{Section~\ref{sec:shared-autonomy}} we combined these latent actions with shared autonomy to handle precise manipulation tasks. Throughout these sections we treated the human's joystick inputs as the latent actions (i.e., $z = u$) since both the joystick inputs and the latent actions are the same dimensionality. However, different users have different expectations for how the robot will interpret their inputs! Imagine that the shared autonomy has assisted us to the tofu, and now we want to control the robot through a cutting motion. One user expects $u = down$ to cause the robot to cut, but another person thinks $u = down$ should cause a stabbing motion. Accordingly, in this section we learn a \textit{personalized} alignment $z = f(u, c)$ that converts the human's joystick inputs $u$ to their preferred latent action $z$ (see Figure~\ref{fig:align_front}). Our goal is to make the robotic system easier to control: instead of forcing the human to adapt to $\phi$, we want the robot to adapt to the user's preferences (without fundamentally changing the latent action space or decoder $\phi$). 

To learn the human's preference we will query the user, showing them example robot motions and then asking them for the corresponding joystick input. But training our alignment model $f$ may require a large number of motion-joystick pairs, particularly in complex tasks where the user must leverage the same joystick input to accomplish several things. It is impractical to ask the human to provide all of these labels. Accordingly, to address the challenge of insufficient training data, we employ a \textit{semi-supervised} learning method \cite{chapelle2009semi}. In this section we first outline our approach, and then formulate a set of intuitive priors that facilitate semi-supervised learning from limited human feedback.

\begin{figure*}[t]
	\begin{center}
		\includegraphics[width=2.0\columnwidth]{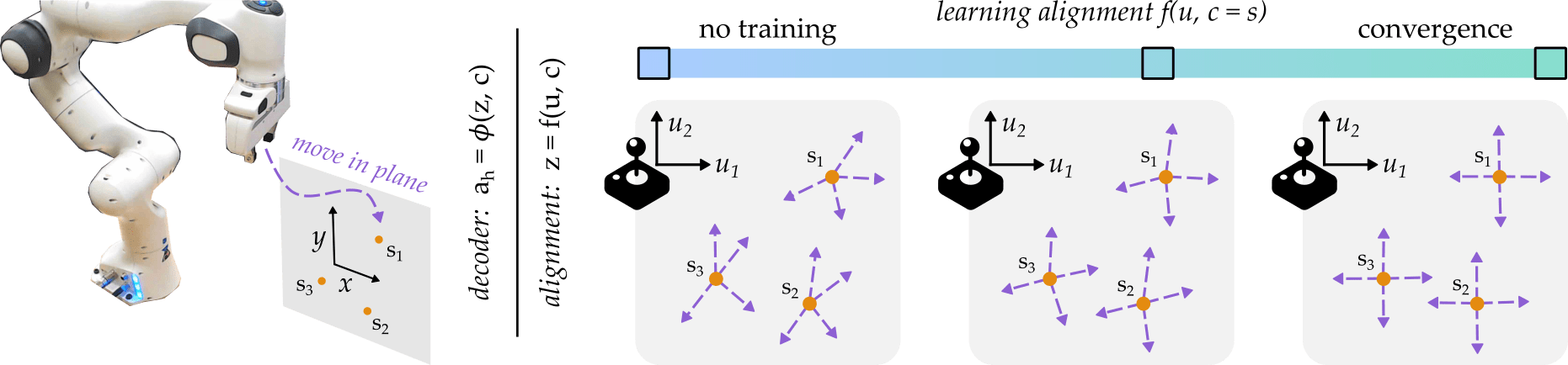}
		
		\caption{Training our alignment model $z = f(u, c)$. Here the context $c$ is equal to the robot's state $s$. (Left) the example task is to move the robot's end-effector in a 2D plane, and the current user prefers for the robot's end-effector motion to align with their joystick axes, so that $u_1$ moves the robot in the $x$-axis and $u_2$ moves the robot in the $y$-axis. (Right) we take snapshots at three different points during training, and plot how the robot actually moves when the human presses up, down, left, and right. Note that this alignment is context dependent. As training progresses, the robot learns the alignment $f$, and the robot’s motions are gradually and consistently pushed to match with the human’s individual preferences.}
		\label{fig:align_align}
	\end{center}

\end{figure*}

\subsection{Alignment Model}

Recall from Equation~(\ref{eq:P4}) that we seek to learn a function approximator $f : \mathcal{U} \times \mathcal{C} \rightarrow \mathcal{Z}$. Importantly, this alignment model is conditioned on the current context $c \in \mathcal{C}$. Consider the person in our motivating example, who is using a 2-axis joystick to control a high-DoF assistive robot arm to reach and cut tofu. The user's preferred way to control the robot is unclear: what does the user mean if they push the joystick right? When the robot is left of the tofu, the user might intend to move the robot towards the tofu --- but when the robot is directly above the tofu, pressing right now indicates that the robot should rotate and start a cutting motion! This mapping from the user input to intended action is not only person dependent, but it is also \textit{context dependent}. In practice, this context dependency prevents us from learning a single transformation to uniformly apply across the robot's workspace; instead, we need an intelligent strategy for understanding the human's preferences in different contexts.

\p{Model} To capture this interdependence we employ a general Multi-Layer Perceptron (MLP). The MLP $f$ takes in the current user input $u^t$ and context $c^t$, and outputs a latent action $z^t$. Combining $f$ with our latent action model, we now have a \textit{two-step} mapping between the human's low-dimensional input and the human's high-dimensional command:
\begin{equation} \label{eq:A1}
    a_h^t =  \phi(z^t, c^t) = \phi(f(u^t, c^t), c^t)
\end{equation}
When using our alignment model \textit{online}, we get the human's commanded action using Equation~(\ref{eq:A1}), and then combine this with shared autonomy to provide the overall robot action $a$. But \textit{offline} --- when we are learning $f$ --- we set $a = a_h$, so that the robot directly executes the human's commanded action. This disentangles the effects of shared autonomy and latent actions, and lets us focus on learning the preferred mapping from joystick inputs $u$ to robot actions $a$. Given that the robot takes action $a^t$ at the current timestep $t$, the state $s^{t+1}$ at the next timestep follows our transition model: $s^{t+1} = \mathcal{T}(s^t, a^t)$. Letting $a = a_h$, and plugging in Equation~(\ref{eq:A1}), we get the following relationship between joystick inputs and robot motion: 
\begin{equation} \label{eq:A2}
    s^{t+1} = \mathcal{T}(s^t, \phi(f(u^t, c^t), c^t)) = T(s^t, u^t)
\end{equation}
Our objective is to learn $f$ so that $s^{t+1} = T(s^t, u^t)$ matches the human's expectations. The overall training process is visualized in Figure~\ref{fig:align_align}.

\p{Loss Function} We train $f$ to minimize the loss function $\mathcal{L}_{\text{align}}$. We emphasize that this loss function (used for training the alignment model $f$) is different than the loss function described in \textbf{Sections~\ref{sec:latent-action}} and \textbf{\ref{sec:shared-autonomy}} (which was used for training the decoder $\phi$). Importantly, $\mathcal{L}_{\text{align}}$ must capture the individual user's expected joystick mapping --- and to understand what the user expects, we start by asking a set of questions. In each separate query, the robot starts in a state $s^t$ and moves to some state $s^*$. We then ask the user to \textit{label} this motion with their preferred joystick input $u$, resulting in the labeled data tuple $(s^t, u^t, s^*)$. For instance, the robot arm starts above the tofu, and then stabs down to break off a piece: you might label this motion by holding down on the joystick (i.e., $u = \text{down}$).

Given a start state $s^t$ and input $u^t$ from the user's labeled data, the robot learns $f$ to minimize the distance between $T(s^t, u^t)$ and $s^*$. Letting $N$ denote the number of queries that the human has answered, and letting $d$ be the distance metric, our alignment function should minimize:
\begin{equation} \label{eq:A3}
    L_{\text{sup}} = \frac{1}{N}\sum_{i=1}^{N} d(s^{*,i}, T(s^i, u^i))
\end{equation}
If we could ask the human as many questions as necessary, then $\mathcal{L}_{\text{align}} = \mathcal{L}_{\text{sup}}$, and our alignment function only needs to minimize the supervised loss. But collecting this large dataset is impractical. Accordingly, to minimize the number of questions the human must answer, we introduce additional loss terms in $\mathcal{L}_{\text{align}}$ that capture underlying priors in human expectations.

\subsection{Reducing Human Data with Intuitive Priors}

Our insight here is that humans share some underlying expectations of how the control mapping should behave \cite{jonschkowski2014state}. We will formulate these common expectations --- i.e., \textit{priors} --- as loss terms that $f$ minimizes within semi-supervised learning.

When introducing these priors, it helps to refine our notation. Recall that $s$ is the system state: here we use $s$ to specifically refer to the robot's joint position, and we denote the forward kinematics of the robot arm as $x = \Psi(s)$. End-effector pose $x$ is particularly important, since humans often focus on the robot's gripper during eating tasks. When the human applies joystick input $u$ at state $s$, the corresponding change in end-effector pose $x$ is: $\Delta x =  \Psi(T(u, s)) - \Psi(s)$. With these definitions in mind, we argue an intuitive controller should satisfy the properties listed below. We emphasize that --- although these properties share some common themes with the latent action properties from \textbf{Section~\ref{sec:latent-action}} --- the purpose of these properties is different. When formalizing the properties for latent actions we focused on enabling the human to complete tasks using these latent actions. By contrast, here we focus on intuitive and task-agnostic expectations for controller mappings.

\p{Proportionality} The amount of change in the position and orientation of the robot's end-effector should be proportional to the scale of the human's input. In other words, for scalar $\alpha$, we expect:
\begin{equation*}
        \alpha \cdot | \Psi(T(u, s)) - \Psi(s) | = | \Psi(T(\alpha \cdot u, s)) - \Psi(s) |
\end{equation*}
We accordingly define the proportionality loss $\mathcal{L}_{\text{prop}}$ as:
\begin{equation} \label{eq:A4}
    \mathcal{L}_{\text{prop}} = \big\| \Psi(T(\alpha \cdot u, s)) - \Psi(s) - \alpha \cdot \Delta x\big\|^2
\end{equation}
where $\alpha$ is sampled from our range of joystick inputs.

\p{Reversability} If a joystick input $u$ makes the robot move \textit{forward} from $s_1$ to $s_2$, then the opposite input $(-u)$ should move the robot \textit{back} from $s_2$ to its original end-effector position. In other words, we expect:
\begin{equation*}
    \Psi(s_1) = \Psi\Big(T\big(-u, T(u, s_1)\big)\Big)
\end{equation*}
This property ensures users can recover from their mistakes. We define the reversability loss $\mathcal{L}_{\text{reverse}}$ as:
\begin{equation} \label{eq:A5}
    \mathcal{L}_{\text{reverse}} = \big\|\Psi(s) - \Psi\big(T(-u, T(u, s))\big)\big\|^2
\end{equation}
Here $\Psi(s)$ is the current position and orientation of the robot's end-effector, and the right term is the pose of the end-effector after executing human input $u$ followed by the opposite input $(-u)$.

\p{Consistency} The same input taken at nearby states should lead to similar changes in robot pose. We previously discussed a similar property in \textbf{Section~\ref{sec:latent-action}} when formalizing latent actions. Here we specifically focus on the input-output relationship between joystick input $u$ and end-effector position $x = \Psi(s)$:
\begin{gather*}
    \Delta x_1 = \| \Psi(T(u, s_1)) - \Psi(s_1) \|^2 \\ \Delta x_2 = \| \Psi(T(u, s_2)) - \Psi(s_2) \|^2
\end{gather*}
We expect $\| \Delta x_1 - \Delta x_2 \| \rightarrow 0$ as $\| s_1 - s_2 \| \rightarrow 0$. Consistency prevents sudden changes in the alignment mapping. We define the consistency loss $\mathcal{L}_{\text{con}}$ as:
\begin{equation} \label{eq:A6}
    \mathcal{L}_{\text{con}} = \|\Delta x(s_1) - \Delta x(s_2)\| \cdot \exp\big\{-\gamma\| s_1 - s_2 \|\big\}
\end{equation}
When the hyperparameter $\gamma \rightarrow 0$, the robot only enforces consistency at local states, and when $\gamma \rightarrow \infty$, the robot tries to enforce consistency at all states.

\p{Semi-Supervised Learning} When learning our alignment model $f$, we first collect a batch of robot motions $(s, s^*)$. The human labels $N$ of these (start-state, end-state) pairs with their preferred joystick input $u$, so that we have labeled data $(s, u, s^*)$. We then train the alignment model to minimize the supervised loss for the \textit{labeled} data, as well as the semi-supervised loss for the \textit{unlabeled} data. Hence, the cumulative loss function is:
\begin{equation}
    \label{eqn:all_loss}
    \mathcal{L}_{\text{align}} = \mathcal{L}_{\text{sup}} + \lambda_1 \mathcal{L}_{\text{prop}} + \lambda_2\mathcal{L}_{\text{reverse}} + \lambda_3 \mathcal{L}_{\text{con}}
\end{equation}
Importantly, incorporating these different loss terms --- which are inspired by human priors over controllable spaces \cite{jonschkowski2014state} --- enables the robot to generalize the labeled human data (which it performs supervised learning on) to unlabeled states (which it can now perform semi-supervised learning on).
\begin{algorithm}[t]
\caption{Latent Control of Assistive Robots}
  \label{deltaco}
\begin{algorithmic}[1]

    \Statex \textbf{Offline:}
    \State Select a discrete set of goals $\mathcal{G}$
    \State Collect a dataset $\mathcal{D} = \{(c_0,a_0), (c_1,a_1), \ldots \}$ from kinesthetic demonstrations
    \State Train latent action model $\phi$ to minimize $\mathcal{L}$ on $\mathcal{D}$    
    \State Query the user to label example motions $\{(s, s^*)\}$ with their preferred joystick direction $u$
    \State Train alignment model $f$ to minimize loss $\mathcal{L}_{align}$ using labeled and unlabeled motions $\{(s, s^*)\}$
    
    \Statex \textbf{Online, at each timestep $t$:}

        \State $z^t \gets f(u^t, c^t)$ \Comment Align latent action with user input
        \State $a_h^t \gets \phi(z^t, c^t)$ \Comment Decode $z^t$ to high-DoF action
        \State $a_r^t \gets \sum_{g \in \mathcal{G}} b^t(g) \cdot (g - s^t)$  \Comment{get robot assistance}
        \State $a^t \gets (1 - \alpha) \cdot a_h^t + \alpha \cdot a_r^t$ \Comment{blend both $a_h$ and $a_r$}
        \State $b^{t+1} \propto P(u^t ~|~ c^t, g)P(g)$ \Comment{update belief over goals}
        \State $s^{t+1} \sim \mathcal{T}(s^t, a^t)$ \Comment{take action}
\end{algorithmic}
\end{algorithm}

\section{Algorithm} \label{sec:alg}

\textbf{Sections~\ref{sec:latent-action}}, \textbf{\ref{sec:shared-autonomy}}, and \textbf{\ref{sec:alignment}} developed parts of our approach. Here we put these pieces together to present our general algorithm for controlling assistive eating robots with learned latent actions. Our approach is summarized in Algorithm~\ref{deltaco} and explained below. 

Given an assistive eating scenario, we start by identifying the food items and other potential goals that the human may want to reach (Line 1). We then collect high-dimensional kinesthetic demonstrations, where a caretaker backdrives the robot through task-related motions that interact with these potential goals (Line 2). Leveraging the properties and models from \textbf{Sections~\ref{sec:latent-action}}, we then train our latent action space and learn decoder $\phi$ (Line 3). Next, we show the user example robot motions --- e.g., by sampling values of $z$ --- and ask the user to label these motions with their preferred joystick input (Line 4). Applying the priors developed in \textbf{Sections~\ref{sec:alignment}}, we generalize from a small number of human labels to learn the alignment $f$ between joystick inputs and latent actions (Line 5).

Once we have learned $\phi$ and $f$, we are ready for the human-in-the-loop. At each timestep the human presses their low-DoF joystick to provide input $u$. We find the latent action $z$ that is aligned with the human's input (Line 6), and then decode that low-DoF latent action to get a high-DoF robot action $a_h$ (Line 7). In order to help the human reach and maintain their high-level goals, we incorporate the shared autonomy approach from \textbf{Sections~\ref{sec:shared-autonomy}}. Shared autonomy selects an assistive action $a_r$ based on the current belief over the human's goal (Line 8), and the robot blends $a_r$ and $a_h$ to take overall action $a$ (Line 9). Finally, the robot applies Bayesian inference to update its understanding of the human's desired goal based on their joystick input (Line 10). We repeat this process until the human has finished eating.

\medskip

\noindent\textbf{How Practical is Our Approach?} One concern is the amount of data required to learn latent actions. In all of the studies reported below --- where the assistive robot makes a mock-up apple pie, assembles dessert, and cuts tofu --- the robot was trained with a maximum of \textbf{twenty minutes} of kinesthetic demonstrations, and all training was done \textbf{on-board} the robot computer. We recognize that this short training time is likely due to the structure of the cVAE model used in these tasks and may not hold true in general; however, this easy implementation holds promise for future use. In the following sections, we demonstrate the objective and subjective benefits of Algorithm~\ref{deltaco}, as well as highlighting some of its shortcomings.

\medskip

\noindent\textbf{Where Do the Goals Come From?} Another question is how the robot detects the discrete set of goals $\mathcal{G}$ that the human may want to reach. Here we turn to \textit{perception}, where recent assistive eating work shows how robot arms can estimate the pose of various objects of interest \cite{feng2019robot,park2019toward}. Determining which objects are potential goals is simplified in eating settings, since the target items are largely consistent (e.g., food items, cups, plates, and bowls). The location of these goals is included in the state $s$ and the robot uses this information when decoding the human's joystick inputs. Although not covered in this paper, it is also possible to condition latent actions directly on the robot's perception, so that $s$ becomes the visual inputs \cite{karamcheti2021learning}.
\section{Simulations}

We performed three separate simulations, one for each key aspect of our proposed method. First we leverage different autoencoder models to learn latent actions, and determine which types of models best capture the user-friendly properties formalized in \textbf{Section~\ref{sec:latent-action}}. Our second simulation then compares learned latent actions alone to latent actions with shared autonomy (\textbf{Section~\ref{sec:shared-autonomy}}), and focuses on how shared autonomy helps users reach, maintain, and change their high-level goals. Finally, we learn the alignment model from \textbf{Section~\ref{sec:alignment}} between joystick inputs and latent actions. We compare versions of our semi-supervised approach with intuitive priors, and see how these priors improve the alignment when we only have access to limited and imperfect human feedback. All three simulations were performed in controlled conditions with simulated humans and simulated or real robot arms. These simulated humans chose joystick inputs according to mathematical models of human decision making, as detailed below.

\subsection{Do Learned Latent Actions Capture our User-Friendly Properties?} \label{sim:icra}

Here we explore how well our proposed models for learning latent actions capture the user-friendly properties formalized in \textbf{Section~\ref{sec:latent-action}}. These properties include \textit{controllability}, \textit{consistency}, and \textit{scalability}.

\p{Setup} We simulate one-arm and two-arm planar robots, where each arm has $n=5$ degrees-of-freedom. The state $s \in \mathbb{R}^{n}$ is the robot's joint position, and the action $a \in \mathbb{R}^{n}$ is the robot's joint velocity. Hence, the robot transitions according to: $s^{t+1} = s^t + a^t \cdot dt$, where $dt$ is the step size. Demonstrations consist of trajectories of state-action pairs: in each of different simulated tasks, the robot trains with a total of $10000$ state-action pairs.

\p{Tasks} The simulated robots perform four different tasks.
\begin{enumerate}
    \item \textit{Sine}: one 5-DoF robot arm moves its end-effector along a sine wave with a 1-DoF latent action
    \item \textit{Rotate}: two 5-DoF robot arms are holding a box, and rotate that box about a fixed point using a 1-DoF latent action
    \item \textit{Circle}: one 5-DoF robot moves back and forth along circles of different radii with a 2-DoF latent action
    \item \textit{Reach}: one 5-DoF robot arm reaches from a start location to a goal region with a 1-DoF latent action
\end{enumerate}

\p{Model Details} We test latent action models which minimize the different loss function described in \textbf{Section~\ref{sec:latent-action}}. Specifically, we test:
\begin{itemize}
    \item Principal Component Analysis (\textbf{PCA})
    \item Autoencoders (\textbf{AE})
    \item Variational autoencoders (\textbf{VAE})
    \item Conditioned autoencoders (\textbf{cAE})
    \item Conditioned variational autoencoders (\textbf{cVAE})
\end{itemize}
The encoders and decoders contain between two and four linear layers (depending on the task). The loss function is optimized using Adam with a learning rate of $1e^{-2}$. Within the VAE and cVAE, we set the normalization weight $<1$ to avoid posterior collapse.

\p{Dependent Measures} To determine \textit{accuracy}, we measure the mean-squared error between the intended actions $a$ and reconstructed actions $\hat{a}$ on a test set of state-action pairs $(s,a)$ drawn from the same distribution as the training set. 

To test model \textit{controllability}, we select pairs of start and goal states $(s_i,s_j)$ from the test set, and solve for the latent actions $z$ that minimize the error between the robot's current state and $s_j$. We then report this minimum state error.

We jointly measure \textit{consistency} and \textit{scalability}: to do this, we select $25$ states along the task, and apply a fixed grid of latent actions $z_i$ from $[-1, +1]$ at each state. For every $(s,z)$ pair we record the distance and direction that the end-effector travels (e.g., the direction is $+1$ if the end-effector moves right). We then find the best-fit line relating $z$ to distance times direction, and report its $R^2$ error.

Our results are averaged across $10$ trained models of the same type, and are listed in the form $mean \pm SD$.

\p{Hypotheses} We have the following two hypotheses:
\begin{displayquote}
\textbf{H1.} \textit{Only latent action models conditioned on the context will accurately reconstruct actions from low-DoF inputs.}
\end{displayquote}
\begin{displayquote}
\textbf{H2.} \textit{Conditioned autoencoders and conditioned variational autoencoders will learn a latent space that is controllable, consistent, and scalable.}
\end{displayquote}

\begin{figure}[t]

	\begin{center}
		\includegraphics[width=1.0\columnwidth]{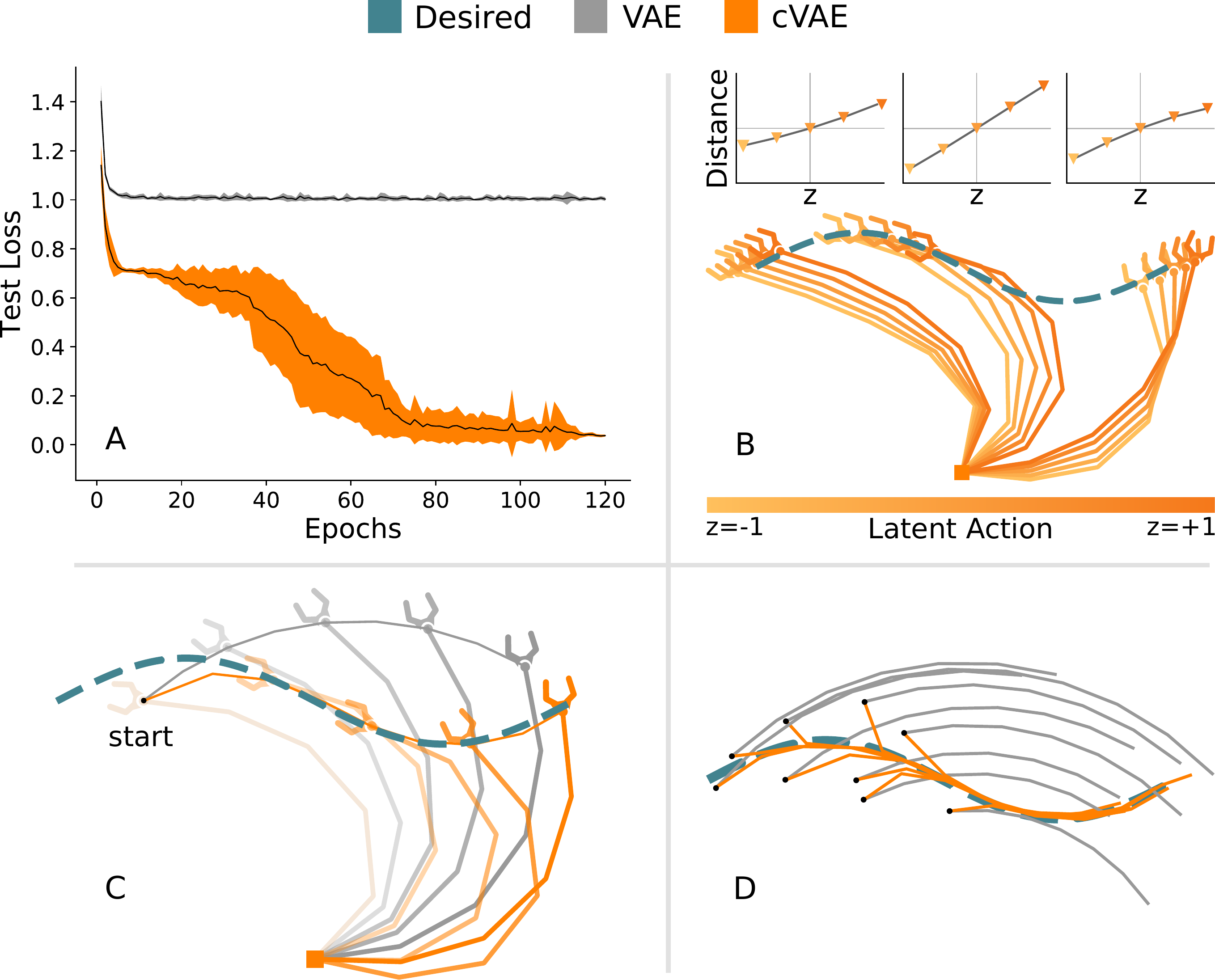}

		\caption{Results for the \textit{Sine} task. (A) mean-squared error between intended and reconstructed actions normalized by PCA test loss. (B) effect of the latent action $z$ at three states along the sine wave for the cVAE model. Darker colors correspond to $z>0$ and lighter colors signify $z<0$. Above we plot the distance that the end effector moves along the sine wave as a function of $z$ at each state. (C) rollout of robot behavior when applying a constant latent input $z=+1$, where both VAE and cVAE start at the same state. (D) end-effector trajectories for multiple rollouts of VAE and cVAE.}

		\label{fig:sine}
	\end{center}
	
	\vspace{-1em}
	
\end{figure}

\p{Sine Task} This task and our results are shown in Figure~\ref{fig:sine}. We find that conditioning the decoder on the current context, i.e., $\phi(z,c)$, greatly improves \textit{accuracy} when compared to the PCA baseline, i.e., $\phi(z)$. Here AE and VAE incur $98.0\pm 0.6\%$ and $100\pm0.8\%$ of the PCA loss, while cAE and cVAE obtain $1.37\pm1.2\%$ and $3.74\pm0.4\%$ of the PCA loss, respectively.

We likewise observe that cAE and cVAE are more \textit{controllable} than their alternatives. When using the learned latent actions to move between $1000$ randomly selected start and end states along the sine wave, cAE and cVAE have an average end-effector error of $0.05\pm0.01$ and $0.10\pm0.01$. Models without state conditioning---PCA, AE, and VAE---have average errors $0.90$, $0.94\pm0.01$, and $0.95\pm0.01$.

When evaluating \textit{consistency} and \textit{scalability}, every tested model has a roughly linear relationship between latent actions and robot behavior: PCA has the highest $R^2 = 0.99$, while cAE and cVAE have the lowest $R^2 = 0.94 \pm 0.04$ and $R^2 = 0.95 \pm 0.01$.

\begin{figure}[t]

	\begin{center}
		\includegraphics[width=1.0\columnwidth]{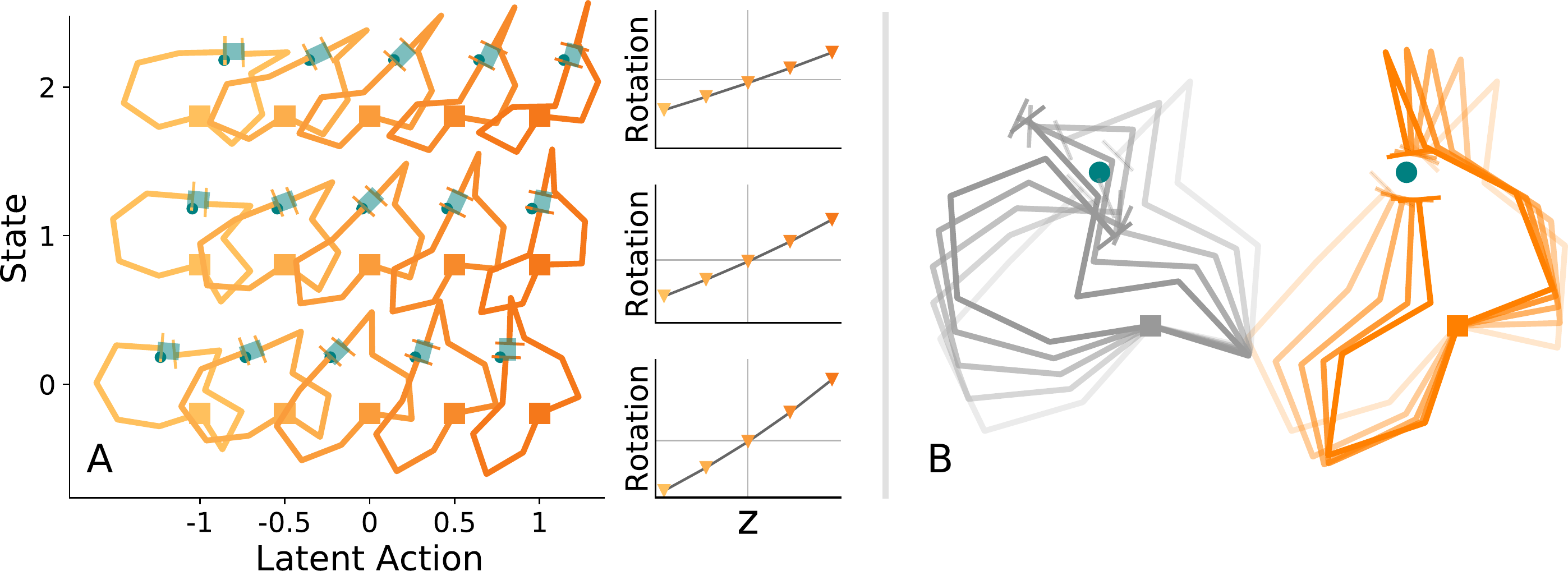}

		\caption{Results for the \textit{Rotate} task. (A) the robot uses two arms to hold a light blue box, and learns to rotate this box around the fixed point shown in teal. Each state corresponds to a different fixed point, and positive $z$ causes counterclockwise rotation. On right we show how $z$ affects the rotation of the box at each state. (B) rollout of the robot's trajectory when the user applies $z=+1$ for VAE and cVAE models, where both models start in the same state. Unlike the VAE, the cVAE model coordinates its two arms.}

		\label{fig:rotate}
	\end{center}
	
\end{figure}

\p{Rotate Task} We summarize the results for this two-arm task in Fig.~\ref{fig:rotate}. Like in the \textit{Sine} task, the models conditioned on the current context are more \textit{accurate} than their non-conditioned counterparts: AE and VAE have $28.7\pm4.8\%$ and $38.0\pm5.8\%$ of the PCA baseline loss, while cAE and cVAE reduce this to $0.65\pm0.05\%$ and $0.84\pm0.07\%$. The context conditioned models are also more \textit{controllable}: when using the learned $z$ to rotate the box, AE and VAE have $56.8\pm9\%$ and $71.5\pm8\%$ as much end-effector error as the PCA baseline, whereas cAE and cVAE achieve $5.4\pm0.1\%$ and $5.9\pm0.1\%$ error.

When testing for \textit{consistency} and \textit{scalability}, we measure the relationship between the latent action $z$ and the change in orientation for the end-effectors of both arms (i.e., ignoring their location). Each model exhibits a linear relationship between $z$ and orientation: $R^2 = 0.995\pm0.004$ for cVAE and $R^2 = 0.996\pm0.002$ for cVAE. In other words, there is an approximately linear mapping between $z$ and the orientation of the box that the two arms are holding.

\begin{figure}[t]

	\begin{center}
		\includegraphics[width=1.0\columnwidth]{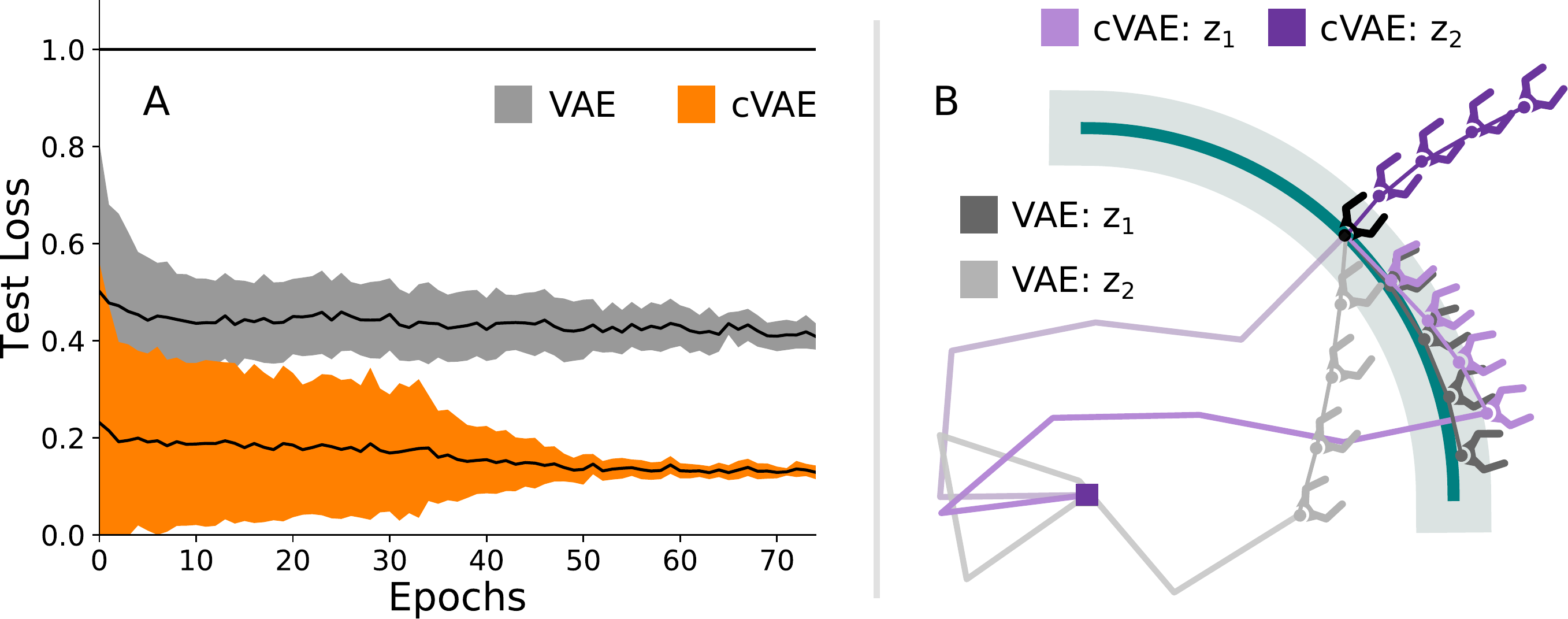}

		\caption{Results for the \textit{Circle} task. (A) mean-squared error between desired and reconstructed actions normalized by the PCA test loss. (B) 2-DoF latent action space $z = [z_1, z_2]$ for VAE and cVAE models. The current end-effector position is shown in black, and the colored grippers depict how changing $z_1$ or $z_2$ affects the robot's state. Under the cVAE model, these latent dimensions move the end-effector tangent or orthogonal to the circle.}

		\label{fig:circle}
	\end{center}
	
\end{figure}

\p{Circle Task} Next, consider the one-arm task in Fig.~\ref{fig:circle} where the robot has a 2-DoF latent action space. We here focus on the learned latent dimensions $z = [z_1,z_2]$, and examine how these latent dimensions correspond to the underlying task. Recall that the training data consists of state-action pairs which translate the robot's end-effector along (and between) circles of different radii. Ideally, the learned latent dimensions correspond to these axes, e.g., $z_1$ controls tangential motion while $z_2$ controls orthogonal motion. Interestingly, we found that this intuitive mapping is \textit{only} captured by the state conditioned models. The average angle between the directions that the end-effector moves for $z_1$ and $z_2$ is $27\pm 20^{\circ}$ and $34 \pm 15^{\circ}$ for AE and VAE models, but this angle increases to $72 \pm 9^{\circ}$ and $74 \pm 12^{\circ}$ for the cAE and cVAE (ideally $90^{\circ}$). The state conditioned models better \textit{disentangle} their low-dimensional embeddings, supporting our hypotheses and demonstrating how these models produce user-friendly latent spaces.

\begin{figure}[t]

	\begin{center}
		\includegraphics[width=1.0\columnwidth]{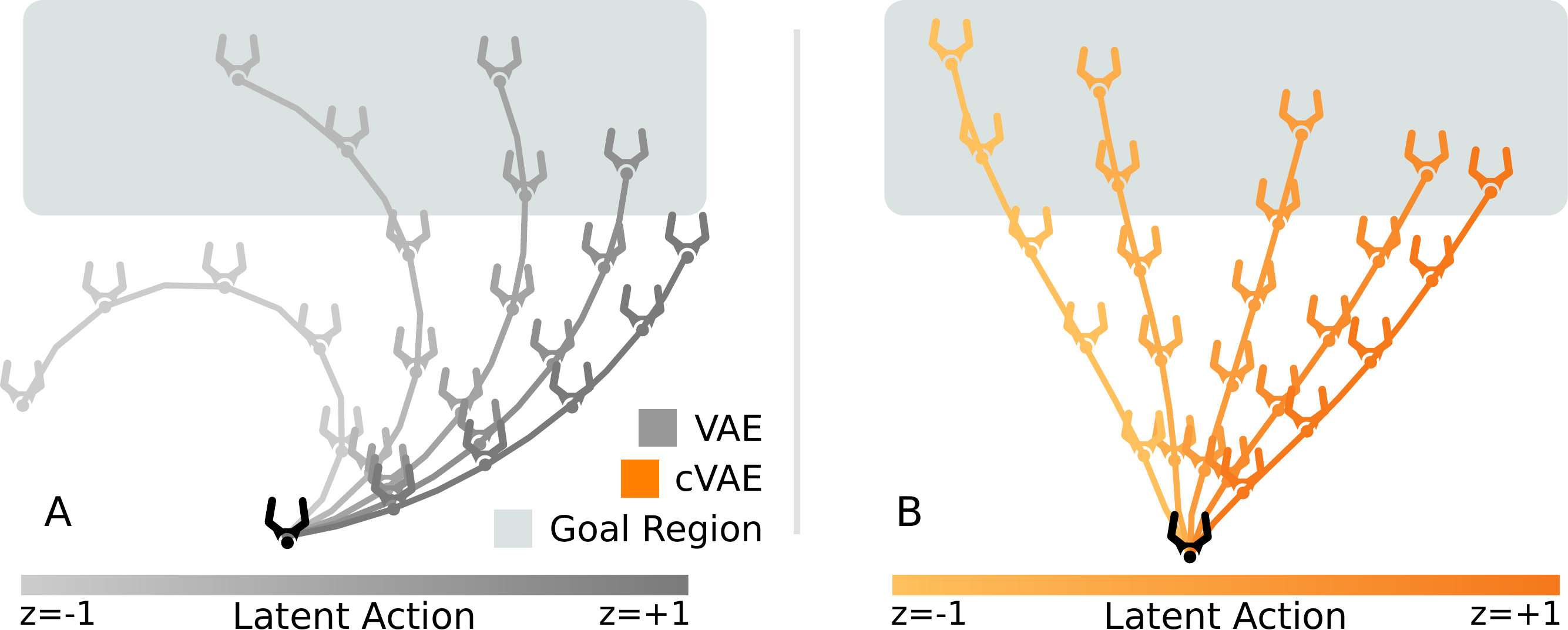}

		\caption{Results for the \textit{Reach} task. In both plots, we show the end-effector trajectory when applying constant inputs $z \in [-1,+1]$. The lightest color corresponds to $z=-1$ and the darkest color is $z=+1$. The goal region is highlighted, and the initial end-effector position is black. (A) trajectories with the VAE model. (B) trajectories with the cVAE model. The latent action $z$ controls which part of the goal region the trajectory moves towards.}

		\label{fig:reach}
	\end{center}
	
	\vspace{-1em}

\end{figure}

\p{Reach Task} In the final task, a one-arm robot trains on trajectories that move towards a goal region (see Fig.~\ref{fig:reach}). The robot learns a 1-DoF latent space, where $z$ controls the direction that the trajectory moves (i.e., to the left or right of the goal region). We focus on \textit{controllability}: can robots utilize latent actions to reach their desired goal? In order to test controllability, we sample $100$ goals randomly from the goal region, and compare robots that attempt to reach these goals with either VAE or cVAE latent spaces. The cVAE robot more accurately reaches its goal: the $L_2$ distance between the goal and the robot's final end-effector position is $0.57\pm0.38$ under VAE and $0.48\pm0.5$ with cVAE. Importantly, using conditioning also improves the movement \textit{quality}. The average start-to-goal trajectory is $5.1\pm2.8$ units when using the VAE, and this length drops to $3.1\pm0.5$ with the cVAE model.

\p{Summary} The results of our \textit{Sine}, \textit{Rotate}, \textit{Circle}, and \textit{Reach} tasks support hypotheses \textbf{H1} and \textbf{H2}. Latent action models that are conditioned on the context more \textit{accurately} reconstruct high-DoF actions from low-DoF embeddings (\textbf{H1}). Moreover, conditioned autoencoders and conditioned variational autoencoders learn latent action spaces which capture our desired properties: \textit{controllability}, \textit{consistency}, and \textit{scalability} (\textbf{H2}).

\subsection{Do Latent Actions with Shared Autonomy Help Users Reach and Change Goals?}

Now that we have tested our method for learning latent actions, the next step is to combine these latent actions with shared autonomy (see \textbf{Section~\ref{sec:shared-autonomy}}). Here we explore how this approach works with a \textit{spectrum of different simulated users}. We simulate human teleoperators with various levels of expertise and adaptability, and measure whether these users can interact with our algorithm to \textit{reach} and \textit{change} high-level goals.

\p{Incorporating Shared Autonomy} In the previous simulations we used latent actions by themselves to control the robot. Now we compare this approach with and without shared autonomy:
\begin{itemize}
    \item Latent actions with no assistance (\textbf{LA})
    \item Latent actions with shared autonomy (\textbf{LA+SA})
    \item Latent actions trained to maximize entropy with shared autonomy (\textbf{LA+SA+Entropy})
\end{itemize}
For both LA and LA+SA we learn the latent space with a conditioned autoencoder (i.e., cAE in the previous section). However, here the context includes both \textit{state and belief}. In other words, $c = (s,b)$. We also test LA+SA+Entropy, where the model uses Equation~(\ref{eq:T6}) to reward entropy in the learned latent space.

\begin{figure}[t]
	\begin{center}
		\includegraphics[width=1\columnwidth]{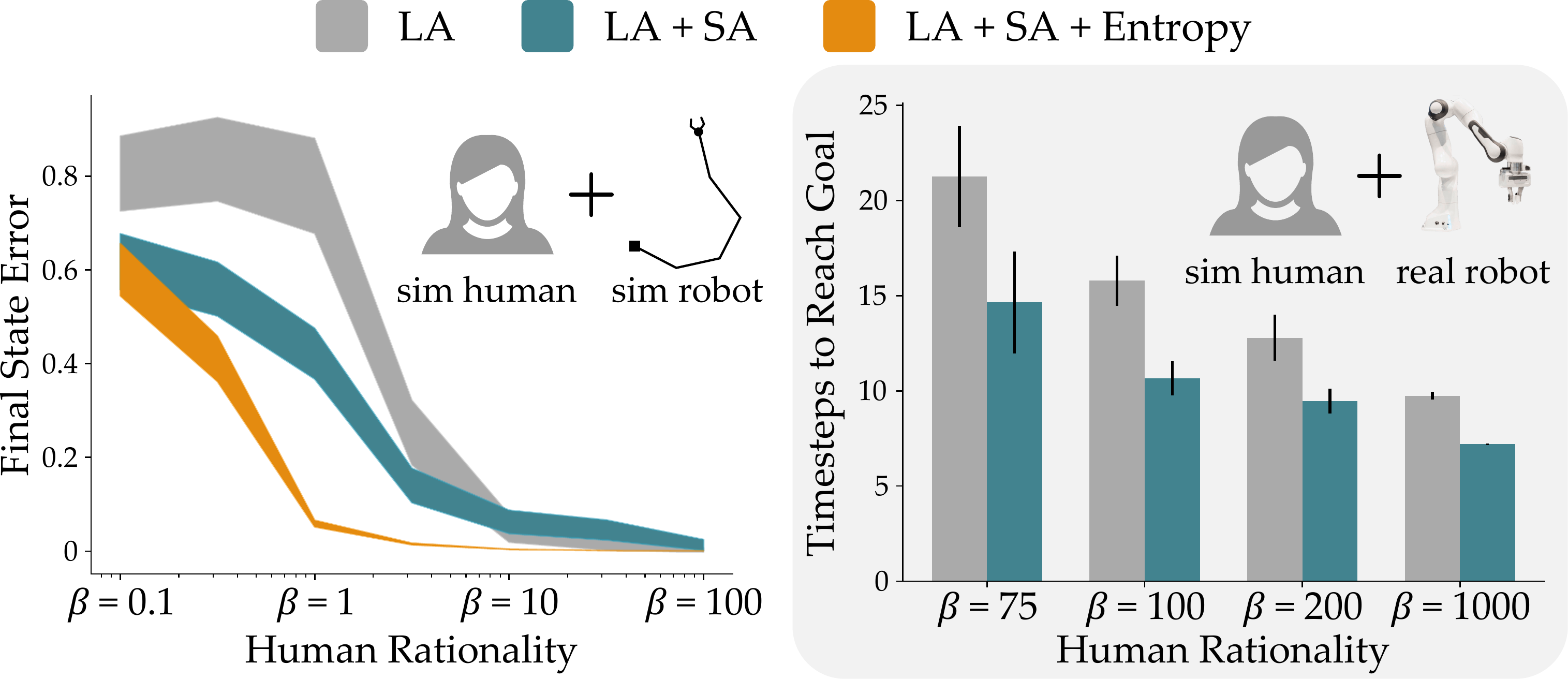}
		
		\caption{Simulated humans for different levels of rationality. As ${\beta \rightarrow \infty}$, the human's choices approach optimal inputs. \textit{Final State Error} (in all plots) is normalized by the distance between goals. Introducing shared autonomy (\textit{SA}) improves the convergence of latent actions (\textit{LA}), particularly when the human teleoperator is noisy and imperfect.}
		\label{fig:sim2a}
	\end{center}

\end{figure}

\begin{figure*}[t]
	\begin{center}
		\includegraphics[width=1.8\columnwidth]{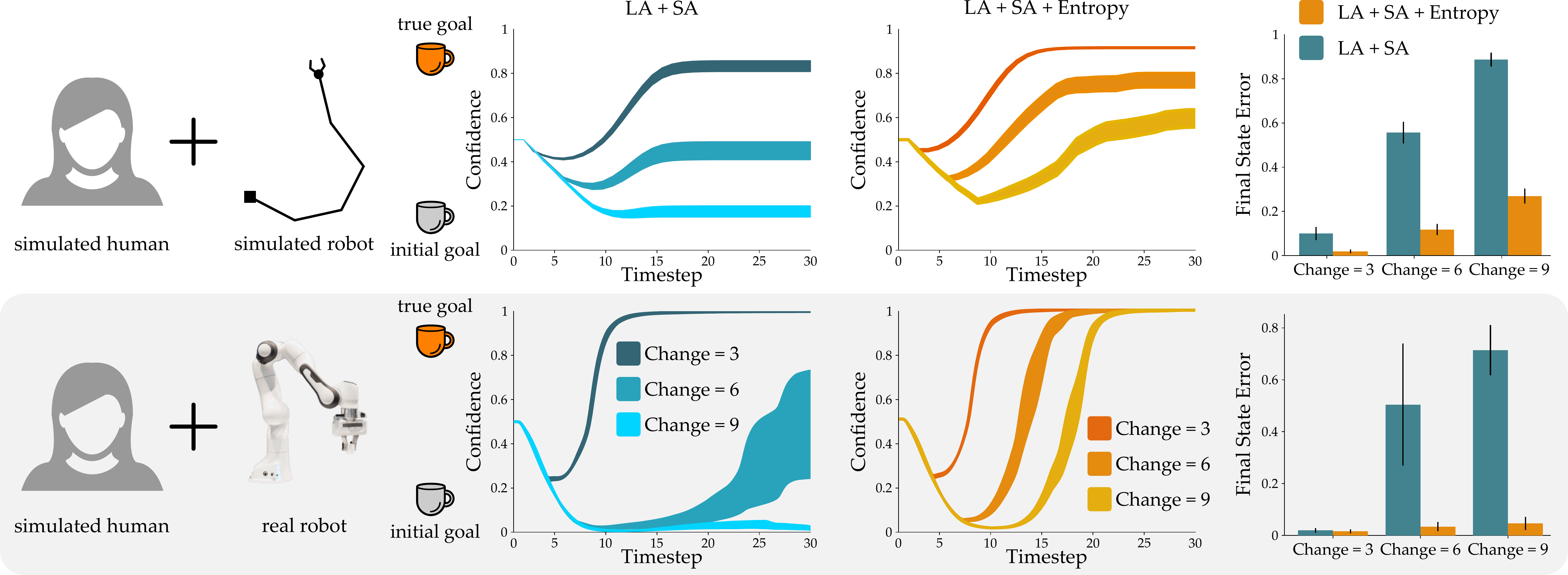}

		\caption{Simulated humans that change their intended goal part-way through the task. \textit{Change} is the timestep where this change occurs, and \textit{Confidence} refers to the robot's belief in the human's true goal. Because of the constraints imposed by shared autonomy, users need latent actions that can overcome misguided assistance and move towards a less likely (but correct) goal. Encouraging entropy in the learned latent space (\textit{LA+SA+Entropy}) enables users to switch goals.}
		\label{fig:sim2b}
	\end{center}

\end{figure*}

\p{Environments} We implement these models on both a \textit{simulated} and a \textit{real} robot. The simulated robot is a $5$-DoF planar arm, and the real robot is a $7$-DoF Franka Emika. For both robots, the state $s$ captures the current joint position, and the action $a$ is a change in joint position, so that: $s^{t+1} = s^t + a^t \cdot dt$.

\p{Task} We consider a manipulation task where there are two coffee cups in front of a robot arm (see Figure~\ref{fig:sim2b}). The human may want to reach and grasp either cup (i.e., these cups are the potential goals). We embed the robot's high-DoF actions into a $1$-DoF input space: the simulated users had to convey both their goal and preference \textit{only by pressing left and right} on the joystick.

\p{Simulated Humans} The users attempting to complete this task are \textit{approximately optimal}, and make decisions that guide the robot accordingly to their goal $g^*$. Remember that $x$ is the position of the robot's end-effector and $\Psi$ is the forward kinematics. The humans have reward function $R = -\|g^* - x\|^2$, and choose latent actions $z$ to move the robot towards $g^*$:
\begin{equation} \label{eq:S1}
	p(z) \propto \exp{ \Big\{ - \beta(t) \cdot \| g^* - \Psi(s + \phi(z, c) \cdot dt) \|^2 \Big\}}
\end{equation}
Here $\beta \geq 0$ is a temperature constant that affects the user's \textit{rationality}. When $\beta \rightarrow 0$, the human selects increasingly random $z$, and when $\beta \rightarrow \infty$, the human always chooses the $z$ that moves towards $g^*$. We simulate different types of users by varying $\beta(t)$.

\p{Users with Fixed Expertise} We first simulate humans that have \textit{fixed} levels of expertise. Here expertise is captured by $\beta$ from Equation~(\ref{eq:S1}): users with high $\beta$ are proficient, and rarely make mistakes with noisy inputs. We anticipate that all algorithms will perform similarly when humans are always perfect or completely random---but we are particularly interested in the spectrum of users \textit{between} these extremes, who frequently \textit{mis-control} the robot.

Our results relating $\beta$ to performance are shown in Figure~\ref{fig:sim2a}. In accordance with our convergence result from \textbf{Section~\ref{sec:shared-autonomy}}, we find that introducing shared autonomy helps humans reach their desired grasp more quickly, and with less final state error. The performance difference between LA and LA+SA decreases as the human's expertise increases --- looking specifically at the real robot simulations, LA takes $45\%$ more time to complete the task than LA+SA at $\beta = 75$, but only $30\%$ more time when $\beta = 1000$. We conclude that shared autonomy improves performance across all levels of expertise, both when latent actions are trained with and without entropy.

\p{Users that Change their Mind} One downside of shared autonomy is over-assistance: the robot may become \textit{constrained} at likely (but incorrect) goals. To examine this adverse scenario we simulate humans that \textit{change} which coffee cup they want to grasp after $N$ timesteps. These simulated users intentionally move towards the \textit{wrong} cup while $t \leq N$, and then try to reach the correct cup for the rest of the task.
We model humans as near-optimal immediately after changing their mind about the goal. 

We visualize our results in Figure~\ref{fig:sim2b}. When the latent action space is trained only to minimize reconstruction loss (LA+SA), users cannot escape the shared autonomy constraint around the wrong goal as $N$ increases. Intuitively, this occurs because the latent space controls the intended goal when the belief $b$ is roughly uniform, and then switches to controlling the preferred trajectory once the robot is confident. So if users change their goal after first convincing the robot, the latent space no longer contains actions that move towards this correct goal! We find that our proposed entropy loss function addresses this shortcoming: LA+SA+Entropy users are able to input actions $z$ that alter the robot's goal. Our results suggest that encouraging entropy at training time improves the robustness of the latent space.

\begin{figure*}[t]
	\begin{center}
		\includegraphics[width=1.8\columnwidth]{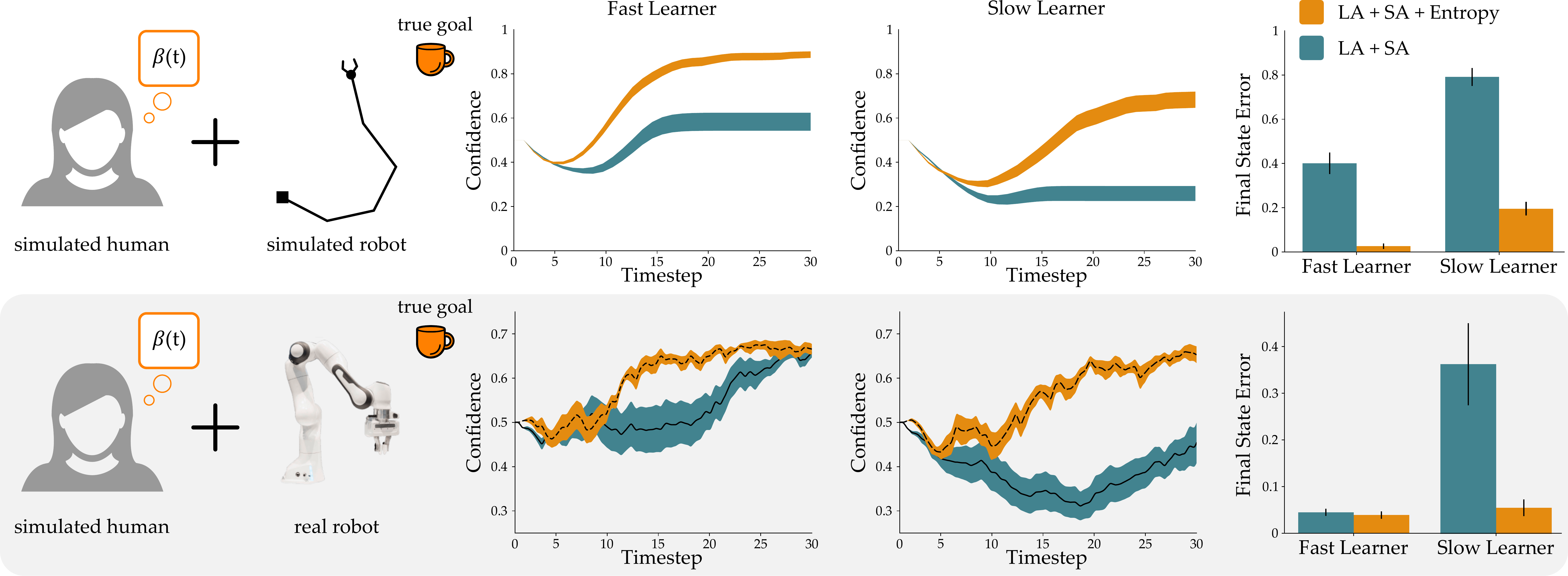}
		
		\caption{Simulated humans that learn how to teleoperate the robot. The human's rationality $\beta(t)$ is linear in time, and either increases with a high slope (\textit{Fast Learner}) or low slope (\textit{Slow Learner}). As the human learns, they get better at choosing inputs that best guide the robot towards their true goal. We find that latent actions learned with the entropy reward (\textit{LA+SA+Entropy}) are more versatile, so that the human can quickly undo mistakes made while learning.}
		\label{fig:sim2c}
	\end{center}
	
	\vspace{-2em}

\end{figure*}

\p{Users that Learn within the Task} We not only expect real users to change their mind when collaborating with the robot, but we also anticipate that these teleoperators will \textit{learn and improve} as they gain experience during the task. For instance, the user might learn that holding left on the joystick causes the robot to grasp the cup from the side, while holding right guides the robot towards a top grasp. To simulate this in-task learning, we set $\beta(t) = m\cdot t$, where the slope $m$ determines how quickly the user learns. All users start with random actions ($\beta = 0$), and either learn \textit{quickly} (high $m$) or \textit{slowly} (low $m$). We point out that slow learners may effectively ``change their mind'' multiple times, since they are unsure of how to control the robot.

Our findings are plotted in Figure~\ref{fig:sim2c}. We see that --- for both fast and slow learners --- LA+SA+Entropy improves in-task performance. We attribute this improvement to the inherent \textit{versatility} of latent spaces that maximize entropy: as humans gain expertise, they can use these latent actions to quickly undo their mistakes and correct the robot's behavior.

\p{Summary} Overall, these simulations show that users are not able to precisely reach and maintain goals when controlling robots with \textit{only} latent actions. Including shared autonomy improves convergence, while training latent actions to maximize entropy ensures that this convergence does not become a burden. When using our proposed combination of shared autonomy and learned latent actions, na\"ive and experenced users are able to reach their preferred goal and change their mind.

\subsection{Can We Efficiently Align Latent Actions with Joystick Inputs?} \label{sim:align}

By combining shared autonomy with latent actions, we have a way for humans to precisely control high-DoF robots. But currently the mapping between joystick inputs and latent actions is arbitrary --- and this makes it challenging for users to know how to leverage latent actions. Here we test our proposed alignment method from \textbf{Section~\ref{sec:alignment}}. We explore how efficiently we can learn a simulated human's preferred alignment by using semi-supervised learning and underlying priors.

\begin{figure*}[t]
	\begin{center}
		\includegraphics[width=2.0\columnwidth]{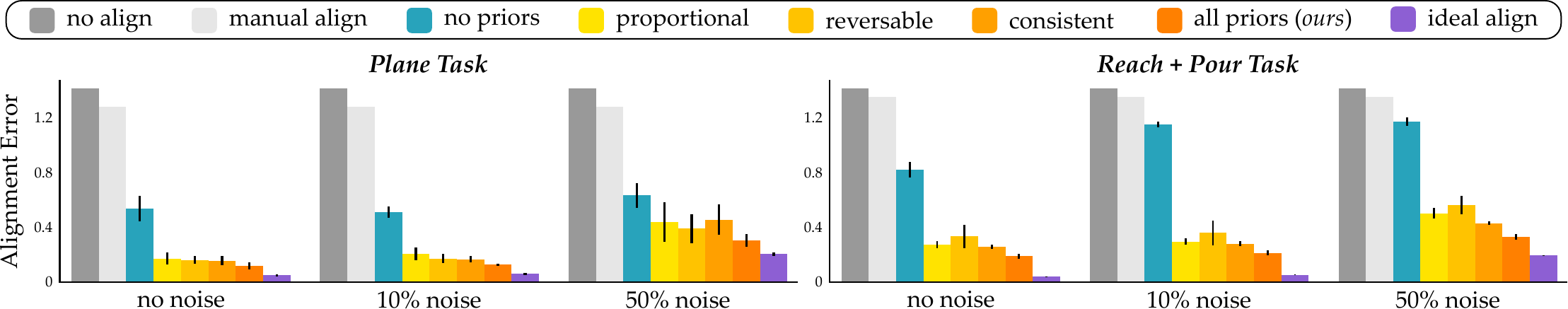}

		\caption{Learning the joystick alignment from simulated and imperfect human feedback. We tested three different tasks with increasing complexity, and here we display the results of the easiest (\textit{Plane}) and the hardest (\textit{Reach + Pour}). \textit{Alignment Error} refers to a the difference between where the simulated human expected the robot to move, and where the robot actually moved. To explore the robustness of our method we varied how noisy the human was when providing their preferences. Across different tasks and levels of human noise, All Priors (our semi-supervised approach) consistently outperformed the other methods, and almost matched an ideal alignment learned from abundant data.}
		\label{fig:sim3}
	\end{center}

\end{figure*}

\p{Setup} Simulated users control the FrankaEmika arm from the previous section. The latent action space and decoder $\phi(z, c)$ were trained using conditional autoencoders. This decoder $\phi$ maps from a 2-DoF latent action $z$ to a $7$-DoF robot arm motion. We now leave the decoder $\phi$ fixed, and focus on learning the alignment model $f$. This alignment model $z = f(u, c)$, takes $2$-DoF inputs from the simulated human and converts them to latent actions for the robot to execute.
   
\p{Tasks} We considered three tasks of increasing complexity. In each task the simulated user interacted with a $2$-DoF joystick.
\begin{enumerate}
    \item \textit{Plane}: The robot moves its end-effector in the $x$-$y$ plane. The human prefers for one joystick DoF to move the robot along the $x$-axis, and the other should move the robot along the $y$ axis.
    \item \textit{Pour}: The robot moves and rotates its end-effector along $z$ axis. The human expects one dimension of the joystick to move the robot up and down, and the other to control a pouring motion.
    \item \textit{Reach \& Pour}: The robot reaches for a bottle, carries it to a bowl, and then pours the contents. The human's preference is divided into two parts: when reaching for the bottle in the $x$-$y$ plane, the human's preference matches \textit{Plane}, and when pouring, the human's preference is the same as in \textit{Pour}.
\end{enumerate}

\p{Learning the Alignment} Before each of these tasks we provided the simulated human with $10$ different motions to label. The human indicated which joystick input $u$ they would expect to use to command the demonstrated motion $(s,s^*)$. In addition to these $10$ labeled datapoints, the robot also collected $1000$ unlabeled motions for self-supervised learning. Given this data, we compared our approach to different baselines. To better understand which priors are useful, we also included an ablation study where the robot learned with only one prior at a time. Overall, the conditions were:
\begin{itemize}
    \item \textbf{No Align}: baseline where $z = u$
    \item \textbf{Manual Align}: the affine transformation that best matches the labeled data
    \item \textbf{No Priors}: trained with the supervised loss $\mathcal{L}_{sup}$ from Equation~(\ref{eq:A3})
    \item \textbf{Proportional}: trained with $\mathcal{L}_{sup} + \mathcal{L}_{prop}$, the proportional prior from Equation~(\ref{eq:A4})
    \item \textbf{Reversable}: trained with $\mathcal{L}_{sup} + \mathcal{L}_{reverse}$, the reversable prior from Equation~(\ref{eq:A5})
    \item \textbf{Consistent}: trained with $\mathcal{L}_{sup} + \mathcal{L}_{con}$, the consistency prior from Equation~(\ref{eq:A6})
    \item \textbf{All Priors}: trained with all of the proposed priors
    \item \textbf{Ideal Align}: supervised loss where the simulated human answers $1000$ queries (instead of $10$).
\end{itemize}
Besides the type of alignment model, we also varied how imperfectly the simulated human answered our queries. We set the coefficient of variance as $0$, $0.1$, and $0.5$ for the simulated human when they answered queries.

\p{Dependent Measures} To determine the quality of each alignment model, we measured the error between where the human \textit{intended} to go and where the robot \textit{actually} went. Specifically, we measured the relative end-effector position and orientation. Let $x^* = \Psi(s^*)$ be the intended end-effector pose, let $x^t$ be the start pose, and let $x^{t+1}$ be where the robot actually ended up: we computed $\| x^* - x^{t-1} \|^2 / \| x^* - x^t \|^2$. For each experiment setting, we reported mean and standard deviation of this metric over $10$ total runs.
\smallskip

\p{Hypotheses} We expected three things during our alignment simulations:
\begin{displayquote}
\textbf{H1.} \textit{With abundant labeled data, the alignment model will learn the human's preferences.}
\end{displayquote}
\begin{displayquote}
\textbf{H2.} \textit{Compared to the fully-supervised baseline, our semi-supervised alignment models that leverage intuitive priors will achieve similar performance with far less human data.}
\end{displayquote}
\begin{displayquote}
\textbf{H3.} \textit{Semi-supervised training with proportional, reversible, and consistent priors will outperform models trained with only one of these priors.}
\end{displayquote}

\p{Results} We highlight results for the \textit{Plane} and \textit{Reach + Pour} tasks in Figure~\ref{fig:sim3}. For models that do not leverage data or personalization --- i.e., No Align and Manual Align --- the error is significantly higher than learning alternatives. With abundant data and noise-free human annotations, Ideal Align provided the gold-standard performance. The success of Ideal Align indicates that our parametrization of the alignment model $f$ is capable of capturing the human's preferences.

In practice, the amount of human feedback will always be limited. We found that our proposed priors were critical when looking at models that only had access to $10$ queries during training. The three semi-supervised models with just one intuitive prior (Proportional, Reversable, and Consistent) performed twice as well as No Priors, the supervised baseline. Putting all of these priors together resulted in even better performance: across different user noise levels, All Priors consistently demonstrated the lowest mean error and standard deviation. This was particularly noticeable when the human oracle is noisy, suggesting that the three priors are indeed \textit{complementary}, and including each of them together brings a performance boost!

Comparing the easier \textit{Plane} task to the more complex \textit {Reach \& Pour} task in Figure~\ref{fig:sim3}, we also saw that using priors became increasingly important as the task got harder. This suggests that --- in complex scenarios --- simply relying on a few labeled examples during supervised learning may lead to severe overfitting. Our intuitive priors for semi-supervised learning effectively mitigate this problem. \smallskip

\p{Summary} Viewed together, the results of our simulations strongly support the hypotheses \textbf{H1}, \textbf{H2}, and \textbf{H3}. Our proposed alignment model successfully learned the mapping between the joystick inputs and latent actions (\textbf{H1}). In settings with limited labels, our proposed alignment model with intuitive control priors reached results that almost match supervised training with abundant data (\textbf{H2}). Finally, in ablation studies, we showed how combining all three proposed priors leads to superior performance and greater training stability than training with a single prior (\textbf{H3}).

\section{User Studies with Non-Disabled Participants}

To evaluate whether actual humans can use learned latent actions to teleoperate robots and perform assistive eating tasks, we conducted four user studies. The participants in these studies are all \textit{non-disabled} adults (we apply our approach with disabled adults in \textbf{Section~\ref{disabled}}). Importantly, we designed these user studies to mimic assistive teleoperation settings. In each study the human user interacts with a 2-DoF joystick, and uses this joystick to control a 7-DoF robot arm. 

The order of the studies roughly parallels \textbf{Sections \ref{sec:latent-action}}, \textbf{\ref{sec:shared-autonomy}}, and \textbf{\ref{sec:alignment}}. We start by comparing latent actions to an existing dataset for assistive eating tasks, and then compare latent actions to the end-effector teleoperation method commonly used on assistive robot arms. Next, we introduce shared autonomy, and perform an ablation study to understand how shared autonomy and latent actions contribute to high-level reaching and precise manipulation tasks. Finally, we learn the alignment between joystick inputs and latent actions by considering intuitive priors, and evaluate how this alignment improves human-robot co-adaptation.

\subsection{Comparing Latent Actions to the HARMONIC Baselines} \label{study1}

\begin{figure}[t]

	\begin{center}
		\includegraphics[width=1.0\columnwidth]{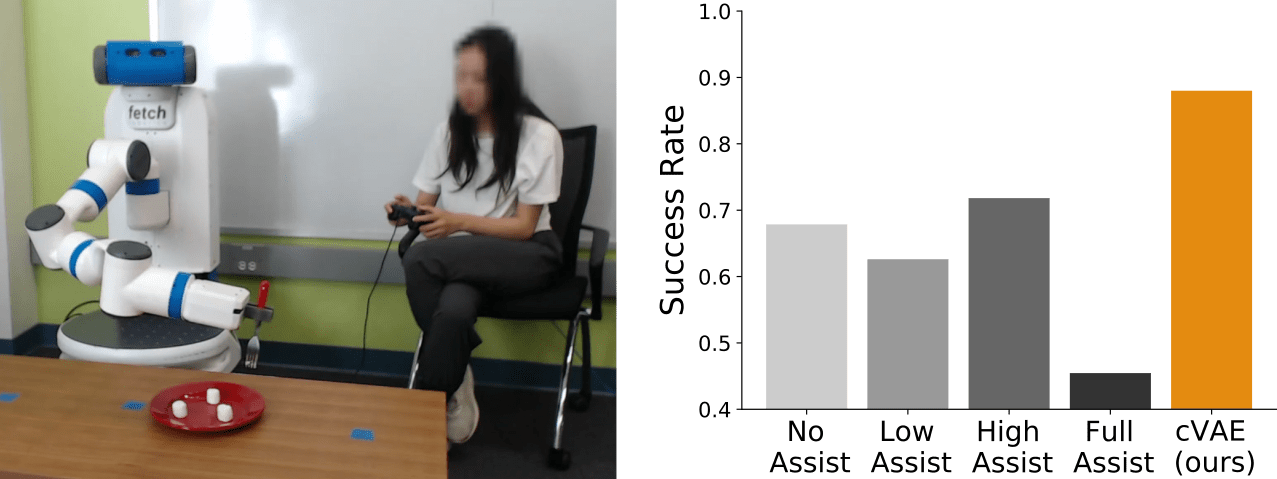}

		\caption{Experimental setup for our user study from Section~\ref{study1}. (Left) in this eating task, the participant uses a two-DoF joystick to guide the robot to reach their desired marshmallow. (Right) we compare our latent action approach to shared autonomy {baselines} from the HARMONIC dataset.}

		\label{fig:harmonic1}
	\end{center}
	
\end{figure}

In our first user study users teleoperate an assistive robot arm using \textit{only learned latent actions}. We baseline their performance against current state-of-the-art approaches for controlling high-DoF assistive robot arms. Specifically, we implement the same assistive eating task as in the HARMONIC dataset \cite{newman2018harmonic}. This task is shown in Figure~\ref{fig:harmonic1}: users guide the robot arm to stab a marshmallow of their choice. There are three different marshmallows on the plate --- i.e., three possible high-level goals --- and only the human knows which of these discrete goals they want to reach. The HARMONIC dataset reports the performance of $24$ people who completed this task with different levels of shared autonomy. Within the HARMONIC baseline, however, the mapping from low-DoF user inputs to high-DoF robot actions is \textit{predefined}, with separate \textit{modes} to control the end-effector's $x$-$y$ position, $z$-$yaw$ position, and $roll$-$pitch$ orientation. We compare the learned latent mapping from \textbf{Section~\ref{sec:latent-action}} to this set of baselines.

\p{Independent Variables} We manipulated the robot's teleoperation strategy with five levels: the four conditions from the HARMONIC dataset plus our proposed learned latent actions. For the first four conditions, the robot uses optimization-based shared autonomy \cite{javdani2018shared} to select assistive actions. The robot either provides no assistance (\textbf{No Assist}) or linearly interpolates between the human's input and the assistive action (\textbf{Low Assist}, \textbf{High Assist}, and \textbf{Full Assist}). \textbf{High Assist} was the most effective strategy from this group: when interpolating, here the assistive action is given twice the weight of the human's commanded action. Within our learned latent actions condition, we applied a conditional variational autoencoder (\textbf{cVAE}) to learn the decoder $\phi(z, c)$ from the demonstrations in the HARMONIC dataset. For now we treat the joystick inputs as the latent actions, so that $z = u$.

\p{Dependent Measures} Before each trial users indicated which marshmallow they want to reach. We measured the fraction of trials in which the robot picks up the correct marshmallow (\textit{Success Rate}), the amount of time needed to complete the task (\textit{Completion Time}), the total magnitude of the human's input (\textit{Joystick Input}), and the distance traveled by the robot's end-effector (\textit{Trajectory Length}).

\p{Hypothesis} We hypothesized that: 
\begin{displayquote}
\textbf{H1.} \emph{Compared the the baselines, latent actions will improve task success while reducing the completion time, joystick inputs, and trajectory length.}
\end{displayquote}

\p{Experimental Setup} Participants interacted with a joystick while watching the robotic arm (see Figure~\ref{fig:harmonic1}). The robot held a fork; during the task, users teleoperated the robot to position this fork directly above their desired marshmallow. We selected the robot's start state, goal locations, and movement speed to be consistent with the HARMONIC dataset.

\p{Participants and Procedure} Our participant pool consisted of ten Stanford University affiliates who provided informed consent ($3$ female, average participant age $23.9\pm2.8$ years). Following the same protocol as the HARMONIC dataset, each participant was given up to five minutes to familiarize themselves with the task and joystick, and then completed five recorded trials using our cVAE approach. At the start of each trial the participant indicates which marshmallow they want the robot to reach; the trial ends after the user indicates that the fork is above their intended marshmallow. We point out that participants only completed the task with the cVAE condition; other teleoperation strategies are benchmarked in Newman \textit{et al.} \cite{newman2018harmonic}.

\begin{figure}[t]

	\begin{center}
		\includegraphics[width=1.0\columnwidth]{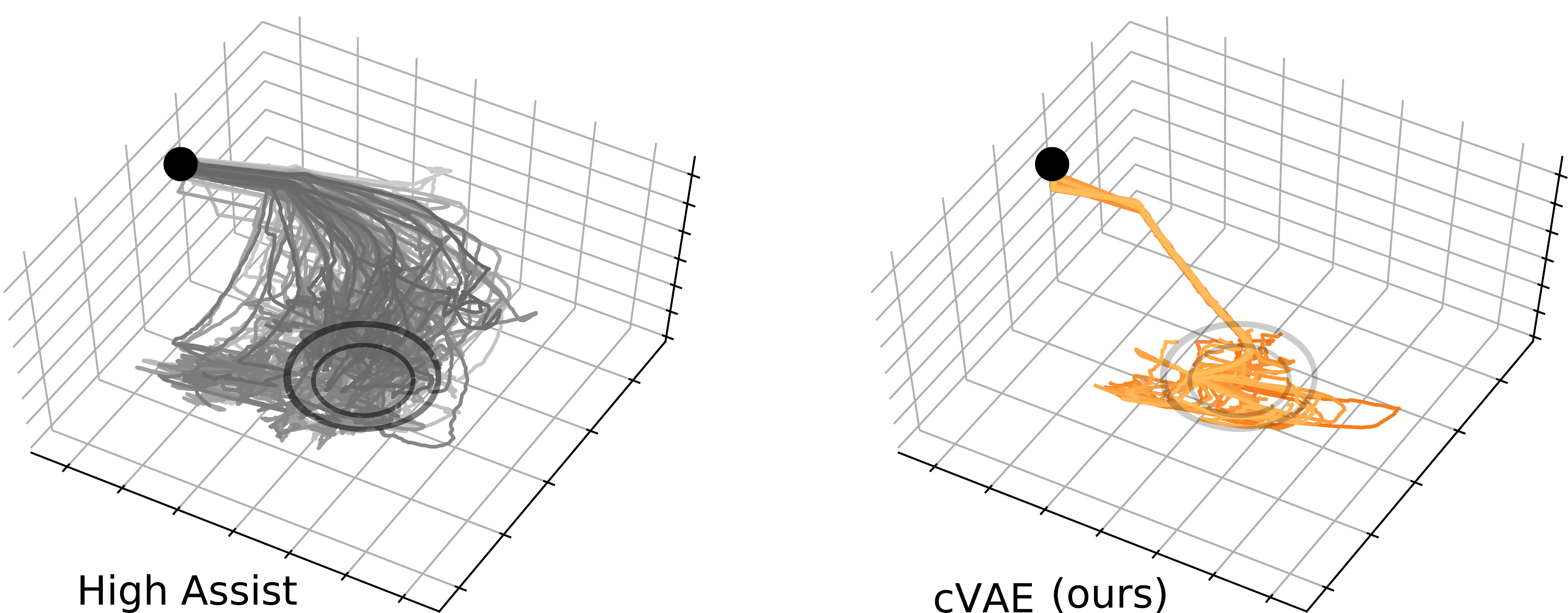}

		\caption{End-effector trajectories from \textbf{High Assist} and \textbf{cVAE} conditions. The robot starts at the black dot, and moves to position itself over the plate.}

		\label{fig:harmonic2}
	\end{center}
	
\end{figure}

\begin{figure}[t]
	\begin{center}
		\includegraphics[width=1.0\columnwidth]{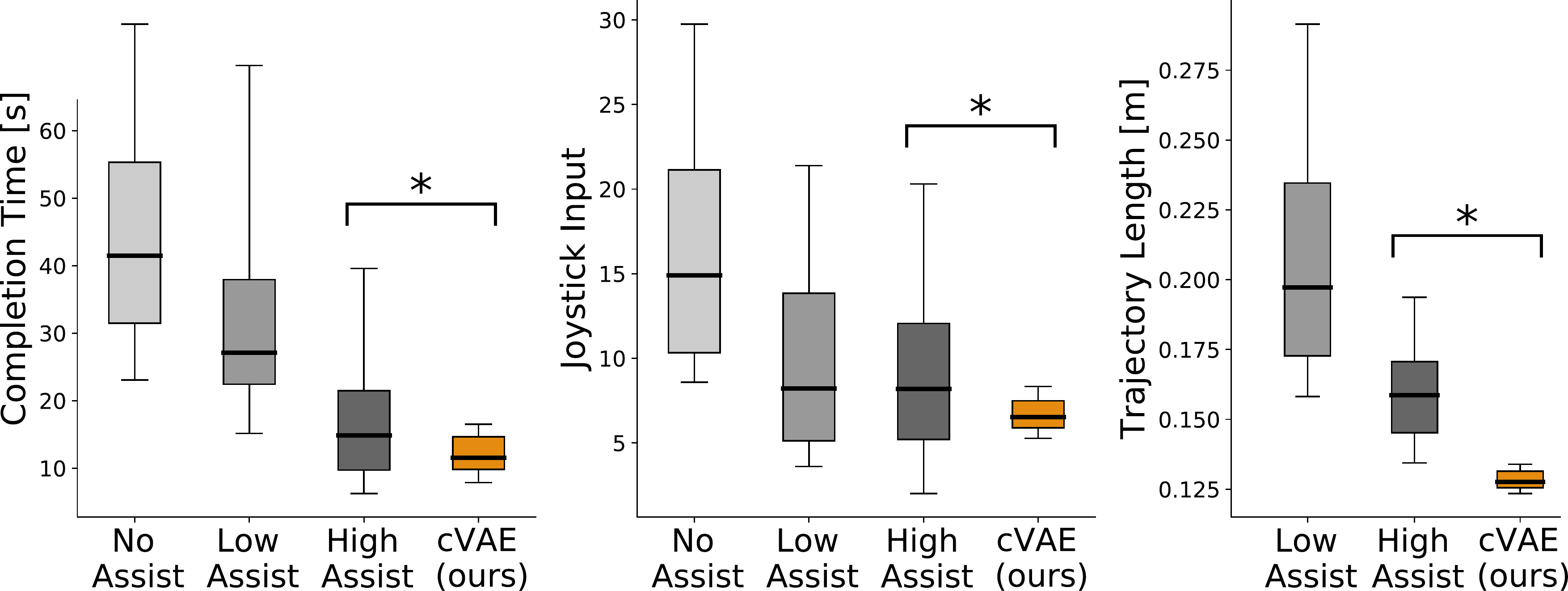}

		\caption{Comparing our objective results to the HARMONIC baseline. We found that \textbf{cVAE} led to faster task completion with less user input and end-effector motion. The \textbf{Full Assist} condition performed worse than \textbf{High Assist} across the board (omitted for clarity). Error bars show the $10$ and $90$ percentiles, and $*$ denotes statistical significance ($p < .05$).}
		
		\label{fig:harmonic3}
	\end{center}
	
\end{figure}

\p{Results} We display example robot trajectories in Figure~\ref{fig:harmonic2} and report our dependent measures in Figures~\ref{fig:harmonic1} and \ref{fig:harmonic3}. Inspecting these example trajectories, we observe that the cVAE model learned latent actions that constrain the robot's end-effector into a region above the plate. Users controlling the robot with cVAE reached their desired morsel in $44$ of the $50$ total trials, yielding a higher \textit{Success Rate} than the assistance baselines. To better compare cVAE to the High Assist condition, we performed independent t-tests. Participants with the cVAE model had significantly lower \textit{Completion Time} ($t(158) = 2.95$, $p < .05$), \textit{Joystick Input} ($t(158) = 2.49$, $p < .05$), and \textit{Trajectory Length} ($t(158) = 9.39$, $p < .001$), supporting hypothesis \textbf{H1}.

\p{Summary} We baselined our learned mapping from low-DoF to high-DoF actions against state-of-the-art shared autonomy approaches with predefined mappings. Users teleoperating an assistive robot with learned latent actions reached their high-level goals more accurately, while requiring less time, effort, and movement.

\subsection{Comparing Latent Actions to End-Effector Teleoperation} \label{study2}

\begin{figure*}[t]

	\begin{center}
		\includegraphics[width=2\columnwidth]{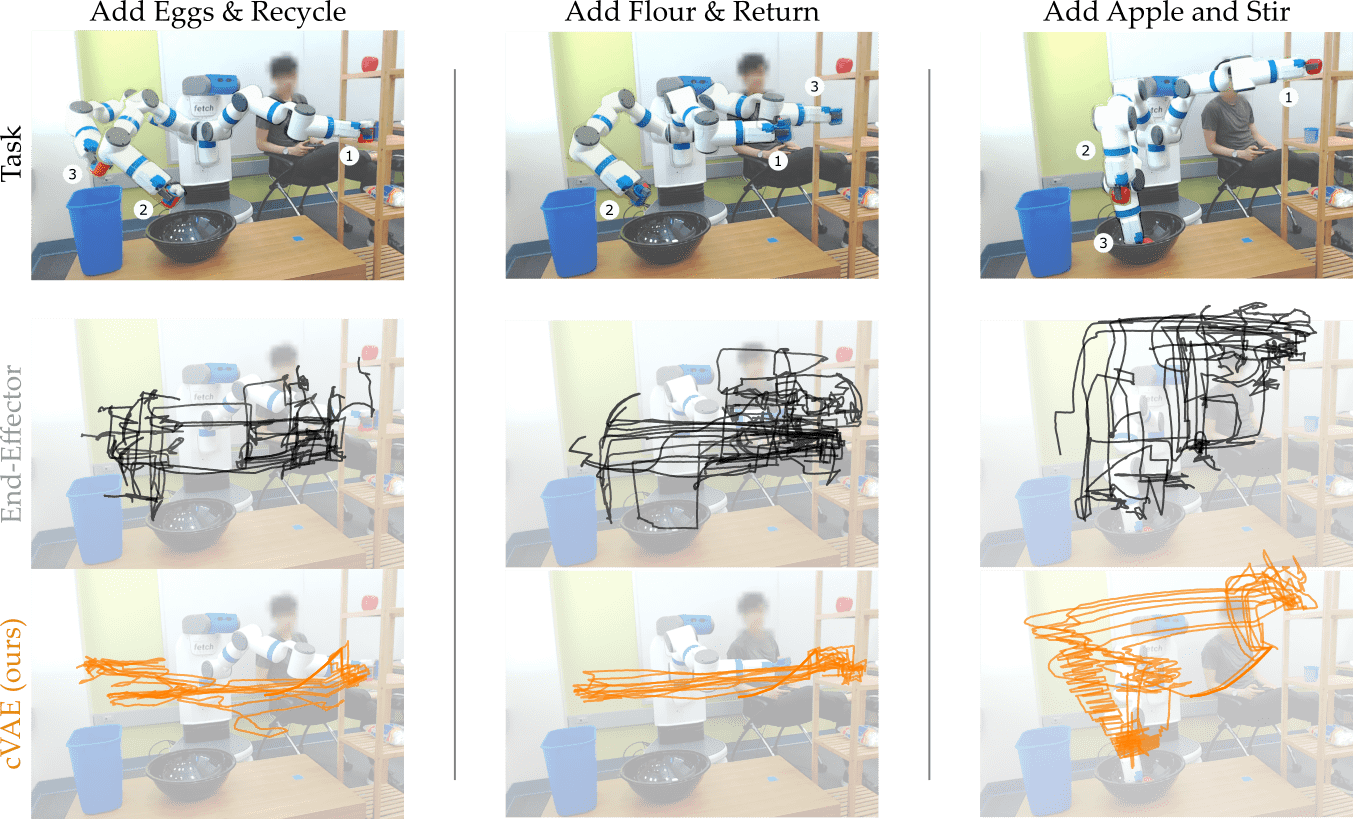}

		\caption{Experimental setup for our user study in Section~\ref{study2}. (Top row) the participant is teleoperating an assistive robot to make their ``apple pie'' recipe. This recipe is broken down into three sub-tasks. On left the robot picks up eggs, pours them into the bowl, then drops the container into the recycling. In middle the robot picks up flour, pours it into the bowl, then returns the container to the shelf. On right the robot grasps an apple, places it in the bowl, then stirs the mixture. (Middle row) example robot trajectories when the person directly controls the robot's \textbf{End-Effector}. (Bottom row) example trajectories when using \textbf{cVAE} to learn latent actions. Comparing the example trajectories, we observe that \textbf{cVAE} resulted in robot motions that more smoothly and directly accomplished the task.}

		\label{fig:cooking1}
	\end{center}
	
\end{figure*}

Real-world assistive eating tasks involve more than just reaching for discrete goals (i.e., stabbing a food morsel). Often objects lie in continuous regions (i.e., anywhere on a shelf), and users must make continuous decisions (i.e., how much water to pour). In our second user study we therefore apply learned latent actions to \textit{continuous} tasks. Consider cooking the ``apple pie'' in Figure~\ref{fig:cooking1}. Assembling this recipe requires picking up ingredients from the shelf, pouring them into a bowl, recycling empty containers --- or returning half-filled containers to the shelf --- and then stirring the mixture. Shared autonomy approaches like \cite{dragan2013policy,newman2018harmonic,javdani2018shared} are not suitable within this setting because: i) the goals lie in continuous regions and ii) the user needs to control both the \textit{high-level goal} that the robot reaches for and the \textit{trajectory} the robot follows to reach that goal (e.g., keeping a cup upright until pouring). Hence, we compare our latent action method against \textit{end-effector teleoperation}. End-effector teleoperation is commonly used by assistive robots \cite{herlant2016assistive}, where the human presses a button to switch modes and control different aspects of the end-effector's motion. For example, in one mode the 2-DoF joystick moves the robot's end-effector in the $x$-$y$ plane.  See videos of this user study here: \\ \href{https://youtu.be/wjnhrzugBj4}{\color{orange}{https://youtu.be/wjnhrzugBj4}}.

\p{Independent Variables} We tested two teleoperation strategies: \textbf{End-Effector} and \textbf{cVAE}. Under End-Effector the user inputs apply a $6$-DoF twist to the robot's end-effector, controlling its linear and angular velocity. Participants interact with two $2$-DoF joysticks, and are given a button to toggle between linear and angular motion \cite{herlant2016assistive,newman2018harmonic,javdani2018shared}. By contrast, in cVAE the participants only interact with one $2$-DoF joystick, i.e., the latent action is $z = [z_1, z_2] \in \mathbb{R}^2$. We emphasize that this cVAE latent action model has the same structure as the one used in our previous user study (Section~\ref{study1}), and does not yet include either shared autonomy or alignment models. We trained cVAE using state-action pairs from kinesthetic demonstrations, where we guided the robot along related sub-tasks such as reaching for the shelf, pouring objects into the bowl, and stirring. Overall, the cVAE was trained with less than $7$ minutes of demonstration data.

\p{Dependent Measures -- Objective} We measured the total amount of time it took for participants to complete the entire cooking task (\textit{Completion Time}), as well as the magnitude of their inputs (\textit{Joystick Input}).

\p{Dependent Measures -- Subjective} We administered a $7$-point Likert scale survey after each condition. Questions were separated into six scales, such as ease of performing the task (\textit{Ease}) {and consistency of the controller (\textit{Consistent})}. Once users had completed the task with both strategies, we asked comparative questions about which they preferred (\textit{Prefer}), which was \textit{Easier}, and which was more \textit{Natural}.

\p{Hypotheses} We had the following hypotheses: 
\begin{displayquote}
\textbf{H1.} \emph{Users controlling the robot arm with low-DoF latent actions will complete the cooking task more quickly and with less overall effort.}
\end{displayquote}
\begin{displayquote}
\textbf{H2.} \emph{Participants will perceive the robot as easier to work with in the cVAE condition, and will prefer the cVAE over End-Effector teleoperation.}
\end{displayquote}

\p{Experimental Setup} We developed a cooking task where the person is making a simplified ``apple pie.'' As shown in Figure~\ref{fig:cooking1}, the assistive robot must sequentially pour eggs, flour, and an apple into the bowl, dispose of their containers, and stir the mixture. The user sat next to the robot and controlled its behavior with a handheld joystick. During the experiment we introduced variance by intermittently changing the location of the shelf, bowl, and recycling bin.

\p{Participants and Procedure} Eleven members of the Stanford University community ($4$ female, age range $27.4 \pm 11.8$ years) provided informed consent to participate in this study. Similar to the other user studies described in this paper, we used a within-subjects design, and counterbalanced the order of our two conditions. Four of our subjects had prior experience interacting with the robot used in our experiment.

Before starting the study, participants were shown a video of the cooking task. Participants then separately completed the three parts of the task as visualized in Figure~\ref{fig:cooking1}; we reset the robot to its home position between each of these sub-tasks. After the user completed these sub-tasks, we re-arranged the placement of the recycling and bowl, and users performed the entire cooking task without breaks. Participants were told about the joystick interface for each condition, and could refer to a sheet that labelled the joystick inputs.

\begin{figure}[t]

	\begin{center}
		\includegraphics[width=0.8\columnwidth]{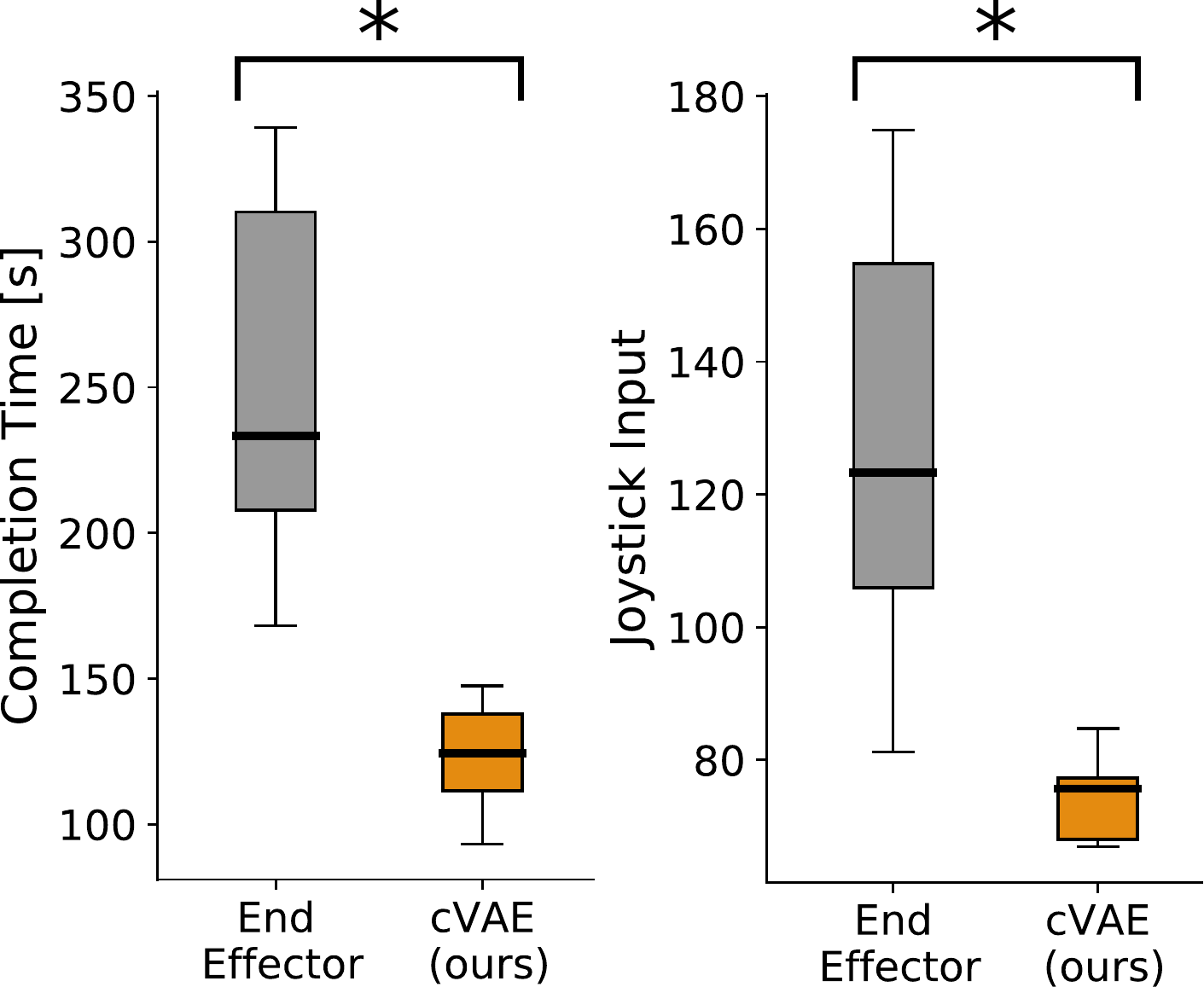}

		\caption{Objective results from assembling an ``apple pie.'' These results were collected across the entire cooking task (combining each sub-task from Figure~\ref{fig:cooking1}). We compare using just latent actions to direct end-effector teleoperation.}

		\label{fig:cooking2}
	\end{center}
	
\end{figure}

\begin{figure}[t]

	\begin{center}
		\includegraphics[width=0.9\columnwidth]{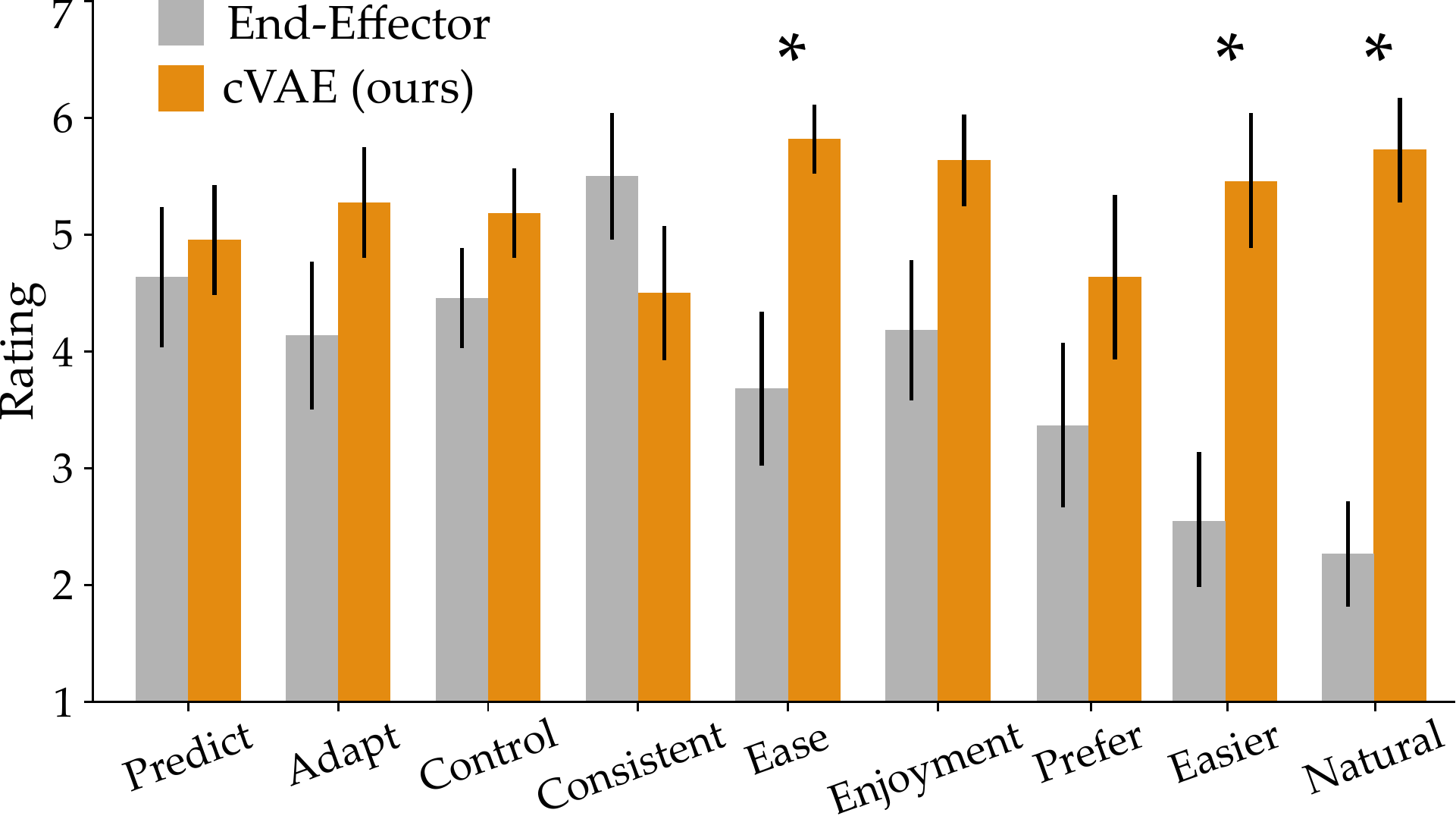}

		\caption{Subjective results from assembling an ``apple pie.'' Higher ratings indicate participant agreement. Participants thought our approach required less effort (\textit{Ease}), made it easier to complete the task (\textit{Easier}), and produced more natural robot motion (\textit{Natural}) as compared to End-Effector control.}

		\label{fig:cooking3}
	\end{center}
	
\end{figure}

\p{Results -- Objective} Our objective results are summarized in Figure~\ref{fig:cooking2}. When using cVAE to complete the entire recipe, participants finished the task in less time ($t(10)=-6.9$, $p<.001$), and used the joystick less frequently ($t(10)=-5.1$, $p<.001$) as compared to direct End-Effector teleoporation.

\p{Results -- Subjective} We display the results of our $7$-point Likert scale surveys in Figure~\ref{fig:cooking3}. Before reporting these results, we first confirmed the reliability of our scales. We then leveraged paired t-tests to compare user ratings for End-Effector and cVAE conditions. We found that participants perceived cVAE as requiring less user effort ($t(10)=2.7$, $p<.05$) than End-Effector. Participants also indicated that it was easier to complete the task with cVAE ($t(10)=2.5$, $p<.05$), and that cVAE caused the robot to move more naturally ($t(10)=3.8$, $p<.01$). The other scales were not significantly different.

\p{Summary} We focused on a cooking task with continuous high-level goals, and compared latent actions to end-effector teleoperation. When controlling the robot with latent actions, users completed the cooking task more quickly and with less effort (\textbf{H1}). Participants believed that the cVAE approach led to more natural robot motion, and indicated that it was easier to perform the task with latent actions. However, participants did not indicate a clear preference for either strategy (\textbf{H2}). We will explore ways to improve user satisfaction in our next user study, where we combine latent actions with shared autonomy to assist the human.

\subsection{Combining Learned Latent Actions with Shared Autonomy} \label{study3}

\begin{figure*}[t]
	\begin{center}
		\includegraphics[width=2\columnwidth]{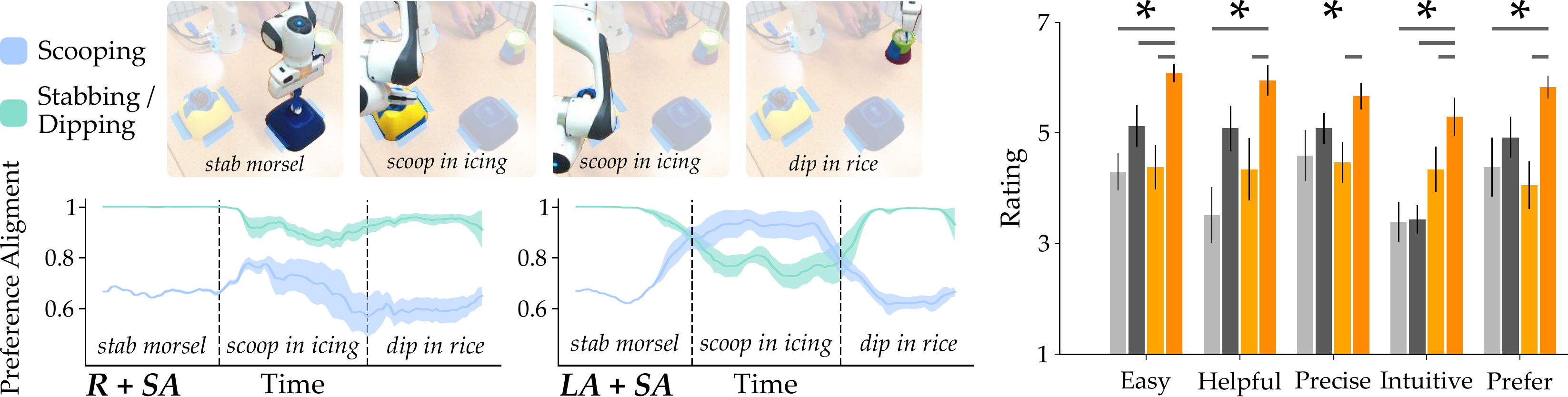}
		\caption{Experimental setup for our user study in Section~\ref{study3}. (Left) The \textit{Dessert} task consists of 3 phases: stabbing the marshmallow, scooping it in icing, and dipping it in rice. We identified the end-effector directions needed to complete these fine-grained preferences, and plotted the average dot product between the desired and actual directions (\textit{Preference Alignment}). In the R+SA condition, users executed the entire task in a stabbing/dipping orientation. By contrast, with LA+SA users correctly adjusted the scooping preference in the second phase of the task. (Right) We plot the results of our 7-point Likert scale surveys. Color to method mappings are consistent with Figure~\ref{fig:dessert2}, and $*$ indicate statistical significance $(p< 0.05)$.}
		\label{fig:dessert1}
	\end{center}

\end{figure*}

In our first two user studies we tested latent actions \textit{by themselves}, without any robotic assistance. Our results indicate that latent actions objectively outperform both the HARMONIC baselines and end-effector control; however, the participant's subjective responses are mixed. Plus, so far we have only dealt with high-level reaching goals --- but assistive eating also involves fine-grained manipulation, where the user must cut, stab, and scoop their food. Here we tackle both issues by performing a user study with \textit{eating tasks} (see Figure~\ref{fig:dessert1}). Participants must control the robot towards their goal plate, and then carefully adjusted the robot's motion to cut, stab, and scoop different foods. We explore how users leverage learned latent actions \textit{with shared autonomy} to complete this task. Specifically, we conduct an ablation study across latent actions and shared autonomy, and determine whether latent actions alone, shared autonomy alone, or their combination best results in high-level reaching and precise manipulation. See videos of this user study here: \\ \href{https://youtu.be/7BouKojzVyk}{\color{orange}{https://youtu.be/7BouKojzVyk}}.

\p{Experimental Setup} Each participant attempted to complete two dishes: an \textit{Entree task} and a \textit{Dessert task}. In \textit{Entree}, users had to perform multiple precise motions at the \textit{same} goal. Here participants i) guided the robot towards a bowl with tofu, ii) cut off a slice of tofu, and iii) stabbed and scooped the slice onto their plate. In \textit{Dessert} the participants had to convey their preferences at \textit{multiple} goals: they i) stabbed a marshmallow in the middle plate, ii) scooped it through icing at the right plate, and then iii) dipped it in rice at the left plate before iv) setting the marshmallow on their plate. In both tasks subjects sat next to the robot, mimicking a wheelchair-mounted arm.

\p{Independent Variables} We conducted a 2x2 factorial design that separately varied  \textit{Control Interface} and \textit{Robot Assistance}.

For the control interface, we tested a state-of-the-art direct teleoperation scheme (\textit{Retargetting}), where the user's joystick inputs map to the 6-DoF end-effector twist of the robot \cite{rakita2017motion}. We compared this direct teleoperation baseline to our learned latent actions: here the robot interprets the meaning of the human's inputs based on the current context.

For robot assistance, we tested \textit{with} and \textit{without shared autonomy}. We implemented the shared autonomy algorithm from \cite{javdani2018shared}, which assists the robot towards likely human goals.

Crossing these two separate factors, we totaled four different conditions: 
\begin{itemize}
    \item \textbf{R}: Retargeting with no shared autonomy
    \item \textbf{R+SA}: Retargetting with shared Autonomy
    \item \textbf{LA}: Latent actions with no shared autonomy
    \item \textbf{LA+SA}: Latent actions with shared autonomy
\end{itemize}
Note that the \textbf{LA+SA} condition is our proposed approach (Algorithm~\ref{deltaco}). However, we still omit the alignment model $f$, so for now the latent action $z$ is set equal to the joystick input $u$.

\p{Model Training} We provided kinesthetic demonstrations $\mathcal{D}$ that guided the robot towards each plate, and then performed cutting, stabbing, and scooping motions at these goals. The robot learned the latent action space from a total of $20$ minutes of kinesthetic demonstrations. Because we are combining latent actions with shared autonomy, we also recorded the belief $b$ during these demonstrations. In the LA+SA condition, the robot used context $c=(s,b)$ to decode human inputs.

\p{Dependent Measures -- Objective}
We recorded the amount of time users took to complete each task (\emph{Total Time}), as well as the amount of time spent without providing joystick inputs (\emph{Idle Time}). We also computed proxy measures of the high-level goal accuracy and low-level preference precision. For goals, we measured the robot's total distance to the closest plate throughout the task (\emph{Goal Error}). For preferences, we recorded the dot product between the robot's actual end-effector direction and the true end-effector directions needed to precisely cut, stab, and scoop (\textit{Preference Alignment}).

\p{Dependent Measures -- Subjective} We administered a $7$-point Likert scale survey after each condition. Questions were organized along five scales: how \textit{Easy} it was to complete the tasks, how \textit{Helpful} the robot was, how \textit{Precise} their motions were, how \textit{Intuitive} the robot was to control, and whether they would use this condition again (\textit{Prefer}).

\p{Participants and Procedure}
We recruited $10$ subjects from the Stanford University student body to participate in our study ($4$ female, average age $23.5 \pm 2.15$ years). All subjects provided informed written consent prior to the experiment. We used a within-subjects design: each participant completed both tasks with all four conditions (the order of the conditions was counterbalanced). Before every trial, users practiced teleoperating the robot with the current condition for up to $5$ minutes.

\p{Hypotheses} We tested three main hypotheses:
\begin{displayquote}
\textbf{H1.} \emph{Users controlling the robot with shared autonomy will more accurately maintain their goals.}
\end{displayquote}
\begin{displayquote}
\textbf{H2.} \emph{Latent actions will help users more precisely perform manipulation tasks.}
\end{displayquote}
\begin{displayquote}
\textbf{H3.} \emph{Participants will complete the task most efficiently with combined LA+SA.}
\end{displayquote}

\begin{figure}[t]
	\begin{center}
		\includegraphics[width=1\columnwidth]{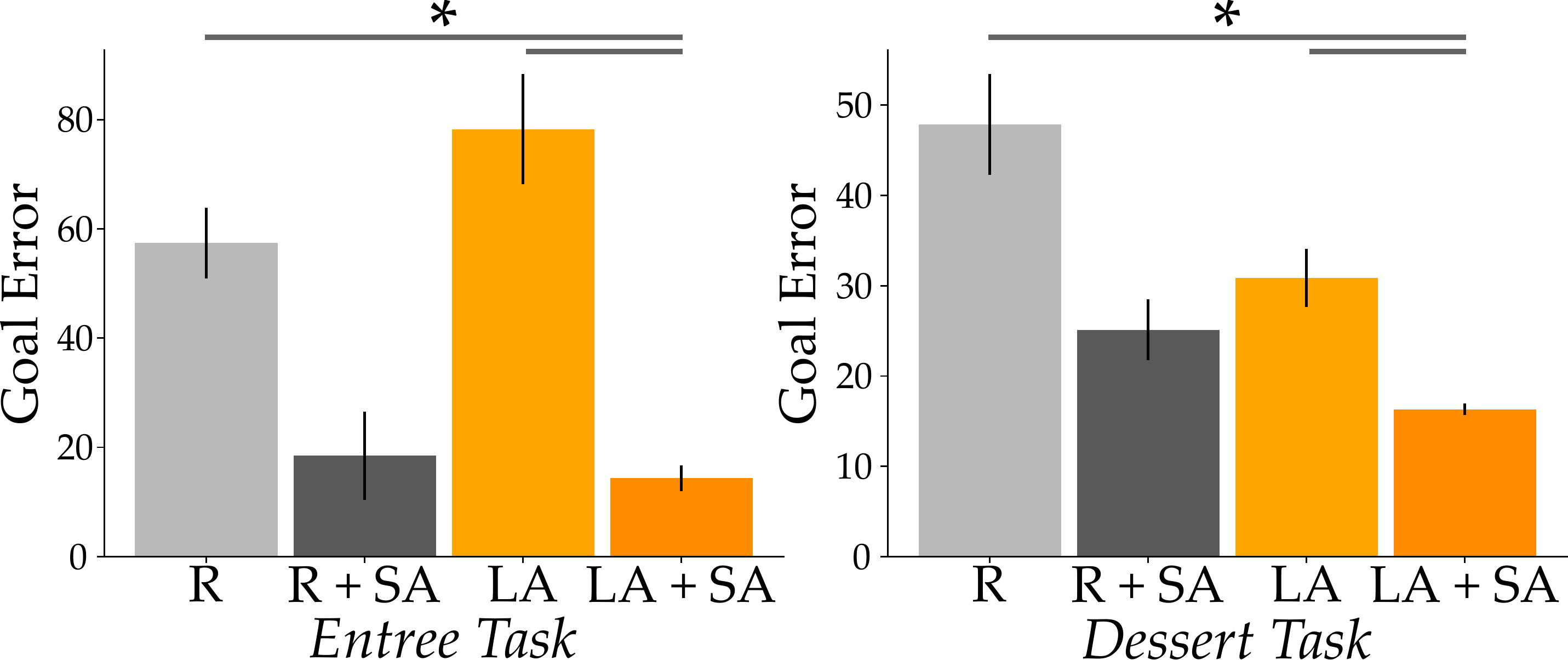}

		\caption{Error between the end-effector and nearest goal during the eating tasks. Adding shared autonomy (SA) decreased this error across both mapping strategies ({R} and {LA}).}
		\label{fig:dessert2}
	\end{center}

\end{figure}

\begin{figure}[t]
	\begin{center}
		\includegraphics[width=1\columnwidth]{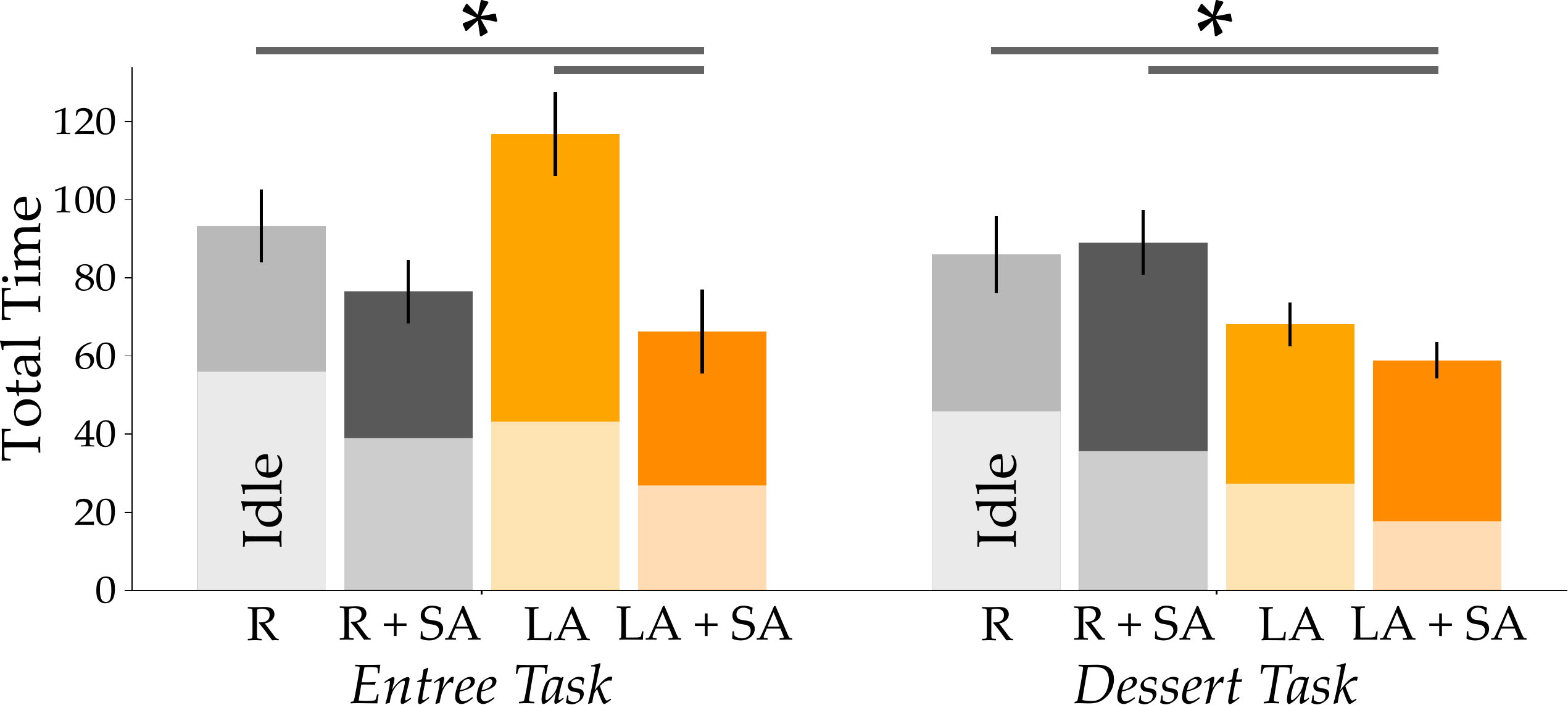}

		\caption{Time taken to complete the eating task (solid) and time spent idle (light). Users completed both eating tasks most efficiently with our proposed combination of shared autonomy and latent actions (LA+SA).}
		\label{fig:dessert3}
	\end{center}

\end{figure}

\p{Results -- Objective} To explore \textbf{H1}, we analyzed the \textit{Goal Error} for methods with and without shared autonomy (see Figure~\ref{fig:dessert2}). Across both tasks, users interacting with shared autonomy reached their intended goals significantly more accurately $(F(1,18)= 29.9, p<.001)$. Breaking this down by condition, users incurred \emph{less} error with LA+SA than with LA ($p<.001$), and --- similarly --- users were more accurate with R+SA than with R ($p<.05$). Building on our prior user study results, this indicates that adding shared autonomy improves the performance of our latent action approach.

So shared autonomy helped users more accurately maintain their goals --- but were participants able to complete the precise manipulation tasks at those goals? We visualize the \textit{Preference Alignment} for \textit{Dessert} in Figure~\ref{fig:dessert1}, specifically comparing R+SA to LA+SA. We notice that --- when using direct teleoperation --- participants remained in a stabbing preference throughout the task. By contrast, users with latent actions \textit{adjusted} between diffrent manipulation tasks: stabbing the marshmallow, scooping it in icing, and dipping it in rice. These results support \textbf{H2}, suggesting that latent actions enable users to precisely manipulate the robot.

\begin{figure*}[t]

	\begin{center}
		\includegraphics[width=2\columnwidth]{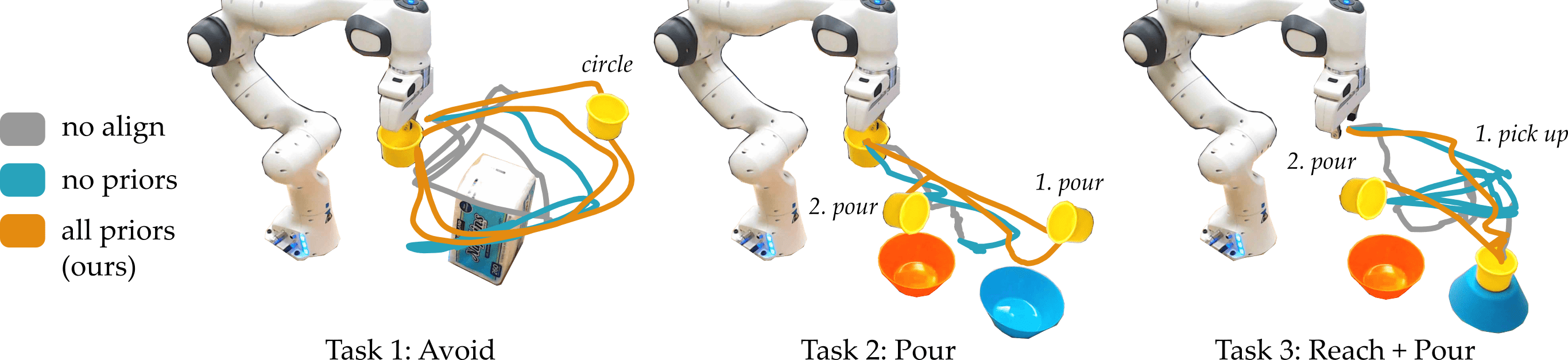}

		\caption{Experimental setup for the user study in Section~\ref{study4}. We visualize a single user's end-effector trajectories for the \textit{Avoid}, \textit{Pour} and \textit{Reach + Pour} tasks. Participants teleoperated the 7-DoF Panda robot arm without any alignment model (No Align), with an alignment model trained only on their supervised feedback (No Priors), and with our proposed method, where the robot generalizes the human's feedback using intuitive priors (All Priors). For both No Align and No Priors baselines, we can see moments where the human gets confused, counteracts themselves, or fails to complete the task.}
		
		\label{fig:alignment1}
		
	\end{center}
	
\end{figure*}

Now that we know the benefits of shared autonomy and latent actions individually, what happens when we focus on their combination? Inspecting Figure~\ref{fig:dessert3}, participants using LA+SA were able to complete both tasks more efficiently. Summing times across both tasks, and then performing pair-wise comparisons between each condition, we found that LA+SA outperformed the alternatives for both \textit{Total Time} ($p < .05$) and \textit{Idle Time} ($p < .05$). Overall, we found that LA+SA users completed the task the fastest and with the least error.

\p{Results -- Subjective} We find further support for \textbf{H3} in the user's feedback. The results of t-tests comparing LA+SA to the other conditions are reported in Figure~\ref{fig:dessert1} (where an $*$ denotes $p<.05$). Responses suggest that users were most ``comfortable" when performing precise manipulation with LA+SA. We emphasize the improvement in these subjective results as compared to Figure~\ref{fig:cooking3}, where users ranked latent actions similarly to direct end-effector control. We conclude that incorporating shared autonomy improves the human's experience when leveraging latent actions.

\p{Summary} We conducted two eating tasks where participants needed to i) reach for high-level goals and ii) precisely manipulate food items at those goals. We found that both shared autonomy and latent actions improved performance: including shared autonomy decreased end-effector error, while using latent actions helped users quickly transition between different fine-grained manipulations. These results compliment Section~\ref{study1} (where we compare latent actions to shared autonomy) and Section~\ref{study2} (where we compare latent actions to end-effector control). But now we also have that the \textit{combination} of shared autonomy and latent actions outperforms either of these approaches alone.

\subsection{Learning the Alignment between Joystick Inputs and Latent Actions} \label{study4}

Now that we have shown the benefits of latent actions --- and how latent actions can incorporate shared autonomy --- we finally turn our attention to aligning the human's joystick inputs with the learned latent space. This final user study builds on \textbf{Section~\ref{sec:alignment}}. Prior to the study, we learn a latent action space that maps from 2-DoF inputs to 7-DoF robot actions. During the study, we ask participants to label a few sample robot motions with their preferred joystick input. The robot uses intuitive priors to generalize from these labels and learn a personalized mapping from joystick inputs $u$ to latent actions $z$. In the previous studies we have simply set $z=u$, and we leverage this as one of the baselines in this user study. We also evaluate the effects of our intuitive priors, and compare learning the alignment model \textit{with} and \textit{without} these priors.  See videos of this user study here: \href{https://youtu.be/rKHka0_48-Q}{\color{orange}{https://youtu.be/rKHka0\_48\-Q}}.

\begin{figure*}[t]

	\begin{center}
		\includegraphics[width=2\columnwidth]{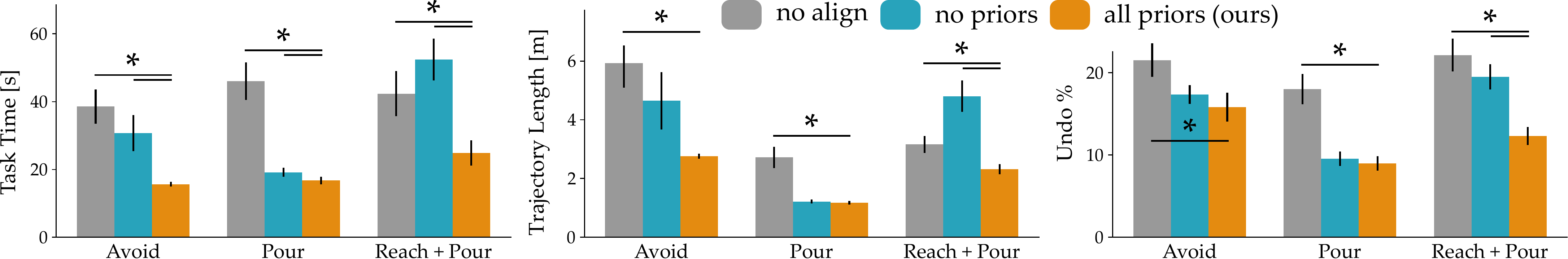}

		\caption{Objective results from our alignment user study. (Left) Average time taken to complete each task. (Middle) Average trajectory length as measured in end-effector space. (Right) Percentage of the time people spend undoing their actions. Error bars show the standard deviation across the $10$ participants, and colors match Figure~\ref{fig:alignment1}. Asterisks denote statistically significant pairwise comparisons between the two marked strategies ($p < .05$).}
		\label{fig:alignment3}
		
	\end{center}
	\vspace{-1em}

\end{figure*}

\begin{figure}[t]
	\begin{center}
		\includegraphics[width=1.0\columnwidth]{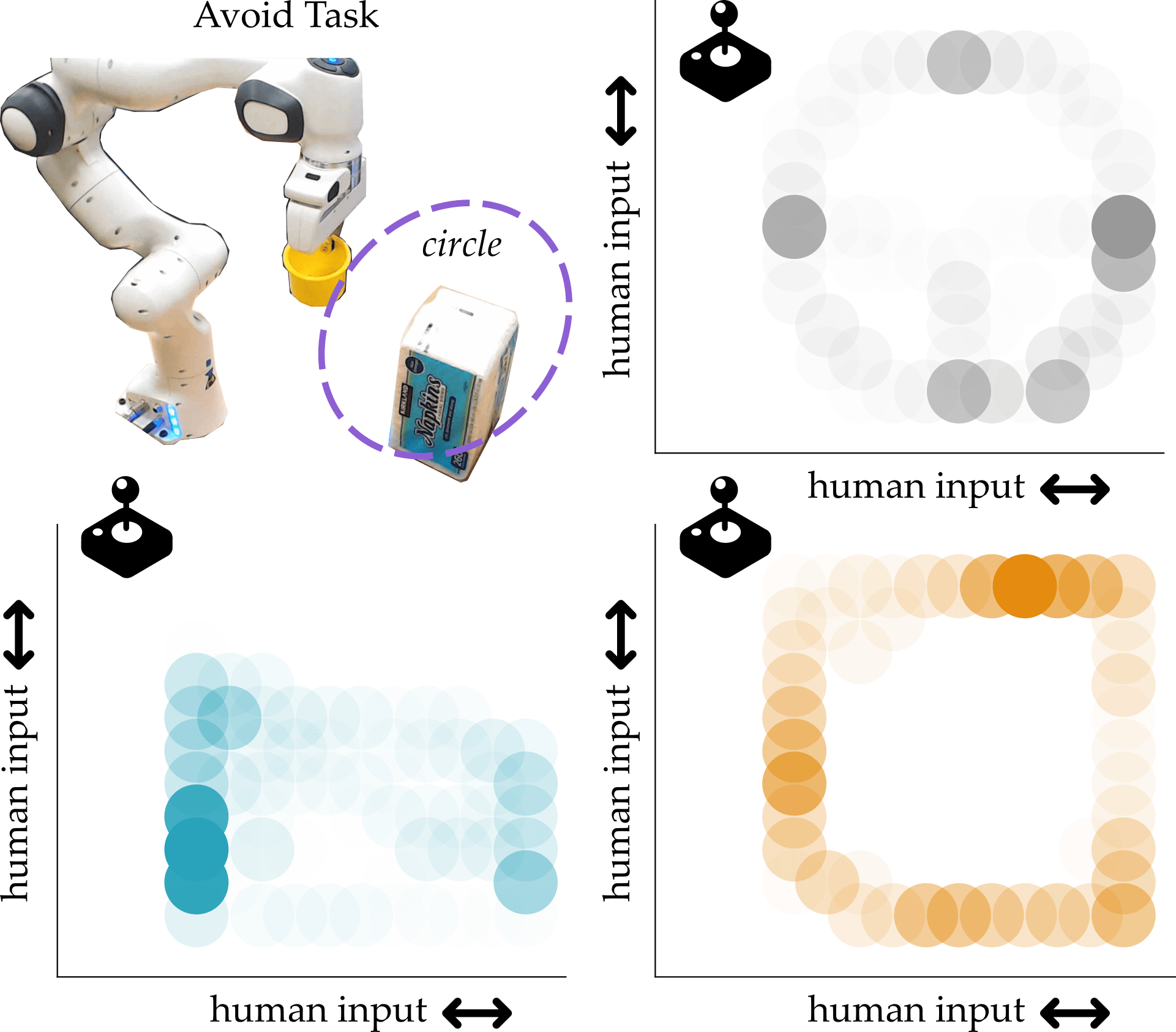}
		\caption{Heatmaps of the participants' joystick inputs during the \textit{Avoid} task. For No Align in the upper right, people primarily used the cardinal directions. For No Priors in the bottom left, the joystick inputs were not clearly separated, and no clear pattern was established. 
		For our All Priors model on the bottom right, however, we observed that the human inputs were \textit{evenly distributed}. This indicates that the users smoothly completed the task by continuously manipulating the joystick in the range $[-1, +1]$ along both axes.}
		\label{fig:alignment2}
	\end{center}

\end{figure}

\begin{figure}[t]

	\begin{center}
		\includegraphics[width=1\columnwidth]{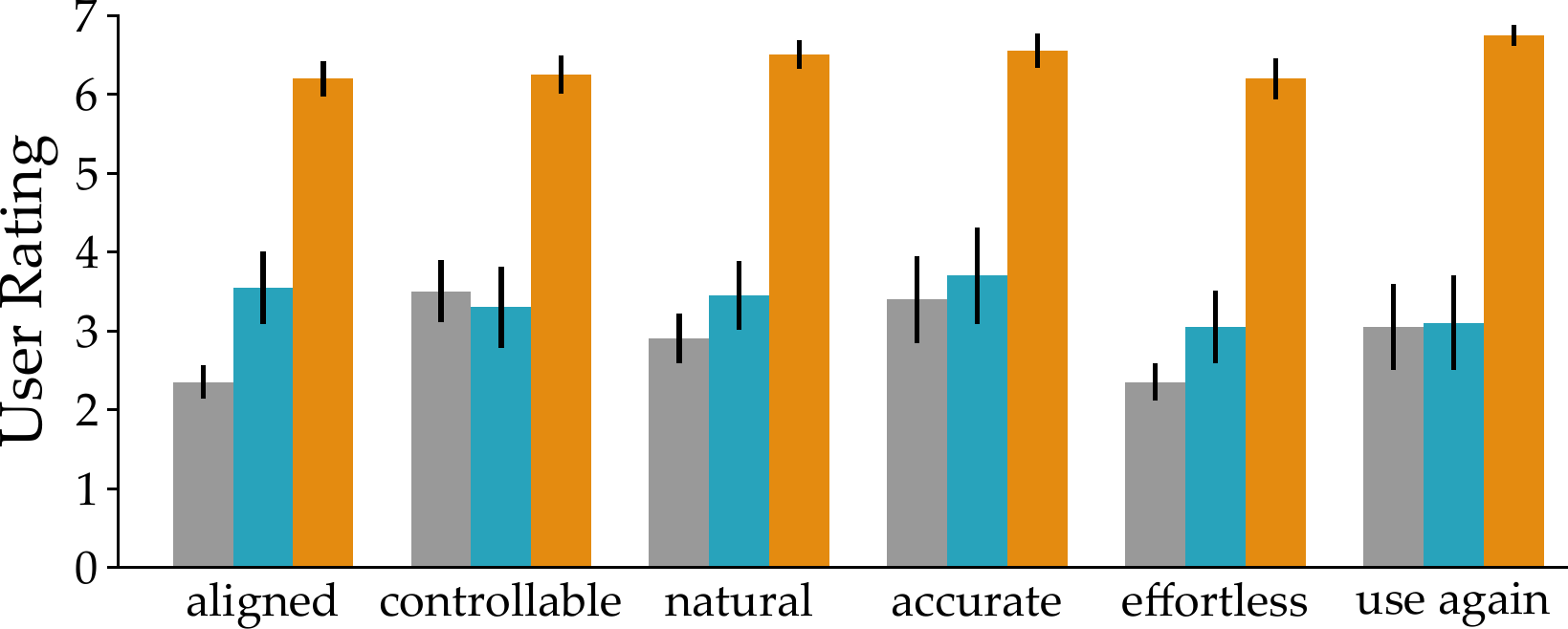}

		\caption{Results from our $7$-point Likert-scale survey after the alignment user study. The legend is the same as in Figure~\ref{fig:alignment3}. Higher ratings indicate agreement. Users thought that our learned model with intuitive priors aligned with their preferences, was easy to control, and improved efficiency --- plus they would choose to use it again. Pairwise comparisons between our approach and the baselines are statistically significant across the board.}
		\label{fig:alignment4}
		
		\vspace{-2em}
		
	\end{center}

\end{figure}

\p{Tasks} Similar to our simulation experiments from Section~\ref{sim:align}, we considered three different tasks. These tasks are visualized in Figure~\ref{fig:alignment1}.

\begin{enumerate}
\item \textit{Avoid}: The robot arm moves its end-effector in a horizontal plane. Users are asked to guide the robot around an obstacle without colliding with it.
\item \textit{Pour}: The robot arm is holding a cup, and users want to pour this cup into two bowls. Users are asked to first pour into the farther bowl, before moving the cup back to the start and pouring into the closer bowl.
\item \textit{Reach \& Pour}: Users start by guiding the robot towards a cup and then pick it up. Once the users reach and grasp the cup, they are asked to take the cup to a target bowl, and finally pour into it.
\end{enumerate}

\p{Independent Variables} For each of the tasks described above, we compared three different alignment models. \textbf{No Align} corresponds to what we have done in the previous user studies: setting $z=u$, so that all users must adapt to the same latent action alignment. We compare this to two \textit{personalized} approaches. First is \textbf{No Priors}, where we train the alignment model $f$ just using the supervised loss (i.e., the robot only learns from the human's labeled data). We contrast this to \textbf{All Priors}, where the robot leverages semi-supervised learning (i.e., the robot also considers the proportional, reversible, and consistent priors we anticipate that the user will expect). Each condition learns from the same human feedback; we emphasize that All Priors uses the same number of human labels as No Priors.

\p{Dependent Measures} To evaluate the effectiveness of these different alignment strategies, we recorded \textit{Task Time} and \textit{Trajectory Length}. We also calculated the percentage of the time that users \textit{Undo} their actions by significantly changing the joystick direction --- undoing suggests that the alignment is not quite right, and the human is still adapting to the robot's control strategy. Besides these objective measures, we also collected subjective feedback from the participants through 7-point Likert scale surveys. 

\p{Participants and Procedure}
We recruited 10 volunteers that provided informed written consent (3 female, ages $23.7 \pm 1.5$). Participants used a 2-axis joystick to teleoperate the 7-DoF robot arm, and completed three manipulation tasks inspired by assistive settings. At the start of each task, we showed the user a set of robot movements and ask them to provide their preferred input on the joystick --- i.e., ``if you wanted the robot to perform the movement you just saw, what joystick input would you provide?'' Users answered $7$ queries for task \textit{Avoid}, $10$ queries for task \textit{Pour} and $30$ queries for task \textit{Reach \& Pour}. After the queries finished, the users started performing tasks sequentially using each of the alignment strategies. The order of alignment strategies was counterbalanced.

\begin{figure*}[t]
	\begin{center}
		\includegraphics[width=1.9\columnwidth]{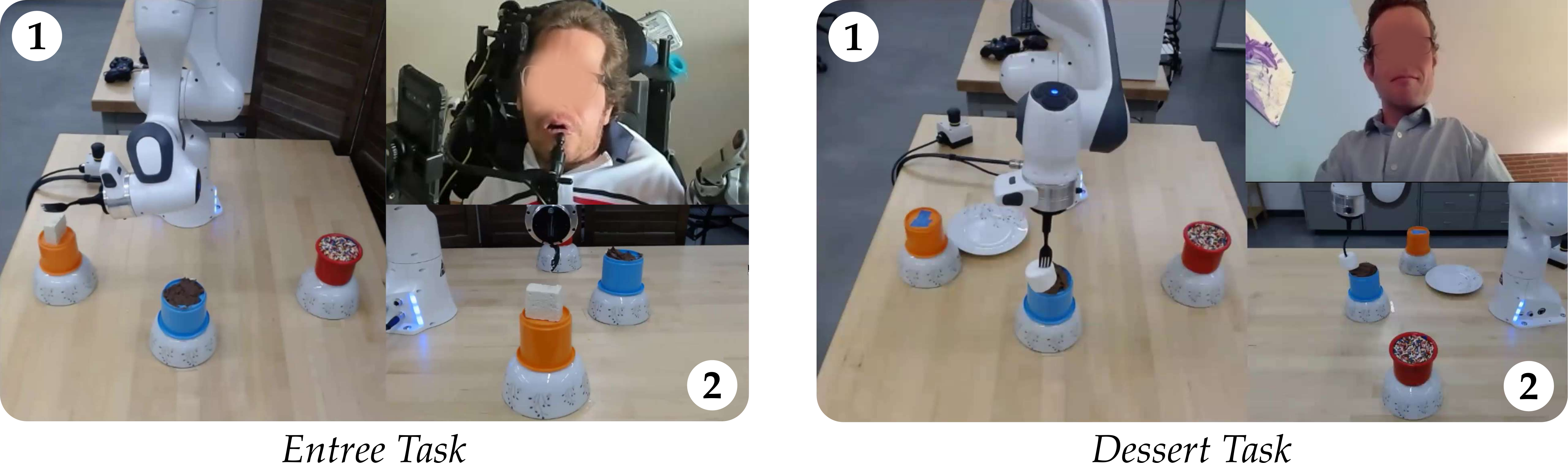}

		\caption{Experimental setup for our case study with disabled persons (Section~\ref{disabled}). Two adult males who employ assistive devices when eating volunteered to participate in the study. Due to safety restrictions, this study was conducted remotely: participants used an online joystick to teleoperate our robot in real-time while watching live-streamed video. Here we show examples of the participants' views during \textit{Entree} and \textit{Dessert} tasks. Camera (1) is a front-view of the robot and the high-level goals, and camera (2) is a side view of the same. Enlarged images of the participants are shown in the top right of each frame.}
		\label{fig:pilot1}
	\end{center}

    \vspace{-1em}

\end{figure*}

\p{Hypothesis} We hypothesize that:
\begin{displayquote}
\textbf{H1.} \emph{An alignment model learned from user-specific feedback and generalized through intuitive priors will make it easier for humans to control the robot and perform assistive manipulation tasks.}
\end{displayquote}

\p{Results}
The objective results of our user study are summarized in Figure~\ref{fig:alignment3}. Across tasks and metrics, our model with All Priors outperforms the two baselines. In addition, our model not only has the best average performance, but it also demonstrates the least variance. Similar to our simulation results from Section~\ref{sim:align}, when the task is difficult (i.e., \textit{Reach \& Pour}), the performance of No Priors drops significantly compared to simpler tasks, reinforcing the \textit{importance of priors} when in learning complex alignment models.

We also illustrate our survey responses in Figure \ref{fig:alignment4}. Across the board, we found that users exhibited a clear preference for our proposed method. Specifically, they perceived All Priors as resulting in better alignment, more natural, accurate, and effortless control, and would elect to use it again. These subjective results highlight the importance of \textit{personalization} when controlling high-DoF systems --- we contrast these results to Figure~\ref{fig:cooking3}, where participants perceived the unaligned controller as somewhat unintuitive.

To better visualize the user experiences, we also display example robot end-effector trajectories from one of the participants in Figure~\ref{fig:alignment1}. Here we observe that the trajectories of our model (in orange) are smooth and do not detour during the tasks, while the trajectories for No Align (in grey) and No Priors (in black) have many movements that counteract themselves, indicating that this user was struggling to understand and align with the control strategy. In the worst case, participants were unable to complete the task with the No Priors model (see the \textit{Avoid} task in Figure~\ref{fig:alignment1}) because no joystick inputs mapped to their intended direction, effectively causing them to get stuck at undesirable states.

To further validate that our model is learning the human's preferences, we illustrate heatmaps over user inputs for the \textit{Avoid} task in Figure~\ref{fig:alignment2}. Recall that this task requires moving the robot around an obstacle. Without the correct alignment, users default to the four cardinal directions (No Align), or warped circle-like motions (No Priors). By contrast, under All Priors the users smoothly moved the joystick around an even distribution, taking full advantage of the joystick's $[-1, +1]$ workspace along both axes.

\p{Summary} In this user study we explored different approaches for learning the alignment model between joystick inputs and latent actions. We compared learning the alignment model with and without intuitive priors, and found that including these priors improved user performance, particularly in challenging tasks (\textbf{H1}). 
\section{Case Study with Disabled Users}\label{disabled}

\begin{figure}[t]
	\begin{center}
		\includegraphics[width=1\columnwidth]{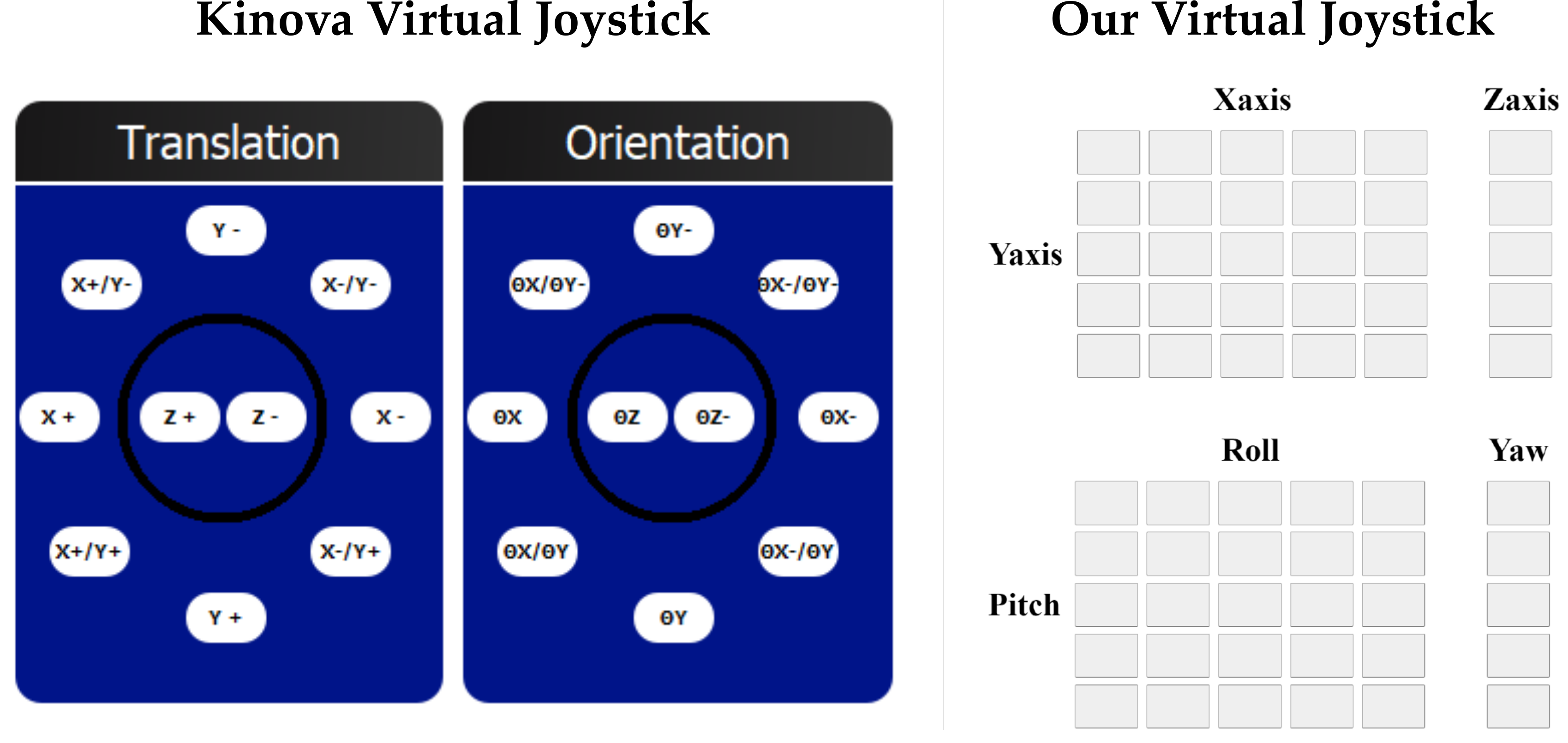}

		\caption{Comparison between the End-Effector condition on a Kinova assistive robot arm \cite{Kinova} (left) and our End-Effector implementation (right). We introduced a grid so that this virtual joystick was as close as possible to the continuous joysticks we previously used during our studies with non-disabled persons. We also consulted with our users while creating this interface to ensure that it matched their expectations.}
		\label{fig:pilot5}
	\end{center}
	
	\vspace{-1em}

\end{figure}

Our previous user studies involved non-disabled participants. Overall, these studies demonstrate the potential advantages of learned latent actions and support our proposed algorithm. But we still need to determine whether these results transfer to our target population; accordingly, here we test whether \textit{disabled} persons can leverage latent actions to teleoperate assistive robots during eating tasks. This case study explores conditions and tasks similar to the user study from Section~\ref{study3}, but now with the added dimension of two disabled users that are familiar with assistive robot arms.

\p{Participants} We recruited two adult males with disabilities (ages $28$ and $42$) who require assistance when eating. The first participant has three years of experience with assistive robot arms, and the second participant has two years of experience with these robots.

\p{Experimental Setup} In order to ensure safety during the COVID-19 pandemic, we developed a \textit{remote control} interface where participants teleoperated an on-campus robot in real-time from their own homes (see Figure~\ref{fig:pilot1}). Participants interacted with a virtual joystick while watching live-streamed video of the robot arm. Video showed the robot and task from two angles: a front-view and a side-view. Just like the previous in-person studies, we used the participant's joystick inputs to control the $7$-DoF motion of the robot arm. We worked with both participants to minimize communication latency and position the cameras effectively; however, we recognize that delays and depth-perception may affect the results of these experiments.

As in Section~\ref{study3}, participants attempted to assemble two dishes: an \textit{Entree} task and a \textit{Dessert} task. The \textit{Entree} task involved reaching for a block of tofu and then precisely cutting off a slice. The \textit{Dessert} task had three steps: i) reaching for and stabbing a marshmallow, ii) carefully scooping that marshmallow in icing, and then iii) dipping the marshmallow in sprinkles.

\p{Independent Variables} We compared two different control schemes: \textbf{End-Effector} and \textbf{LA+SA}. With End-Effector the user directly controls the position and orientation of the fork attached to the end of the assistive robot arm \cite{herlant2016assistive,newman2018harmonic,javdani2018shared}. Participants here interact with two separate sets of joysticks, one for linear motion and a second for angular motion. We collaborated with both participants to ensure that this End-Effector setup closely resembled the control interface they typically use on their assistive robot arms (see Figure~\ref{fig:pilot5}). The LA+SA condition is our proposed approach (Algorithm~\ref{deltaco}), which combines latent actions with shared autonomy. We omit the alignment model $f$ because of the time constraints of our voluntary participants; thus, the latent action $z$ is set equal to the joystick input $u$.

\p{Dependent Measures} To understand the effects of learned latent actions, we measured the total time taken to complete each task (\textit{Total Time}) and the amount of time during the task where users were not providing joystick inputs (\textit{Idle Time}). We also recorded the distance between the robot's fork and the closest goal --- i.e., the tofu, marshmallow, icing, or sprinkles. This \textit{Goal Error} serves as a proxy metric for the accuracy of the robot's high-level reaching. After participants completed the entire experiment we administered a short, open-ended survey to elicit their free-response feedback.

\begin{figure}[t]
	\begin{center}
		\includegraphics[width=1\columnwidth]{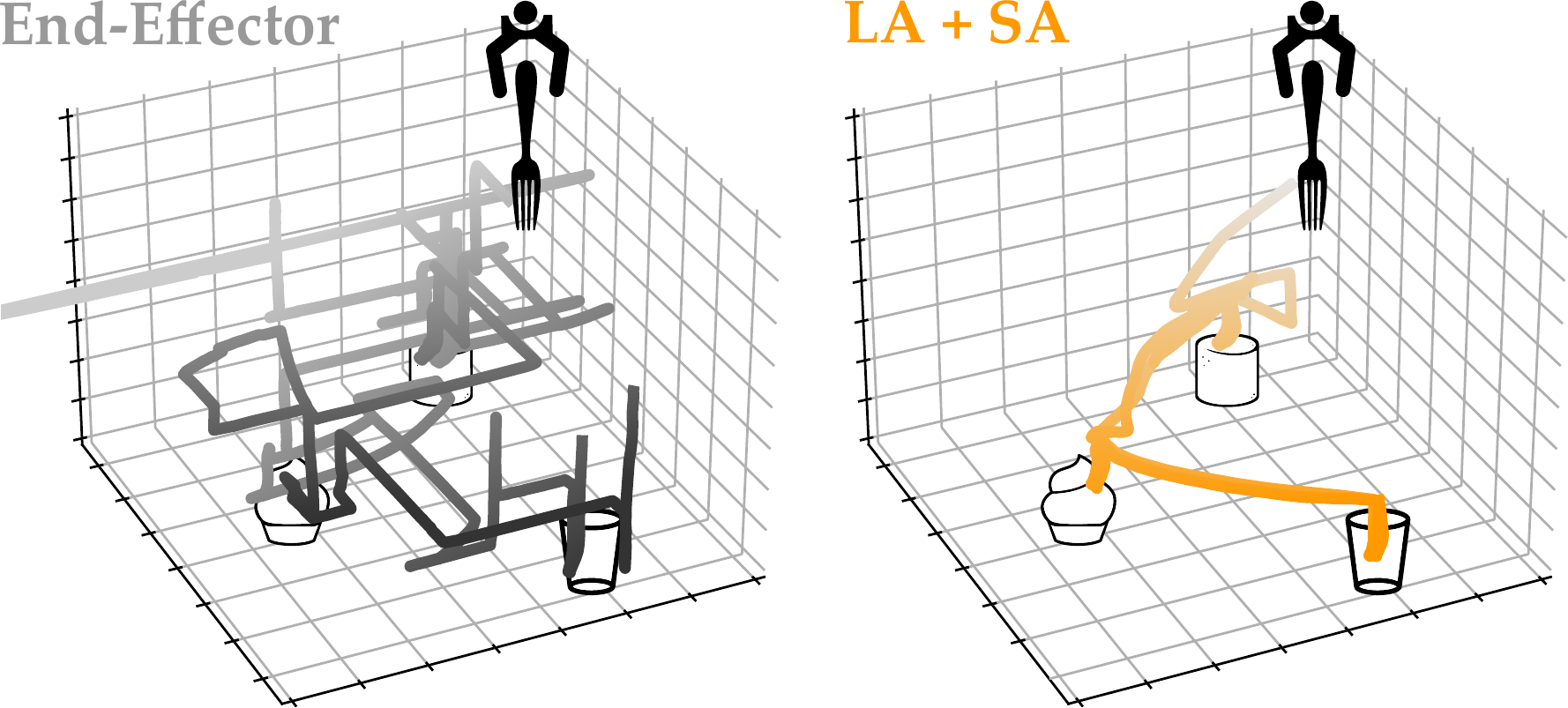}

		\caption{The trajectories the robot's fork follows with \textbf{End-Effector} and \textbf{LA+SA} conditions in the \textit{Dessert} task. The fork starts above the goals and moves to stab a marshmallow, scoop it in icing, and then dip it in a cup of sprinkles. We overlay the trajectories for both disabled participants.}
		\label{fig:pilot2}
	\end{center}

\end{figure}

\begin{figure}[t]
	\begin{center}
		\includegraphics[width=1\columnwidth]{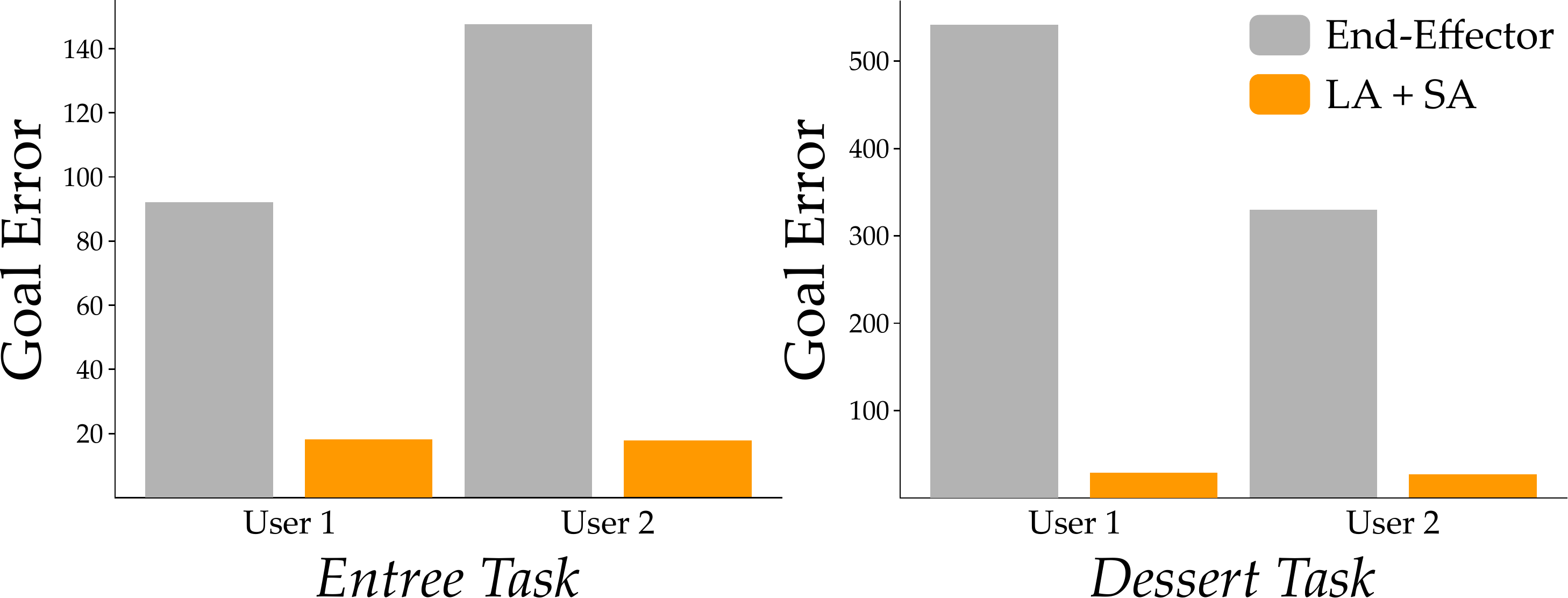}

		\caption{Error between the robot's fork and the closest goal during both \textit{Entree} and \textit{Dessert} tasks. We separately show the results for each disabled user.}
		\label{fig:pilot3}
	\end{center}

\end{figure}

\begin{figure}[t]
	\begin{center}
		\includegraphics[width=1\columnwidth]{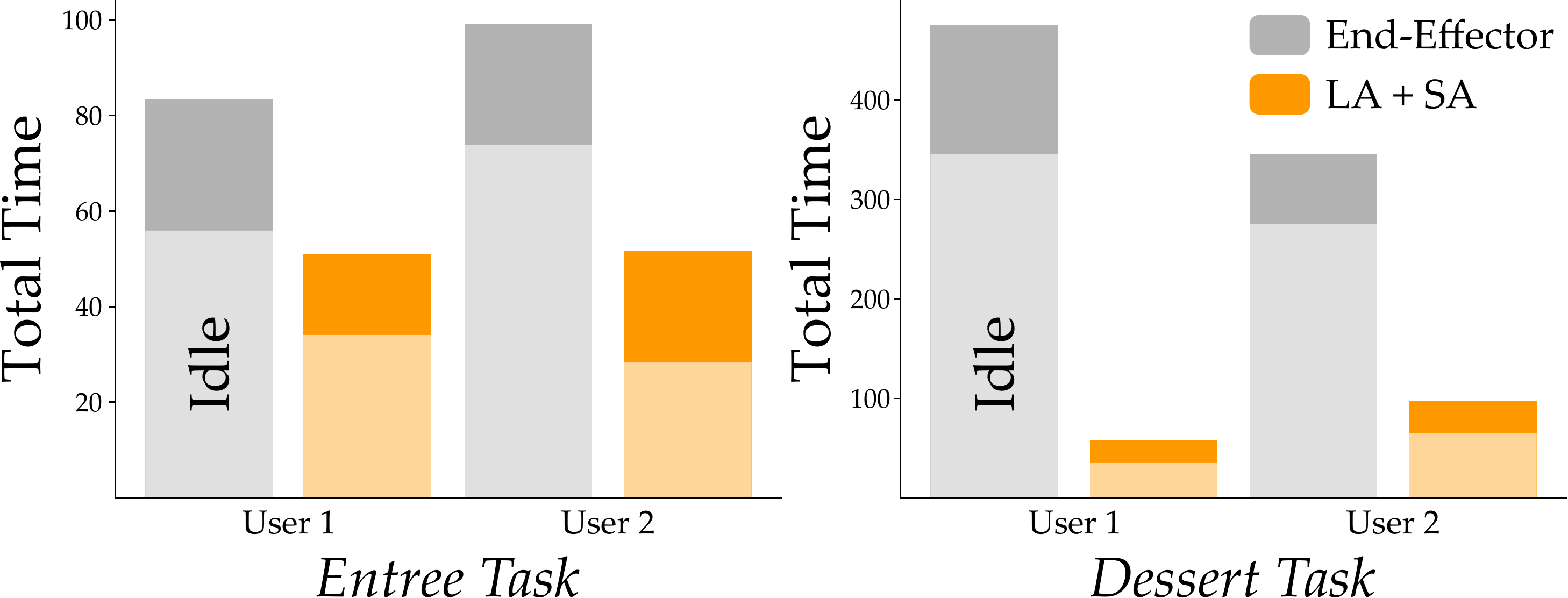}

		\caption{Total time taken to complete the task (solid) and the idle time where users were trying to select their joystick input (light). Time is measured in seconds. As before, we separately show the results for each disabled user. Notice that the \textit{Dessert} task took both users over $5$ minutes with \textbf{End-Effector}, but less than $2$ minutes with \textbf{LA+SA}.}
		\label{fig:pilot4}
	\end{center}

\end{figure}

\p{Procedure} We applied a within-subjects design, where both participants completed the Dessert and Entree tasks using End-Effector and LA+SA conditions. User 1 started with the End-Effector condition and User 2 started with LA+SA. Both users completed the Dessert task first before doing the Entree task. So that the users could familiarize themselves with the controller and environment, participants were allotted up to $5$ minutes of practice time with each condition.

\p{Hypotheses} We tested two main hypotheses:
\begin{displayquote}
\textbf{H1.} \emph{Disabled users will complete the eating tasks more quickly and accurately with our combination of latent actions and shared autonomy.}
\end{displayquote}
\begin{displayquote}
\textbf{H2.} \emph{Disabled users will subjectively prefer LA+SA to their current control approach (End-Effector).}
\end{displayquote}

\p{Results} The objective results of our case study are visualized in Figures~\ref{fig:pilot2}, \ref{fig:pilot3}, and \ref{fig:pilot4}. In Figure~\ref{fig:pilot2} we display the motion of the robot's end-effector for the \textit{Dessert} task: comparing the trajectories, we observe that End-Effector resulted in disjointed motions where users constantly switched between controlling the $x$, $y$, or $z$ position of the robot's fork. By contrast, LA+SA helped users smoothly and directly move between goals.

To explore \textbf{H1} we specifically analyzed the \textit{Goal Error} and \textit{Total Time} for both tasks and participants (Figures~\ref{fig:pilot3} and \ref{fig:pilot4}). Across disabled users we noticed a clear trend: our proposed LA+SA helped participants accurately reach their high-level goals while reducing the \textit{Total Time} and \textit{Idle Time} required to complete the tasks. These trends match our results with non-disabled users (see Section~\ref{study3}), and suggest that combining latent actions with shared autonomy improves objective performance during eating tasks.

We also asked for free-form feedback after the experiment to better understand the perspective of disabled users. The participants' responses generally supported \textbf{H2}. User $1$ stated that\footnote{We have added the condition names for clarity in these quotes. Participants did not know what the conditions were during the user study, and referred to them as ``the first one'' or ``the second one.''}: 
\begin{displayquote}
\textit{``Comparing End-Effector to LA+SA, the former was a lot harder, and LA+SA was way easier in probably every aspect. LA+SA was a little bit confusing in the sense that I wasn't sure right off the bat what the joystick directions meant, but after using it for a minute or two it was really, really intuitive. Overall, LA+SA was great and I would definitely use it.''}
\end{displayquote}
User 2 similarly mentioned that:
\begin{displayquote}
   \textit{ ``With LA+SA it was much easier to get those broad strokes of where I wanted to go, and more intuitive in how the robot moved, together with a simpler interface.''}
\end{displayquote}

\p{Summary} This small-scale case study with two disabled users compared their typical end-effector control interface to our proposed latent action approach. Both participants performed better with learned latent actions (\textbf{H1}): they moved the robot closer to their high-level goals and completed the task in less total time. Users also perceived latent actions as a better approach for the two assistive eating tasks (\textbf{H2}).
\section{Conclusion}

Our user and case studies separately and collectively test the key parts of our approach from Algorithm~\ref{deltaco}. We learn latent actions from kinesthetic demonstrations, and then enable users to control assistive robots with these learned actions. Our results demonstrate that controlling robots with learned latent actions outperforms the baseline shared autonomy dataset as well as direct end-effector teleoperation. Incorporating shared autonomy with latent actions further increases performance: shared autonomy helps the user reach and maintain high-level goals, while latent actions focus on low-level, precise manipulation. We also leveraged our semi-supervised approach to learn the human's personalized alignment between joystick inputs and latent actions --- in practice, this reduced the number of queries the human needed to answer, and resulted in more efficient task completion than one-size-fits-all alternatives. Overall, our latent action approach is a step towards intuitive, user-friendly control of assistive feeding robots. Our case study with two disabled users suggests that, in practice, controlling robots with learned latent actions makes assistive eating easier.

\p{Limitations} One key limitation of our approach occurs when the user encounters a new task never seen when training the latent actions. Here we cannot rely on latent actions --- but we can revert to a baseline teleoperations scheme (e.g., end-effector control), and let the user complete the task using this default teleoperation mapping. We then include the demonstrated behavior within $\mathcal{D}$, retrain, and leverage learned latent actions \textit{the next time} we encounter this new task. We emphasize that disabled persons can always provide new demonstrations by reverting to the standard, pre-defined teleoperation scheme to control the robot. However, directly retraining our learned latent actions on this updated dataset presents some new challenges: i) determining if we have seen enough data so that the learned latent actions will perform robustly and ii) ensuring that learning new latent actions does not interfere with or override previously learned latent actions. Our future work focuses on this challenge --- we envision assistive robots that continuously alternate between learned latent actions and end-effector control to balance between compact, intuitive embeddings and full, high-dimensional control over the robot.

\bibliographystyle{spmpsci}
\bibliography{citations}


\end{document}